\documentclass[letterpaper]{article} 
\usepackage[preprint]{aaai2027}  
\usepackage[hyphens]{url}  
\usepackage{graphicx} 
\usepackage{natbib}  
\usepackage{caption} 
\usepackage{algorithm}
\usepackage{algpseudocode}
\usepackage[most]{tcolorbox}
\usepackage{animate}
\usepackage{amsmath}
\usepackage{amsfonts}
\usepackage{array} 
\usepackage{minibox}
\usepackage{subcaption}
\usepackage{multirow}
\usepackage{pifont}
\usepackage{makecell}
\newcommand{\cmark}{\ding{51}}
\newcommand{\xmark}{\ding{55}}
\usepackage[table]{xcolor}

\definecolor{bgheader}{RGB}{221,235,247}    
\definecolor{geoheader}{RGB}{252,228,214}  
\definecolor{fgheader}{RGB}{252,228,214}
\definecolor{baseheader}{RGB}{242,242,242} 
\definecolor{videoheader}{RGB}{232,226,242}
\definecolor{bgtext}{RGB}{70,120,165}
\definecolor{geotext}{RGB}{190,105,65}
\definecolor{fgtext}{RGB}{190,105,65}
\definecolor{videotext}{RGB}{105,80,145}
\usepackage{newfloat}
\usepackage{listings}
\DeclareCaptionStyle{ruled}{labelfont=normalfont,labelsep=colon,strut=off} 
\floatstyle{ruled}
\newfloat{listing}{tb}{lst}{}
\floatname{listing}{Listing}

\usepackage{booktabs}
\newcommand{\abbr}{\textbf{ScaleVid}}
\title{ScaleVid: Geometry-Aware Video Object Scaling with Mesh-Free Inference}
\author{
    Youze Huang\equalcontrib,
    Penghui Ruan\equalcontrib,
    Bojia Zi\equalcontrib,
    Xianbiao Qi\corresponding,
    Shihao Zhao,
    Xiao Rong
}
\affiliations{

}

\title{ScaleVid: Geometry-Aware Video Object Scaling with Mesh-Free Inference}
\author {
    Youze Huang\textsuperscript{\rm 1}\equalcontrib,
    Penghui Ruan\textsuperscript{\rm 2}\equalcontrib,
    Bojia Zi\textsuperscript{\rm 3}\equalcontrib,
    Xianbiao Qi\textsuperscript{\rm 4}\corresponding,
    Shihao Zhao\textsuperscript{\rm 5},
    Xiao Rong\textsuperscript{\rm 4}
}
\affiliations {
    \textsuperscript{\rm 1}University of Electronic Science and Technology of China\\
    \textsuperscript{\rm 2}The Hong Kong Polytechnic University\\
    \textsuperscript{\rm 3}The Chinese University of Hong Kong\\
    \textsuperscript{\rm 4}Intellif Inc.\\
    \textsuperscript{\rm 5}The University of Hong Kong\
}

\begin{document}

\maketitle

\begin{abstract}
Geometry-aware video object scaling aims to anisotropically resize the object along object-centric axes while preserving geometric plausibility, temporal coherence, and background consistency. Existing text-guided methods mainly operate in the 2D image plane, while depth-guided approaches provide coarse control and mesh-based methods require costly 3D reconstruction. We present a progressive two-stage training framework that decouples geometry-aware foreground transformation from background preservation and realistic video composition, without mesh-pixel alignment and explicit 3D reconstruction at inference. In both stages, geometrically perturbed pseudo-sources are constructed from real videos, while the original complete videos are retained as reconstruction targets. The first stage uses planar transformations to learn robust foreground-background composition, whereas the second introduces object-centric 3D deformation guidance for geometry-aware scaling. This pseudo-source reconstruction formulation enables real-video synthesis without paired real-world scaling targets. We construct complementary paired-geometry and real-background benchmarks and further evaluate on in-the-wild videos. Extensive experiments demonstrate superior geometric consistency, foreground fidelity, and background preservation, together with faster and more practical inference than methods requiring explicit 3D reconstruction.
\end{abstract}


\section{Introduction}

\begin{figure*}[!t]
    \centering
      \includegraphics[width=0.985\linewidth]{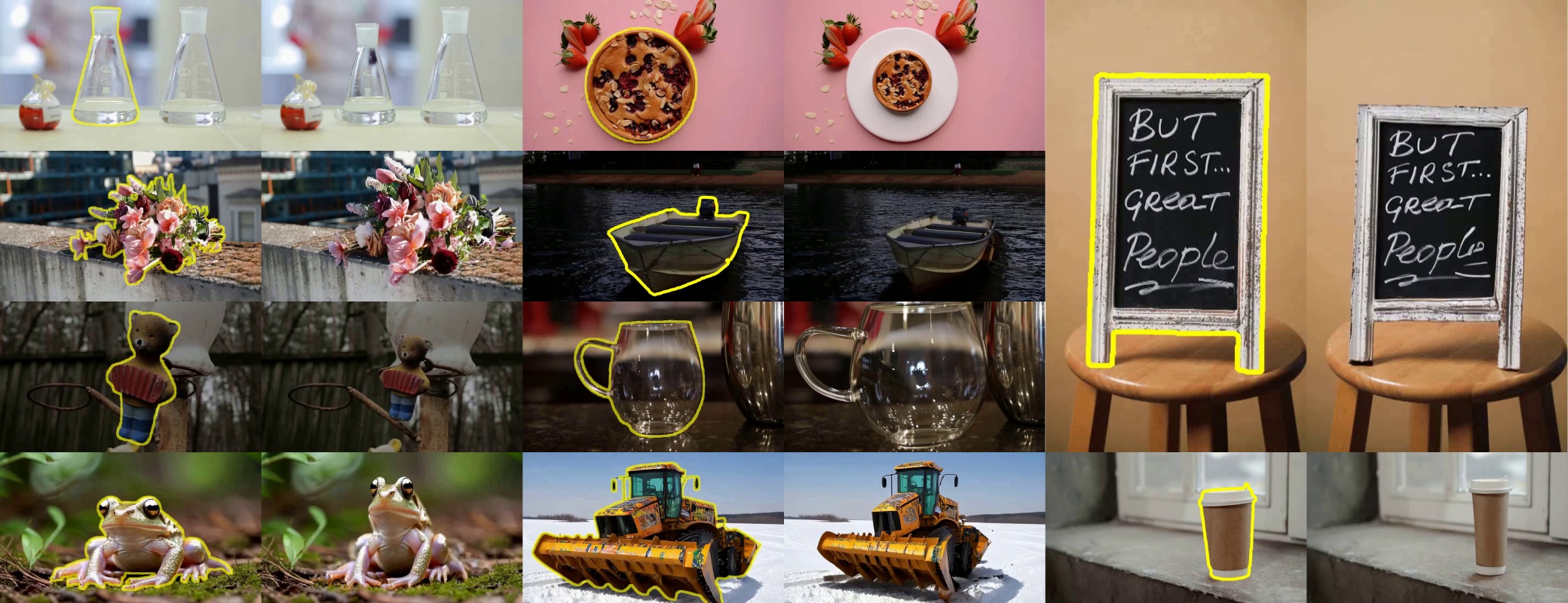}
      \caption{
Visual results of ScaleVid on diverse objects and scenes.
In each pair, the left frame is the source, and the right frame is the scaled result.
The examples cover transparent and geometrically complex objects, anisotropic scaling along object-centric width, height, and depth axes, isotropic enlargement and shrinkage, and inputs of varying resolutions.
}
      \label{fig:teaser}
\end{figure*}

Diffusion models have achieved remarkable progress in visual generation and editing, spanning image inpainting, instruction-based image manipulation~\cite{blattmann2023align, rombach2022high, zhao2024ultraedit, hui2025hqedit, chen2023pixart, wang2025styleadapter}, and video editing~\cite{guo2024animatediff, chen2024videocrafter2, yang2024cogvideox, wan2025wan, zhuang2024task,  wu2025qwen}. 
By learning strong appearance priors from diverse data, these models can synthesize highly realistic content and have improved the quality of controllable visual generation.

However, many real-world editing tasks are not purely appearance-driven, but inherently geometric. 
Representative examples are object rotation, translation and scaling, where the target object should change its position or size while remaining geometrically plausible under perspective projection. 
To address this limitation, a number of works have attempted to incorporate 3D knowledge into diffusion-based editing
~\cite{liu2026reconx,wang2026objctrl,chen2026disco,hu2025ex,ruan2026ctrl,chen2025blenderfusion,wu2024neural,he2026geoeditgeometryawareobjectediting}. Moreover, some pixel-to-mesh works~\cite{liu2023zero,zhao2025hunyuan3d,sabathier2026actionmesh} also facilitate incorporating 3D knowledge into pixel domain. 
These methods show that geometric priors are crucial for moving beyond purely 2D manipulation.

Despite recent progress, existing methods remain limited for
controllable video object scaling. Text- and depth-guided approaches
provide imprecise geometric control, while explicit 3D pipelines
require reconstruction, camera estimation, rendering, and sometimes
manual editing. Extending such control to videos is further
complicated by temporal consistency and the absence of paired
real-world training videos with controllable geometry-aware
transformations. Recent attempts to address this issue by constructing
paired training data with 3D assets still require expensive
video-specific 3D reconstruction and frame-wise alignment between
meshes and real videos, limiting scalability for dataset.

In this paper, we study a sub-task: geometry-aware video object scaling, with a particular focus on controllable 3D object scaling in videos. 
Our goal is to scale the object according to user-specified 3D scaling factors, while preserving geometric plausibility, temporal coherence, object realism, and background consistency, without resorting to explicit 3D reconstruction at inference time. Controllable object scaling is useful for short-video editing, advertising,  and AR/VR, where object size affects composition and visual emphasis.

Our key insight is to decouple controllable geometric transformation
from realistic video synthesis through pseudo-source reconstruction.
Since paired real-world scaling videos are unavailable, we construct
geometrically perturbed pseudo-sources while always retaining the
original complete videos as real-video targets. Our design allows geometric
control to be learned from paired geometric supervision, while
composition, appearance refinement, temporal coherence, and background
preservation are learned in the real-video domain, without explicit
mesh--pixel alignment. We further adopt a progressive 2D-to-3D strategy:
Stage~I learns robust composition from planar transformations, and
Stage~II introduces object-centric 3D deformation guidance, enabling
geometry-aware reshaping upon the synthesis prior learned in Stage~I.
Our contributions are three-fold:
\begin{itemize}
    \item We formulate video object scaling through object-centric anisotropic factors and learn its projected geometric effects without explicit 3D reconstruction at inference.
    \item We propose a progressive 2D-to-3D training strategy that constructs pseudo-sources while always using the original
complete real video as the target, enabling robust composition learning
from planar transformations and geometry-aware scaling from 3D
deformation guidance.
    \item We establish complementary paired-geometry, real-background, and real-world video evaluations for scale control, geometric alignment, appearance fidelity, background preservation, and temporal quality.
\end{itemize}

\begin{figure*}[!t]
    \centering
    \includegraphics[width=0.985\textwidth]{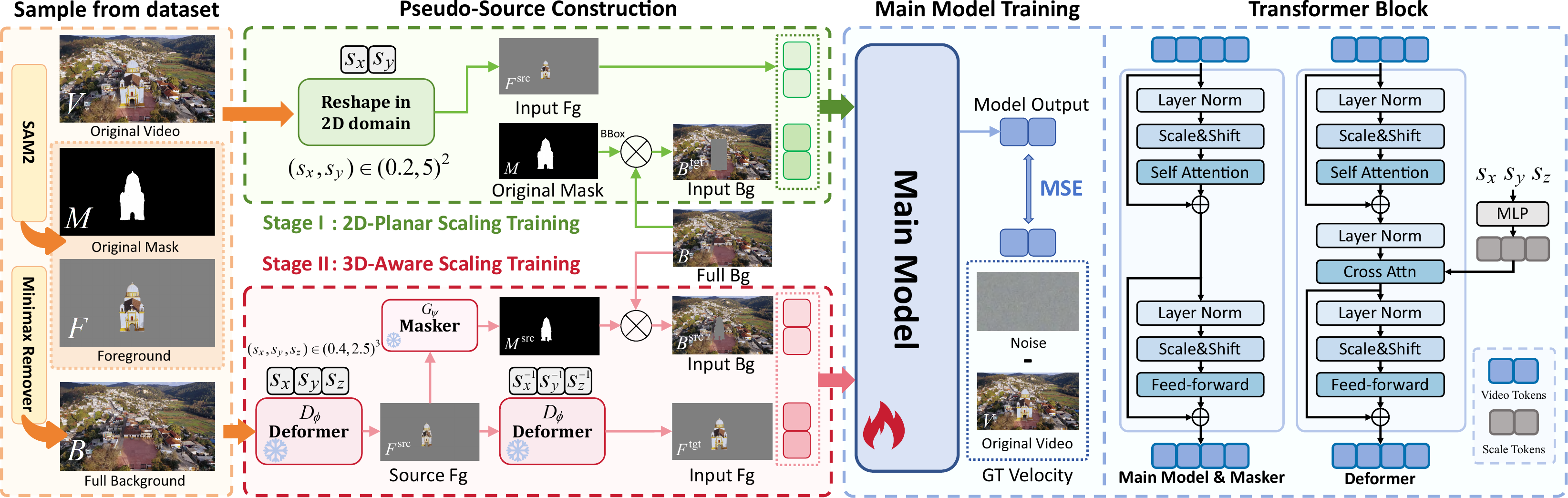}
    \caption{
Overview of the ScaleVid training pipeline and model architecture.
{Left}: given a video $V$, its mask $M$, and completed
background $B$, Stage I constructs planar pseudo-sources using
$(s_x,s_y)$ and a bounding-box background mask, whereas Stage II uses
the pretrained Deformer and Masker to construct geometry-aware
foreground, mask, and background conditions under
$(s_x,s_y,s_z)$.
{Middle}: the Main Model is trained with flow matching to reconstruct the
original complete video.
{Right}: model architecture. We remove cross-attention modules from the Main Model and Masker.
}
    \label{fig:train_pipe}
\end{figure*}

\section{Related Works}
Recent works highlight the importance of geometric priors for overcoming the limitations of purely 2D diffusion-based editing. 
Early attempts mainly exploit depth or pseudo-3D cues. Parihar et al.~\cite{parihar2025zero} propose a training-free framework that decomposes scenes into depth-ordered layers, while GeoDiffuser~\cite{sajnani2025geodiffuser} injects geometric transformations into diffusion attention through DDIM inversion for object-level manipulation. Beyond depth-based methods, several works explicitly lift images into 3D representations. BlenderFusion~\cite{chen2025blenderfusion} and Image Sculpting~\cite{yenphraphai2024image} reconstruct object-centric meshes, perform editing in external 3D environments, and then re-render and refine the results with diffusion models. 3DIT~\cite{michel2023object} further combines language guidance with 3D-aware scene representations. Another line of work introduces implicit geometry control. ShapeWords~\cite{petrov2025shapewords} encodes 3D shape information into tokenized embeddings for text-to-image diffusion models, and Neural Assets~\cite{wu2024neural} adopts object-level tokens for multi-object generation. 

These ideas have also been extended to video. VideoHandles~\cite{koo2025videohandles} and Shape-for-Motion~\cite{liu2025shape} lift videos into 3D proxy representations and propagate geometry and texture across frames for temporal consistency. Lee et al.~\cite{lee2026generative} use 3D point trajectories as control signals for manipulating object motion and camera dynamics, while Gu et al.~\cite{gu2025diffusion} formulate video generation as coloring tracked 3D points for unified controllable generation. Ctrl\&Shift~\cite{ruan2026ctrl} introduces camera-aware embeddings to enhance geometric controllability, and Refa\c{c}ade~\cite{huang2026refacade} disentangles structure and texture via a texture remover.

\section{Methodology}

\subsection{Problem Formulation and Preliminaries}\label{sec:problem_formulation}

Given an input video
$V=\{I_n\}_{n=1}^{N}$, where
$I_n\in\mathbb{R}^{H\times W\times 3}$,
its corresponding target-object masks
$M=\{M_n\}_{n=1}^{N}$, where
$M_n\in\{0,1\}^{H\times W}$,
and user-specified anisotropic scaling factors
$\mathbf{s}=(s_x,s_y,s_z)\in\mathbb{R}_{>0}^{3}$,
our goal is to generate an edited video
$V'=\{I'_n\}_{n=1}^{N}$ in which the target object is
rescaled according to $\mathbf{s}$, while preserving object
identity, temporal coherence, and background fidelity.

\paragraph{Object-Centric 3D Scaling.}
A geometrically accurate realization of object scaling is to
transform the underlying object geometry in 3D space and then
project the transformed object onto the image plane. Let
$\mathbf{x}_n\in\mathbb{R}^{3}$ denote a mesh vertex at frame $n$
in the original mesh-local coordinate system, and let
$\mathbf{c}_n$ denote the scaling center. We define
$\mathbf{A}\in\mathbb{R}^{3\times3}$ as an orthonormal basis whose
columns correspond to the canonical object-centric axes. The desired
anisotropic scaling is performed in this canonical coordinate system:
\begin{equation}
\label{equ:rescale}
\mathbf{x}'_n
=
\mathbf{c}_n
+
\mathbf{A}\mathbf{S}\mathbf{A}^{\top}
\left(\mathbf{x}_n-\mathbf{c}_n\right),
\,
\mathbf{S}
=
\operatorname{diag}(s_x,s_y,s_z).
\end{equation}
Here, $\mathbf{x}'_n$ is the transformed vertex, $\mathbf{A}^{\top}$ transforms the centered vertex into the
canonical object-centric frame, $\mathbf{S}$ applies axis-wise
scaling, and $\mathbf{A}$ maps the transformed vertex back to the
original mesh-local coordinate system.

\paragraph{Canonical Axis Alignment.}
In this paper, mesh is stored as a standard 3D asset (GLB in our
implementation). Its vertices are represented in an asset-specific
mesh-local coordinate system, whose axes are not necessarily
consistent across different objects. Directly applying
$\mathbf{S}$ along these raw axes may therefore cause the same
scaling factor to control different geometric directions.

We estimate an oriented bounding box (OBB) of the mesh geometry, use its three orthogonal directions as the dominant object-centric
axes and use the center of the OBB as the scaling center $\mathbf{c}_n$. However, the OBB axes are ambiguous in ordering,
and therefore do not by themselves provide a consistent axis
convention across objects. To assign unified scaling semantics, we
align them with the renderer's right-handed canonical frame, in which
the $y$-axis is upright and the $x$- and $z$-axes denote width and
depth, respectively. Specifically, we determine the one-to-one axis
assignment by maximizing the absolute directional agreement with the
renderer axes, obtaining the basis $\mathbf{A}$ used in
Eq.~\eqref{equ:rescale}. Finally, the transformed surface points are
projected onto the image plane by the renderer. $\mathbf{A}$ is fixed across frames to ensure temporal consistency, while the scaling center $\mathbf{c}_n$ is recomputed from the current-frame OBB for deformable objects.


\paragraph{Conditional Flow Matching.}
Given a target sample
$\boldsymbol{x}_1$,
we sample Gaussian noise
$\boldsymbol{x}_0 \sim \mathcal{N}(\boldsymbol{0},\boldsymbol{I})$
and $t\sim\mathcal{U}(0,1)$, then construct the interpolated noisy sample
$\boldsymbol{x}_t=(1-t)\boldsymbol{x}_0+t\boldsymbol{x}_1$,
with target velocity
$\boldsymbol{v}=\boldsymbol{x}_1-\boldsymbol{x}_0$.
For a predictor $g$ with condition
$\boldsymbol{c}$, the training objective is
\begin{equation}
\mathcal{L}_{\mathrm{FM}}
(g;\boldsymbol{x}_1,\boldsymbol{c})
=
\mathbb{E}_{t,\boldsymbol{x}_0}
\left[
\left\|
g(\boldsymbol{x}_t,t,\boldsymbol{c})
-
\boldsymbol{v}
\right\|_2^2
\right].
\label{eq:general_fm}
\end{equation}




\subsection{ScaleVid Overview}

ScaleVid performs geometry-aware object scaling directly in the
video domain, without explicit 3D reconstruction, camera estimation,
or mesh rendering at inference time. Our key idea is to decouple
geometry transformation from high-quality video synthesis through progressive pseudo-source reconstruction. The original complete real video is always
used as the reconstruction target. The pipeline
is shown in Fig.~\ref{fig:train_pipe}.
All generative modules in ScaleVid operate in the latent space of a pretrained video VAE. For simplicity, we omit the latent encoding operations in the following formulations.

\subsection{Geometry-Aware Guidance Modules}
\label{sec:deformer_masker}

We first describe how the geometry-aware guidance used by the Main
Model is learned from paired synthetic supervision.

\subsubsection{Synthetic Paired Supervision}

\begin{figure*}[t] \centering \includegraphics[width=0.985\textwidth]{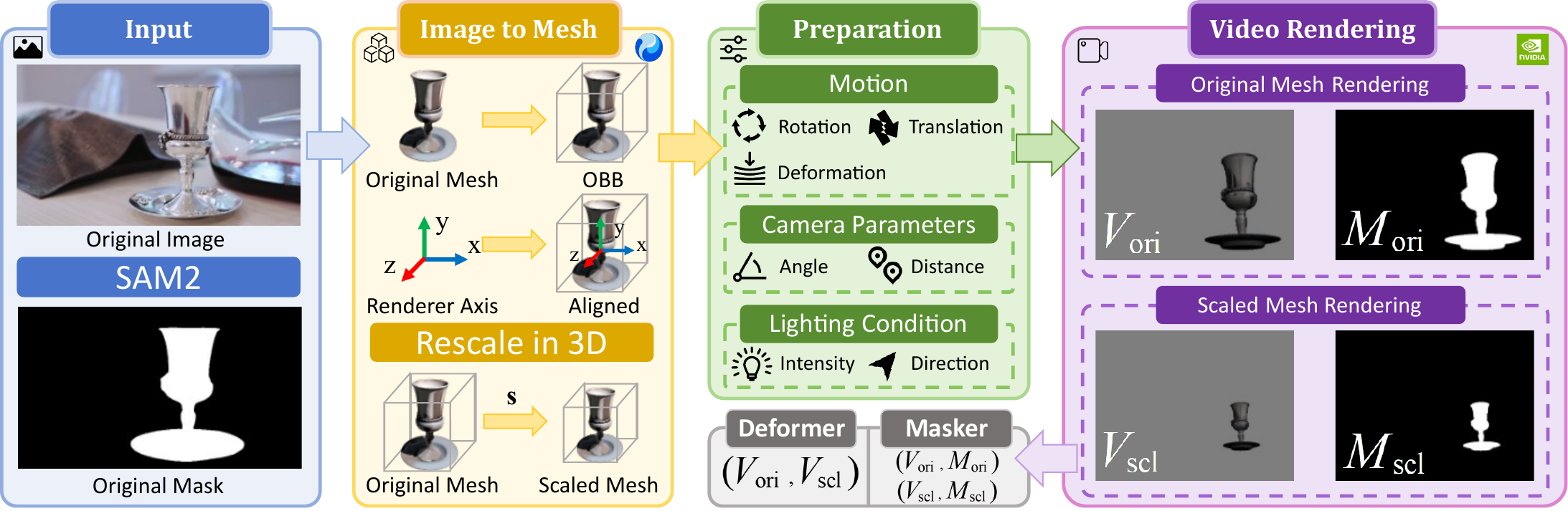} \caption{Overview of our synthetic data construction pipeline for training the Deformer and Masker modules. Given an input image and its SAM2 mask, we reconstruct a 3D mesh,
align its OBB axes with the renderer frame to define consistent
object-centric scaling directions, and apply 
${s}=(s_x,s_y,s_z)$.
The original and scaled meshes are rendered using identical motion,
camera, and lighting configurations, yielding paired tuples 
$(V_{\rm ori},M_{\rm ori},V_{\rm scl},M_{\rm scl},
{s})$ for training the
Deformer and Masker.} \label{fig:dataset} \end{figure*}
To train the Deformer and Masker, we construct a large-scale
synthetic paired-video dataset, as illustrated in
Fig.~\ref{fig:dataset}.
The meshes are reconstructed using Hunyuan3D~\cite{zhao2025hunyuan3d}
and canonicalized through the OBB-based alignment procedure.
We sample anisotropic scaling factors
${s}=(s_x,s_y,s_z)\in\mathbb{R}_{>0}^{3}$ and apply
Eq.~\eqref{equ:rescale} along the resulting object-centric axes.
The canonical basis $\mathbf{A}$ is used only to construct consistent
synthetic supervision and is not required at inference time.
The Deformer instead learns the projected deformation effects
associated with these canonical axes directly from the input video.

The foreground masks are obtained from rasterized mesh silhouettes
using Kaolin~\cite{jatavallabhula2019kaolin}.
Each training sample is represented as
$\left(
V_{\rm ori},M_{\rm ori},
V_{\rm scl},M_{\rm scl},
\mathbf{s}
\right)$,
providing strict supervision for Deformer and Masker.

\subsubsection{Deformer}

The Deformer $D_{\phi}$ is a latent-space conditional flow model
that learns geometry-aware transformations according to continuous
anisotropic scaling parameters. We denote the original foreground video and its scaled counterpart as 
$V_{\mathrm{ori}}$ and $V_{\mathrm{scl}}$. 
Scaling factor $\mathbf{s}$ is projected by an MLP into scale tokens and injected
into the transformer blocks through cross-attention. Conditioned on
$V_{\mathrm{ori}}$ and $\mathbf{s}$, the Deformer's forward objective is
\begin{equation}
\label{equ:deformer_forward}
\mathcal{L}_{\mathrm{fwd}}
=
\mathcal{L}_{\mathrm{FM}}
\left(
D_{\phi};
V_{\mathrm{scl}},
\{V_{\mathrm{ori}},\mathbf{s}\}
\right).
\end{equation}

\subsubsection{Bidirectional Training}

The Deformer is applied in both scaling directions in our training
pipeline: it first constructs a geometrically perturbed foreground
and is subsequently applied with the inverse scaling factors to
produce target-aligned guidance. We therefore encourage the learned
transformation to remain consistent under bidirectional scaling.

Since direct cycle reconstruction requires backpropagation through
two multi-step sampling trajectories, we instead avoid end-to-end
cycle optimization and impose reverse supervision in a step-wise
manner at the velocity-prediction level. Specifically, the inverse
objective is
\begin{equation}
\label{equ:deformer_inverse}
\mathcal{L}_{\mathrm{inv}}
=
\mathcal{L}_{\mathrm{FM}}
\left(
D_{\phi};
V_{\mathrm{ori}},
\{V_{\mathrm{scl}},\mathbf{s}^{-1}\}
\right).
\end{equation}
Combining \eqref{equ:deformer_forward} and \eqref{equ:deformer_inverse}, the bidirectional objective is
\begin{equation}
\mathcal{L}_{\mathrm{bi}}
=
\frac{1}{2}\mathcal{L}_{\mathrm{fwd}}
+
\frac{1}{2}\mathcal{L}_{\mathrm{inv}}.
\end{equation}
During training, we use the standard forward objective with
probability $1-\lambda$ and the bidirectional objective with
probability $\lambda$, where $\lambda \in [0,1]$. The expected objective is
\begin{equation}
\label{equ:deformer_total}
\mathcal{L}_{\mathrm{def}}
=
(1-\lambda)\mathcal{L}_{\mathrm{fwd}}
+
\lambda\mathcal{L}_{\mathrm{bi}}.
\end{equation}

\subsubsection{Masker}

The Masker $G_{\psi}$ predicts the temporally consistent spatial
support of a transformed foreground video. It is trained using both
directions of the rendered supervision:
\[
G_{\psi}(V_{\mathrm{ori}})
\rightarrow M_{\mathrm{ori}},
\qquad
G_{\psi}(V_{\mathrm{scl}})
\rightarrow M_{\mathrm{scl}}.
\]
The Masker adopts the same latent flow-matching backbone as the
Deformer and is optimized using the general objective in
Eq.~\eqref{eq:general_fm}, with the corresponding mask video as the
target and the input foreground video as the condition.

Its predicted mask is used to construct the masked background
condition and localize the transformed foreground for the Main Model.
To reduce the computational cost and accelerate the training of the Main Model, we further distill it into
a three-step generator using DMD~\cite{yin2024improved}.

\subsection{Progressive Training of the Main Model}

We progressively train the Main Model using the two stages illustrated
in Fig.~\ref{fig:train_pipe}.
Given a video $V$ and its mask $M$, we first obtain
a clean background $B$ using Minimax Remover~\cite{zi2025minimax}
and extract the foreground $F$. 
With the target $V$ and condition $\mathbf{c}$, the Main Model objective is
\begin{equation}
\label{equ:main_loss}
\mathcal{L}_{\mathrm{main}}
=
\mathcal{L}_{\mathrm{FM}}
\left(
f_{\theta};
V,
\mathbf{c}
\right).
\end{equation}

\subsubsection{Stage I: 2D-Planar Scaling Training}

Given the original mask $M$, we compute its bounding box
$\operatorname{BBox}(M)$ and use the box center as the scaling center.
We randomly sample planar scaling factors
$(s_x,s_y)\in(0.2,5)^2$ and apply them to the foreground $F$,
producing a transformed foreground $F^{\mathrm{src}}$.
We further construct the target background by removing the rectangular
region specified by $\operatorname{BBox}(M)$ from the original
background $B$, i.e.,
$B^{\mathrm{tgt}}=(1-\operatorname{BBox}(M))\odot B$. The condition $\mathbf{c}=\{
F^{\mathrm{src}},B^{\mathrm{tgt}},\operatorname{BBox}(M)\}$.
Bounding box avoids explicit transformed-silhouette leakage. This pseudo-source construction teaches the model to recover a
high-quality object from geometrically perturbed foreground guidance
while completing the removed region and preserving the surrounding
scene. Since only planar transformations are required, this stage can
be trained efficiently on large-scale data.

\subsubsection{Stage II: 3D-Aware Scaling Training}

Stage II introduces geometry-aware transformations using the
pretrained Deformer and Masker. Given the original foreground, we
sample anisotropic scaling factors
$\mathbf{s}=(s_x,s_y,s_z)\in\mathbb{R}_{>0}^{3}$.
Conditioned on the original foreground and $\mathbf{s}$, the Deformer
generates a geometrically transformed foreground
$F^{\mathrm{src}}$, while the Masker predicts its corresponding
spatial support $M^{\mathrm{src}}$.

Directly providing the original foreground to the Main Model would
allow it to learn a trivial copy-and-paste solution. We therefore
apply the Deformer again with $\mathbf{s}^{-1}=(s_x^{-1},s_y^{-1},s_z^{-1})$, using $F^{\mathrm{src}}$ as the visual condition, and resulting in $F^{\mathrm{tgt}}$ which is aligned with the target geometry
but retains realistic deformation and generation errors introduced
by the Deformer. It therefore serves as non-trivial foreground
guidance without exposing the ground-truth foreground.

Finally, we obtain the input background using $B^{\mathrm{src}}
    =B\odot(1-M^{\mathrm{src}})$. Thus the Main Model is conditioned on $\mathbf{c}=\left\{
F^{\mathrm{tgt}},
B^{\mathrm{src}},
M^{\mathrm{src}}
\right\}$. In this way,
the Deformer provides geometry-aware transformation guidance, while
the Main Model focuses on appearance refinement, temporal coherence,
and seamless foreground--background composition.

The complete training procedure is summarized in
Algorithm~1 in the Supplementary.

\paragraph{Inference}
The user specifies $\mathbf{s}$ and selects the object to be edited through SAM2~\cite{ravi2024sam} to provide source mask $M^{\mathrm{src}}$. We directly mask the
selected object from the input video to construct
$B^{\mathrm{src}}$, without using an additional object-removal
model. The Deformer is applied only once to produce the desired
target-aligned foreground $F^{\mathrm{tgt}}$, which is then integrated
with $B^{\mathrm{src}}$ by the Main Model. The learned Masker and
inverse deformation are not required. This works because Stage II training motivates the Main Model to treat its foreground condition's geometry as the specification of the desired output; supplying a single target-scale deformation at inference therefore suffices.



\begin{figure}[t]
  \centering
  \renewcommand{\arraystretch}{1.0}
  \setlength{\tabcolsep}{0pt}
  {%
    \fontsize{8pt}{9.6pt}\selectfont
    \begin{tabular}{@{}*{4}{>{\centering\arraybackslash}m{0.246\linewidth}}@{}}
      \minibox{Original Video} & \minibox{Original Mask} & \minibox{2D Planar Scaling} & \minibox{ScaleVid} \\
    \end{tabular}%
  }

  \vspace{-1pt}

    \centering
    \includegraphics[width=0.985\linewidth]{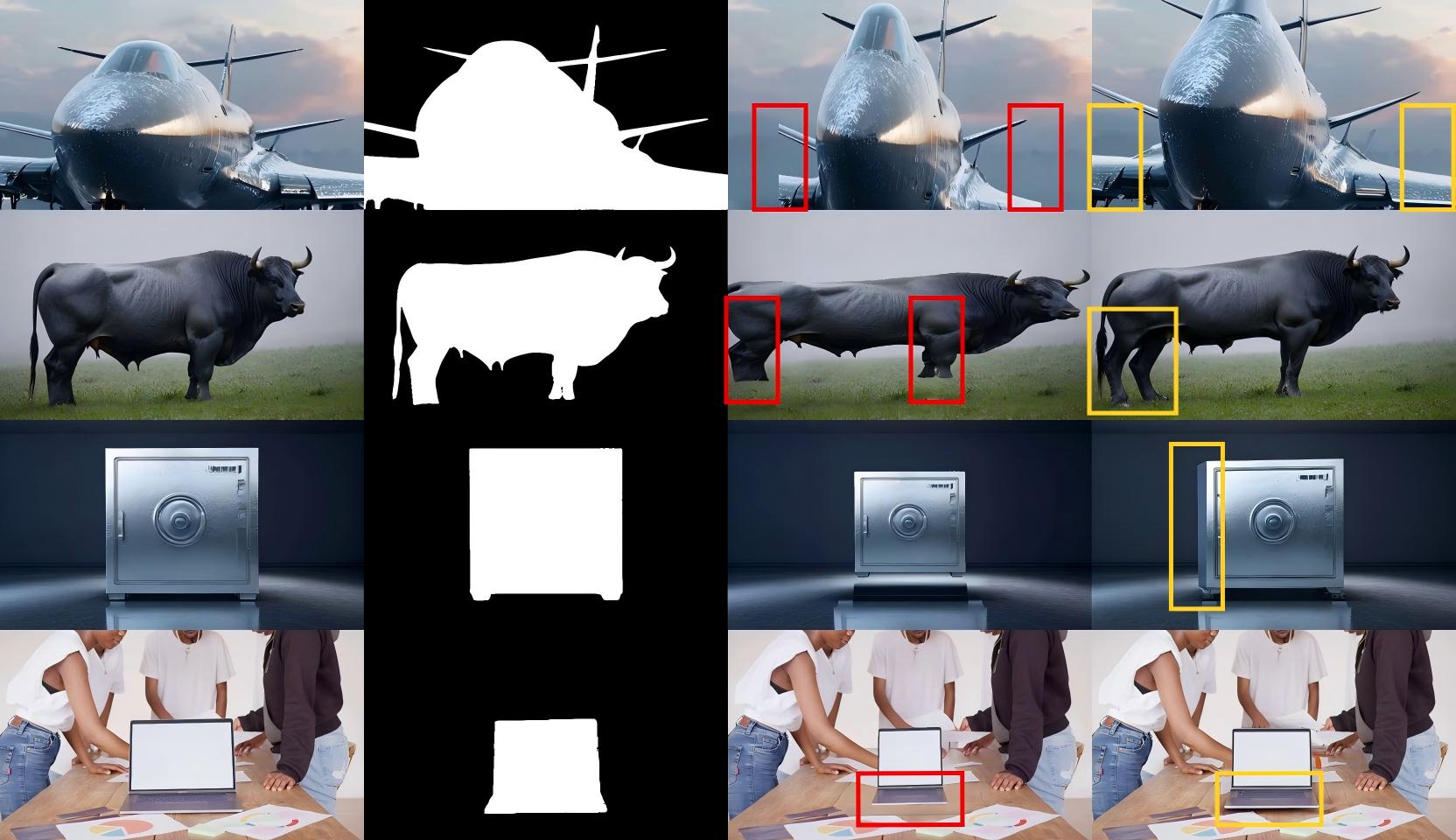}

  \caption{Visual comparison between 2D scaling \& ScaleVid.}
  \label{fig:2d_3d}
\end{figure}

\begin{figure*}[!t]
  \centering
  \renewcommand{\arraystretch}{1.0}
  \setlength{\tabcolsep}{0pt}





  {%
    \fontsize{8pt}{9.6pt}\selectfont
    \begin{tabular}{@{}*{8}{>{\centering\arraybackslash}m{0.123\linewidth}}@{}}
      \minibox{Source} & \minibox{Ground Truth} & \minibox{DiffHandles} & \minibox{Ditto} &
      \minibox{Flux Fill} & \minibox{Flux Kontext} & \minibox{FreeFine} & \minibox{GeoDiffuser}
    \end{tabular}%
  }

  \vspace{-1pt}
  \begin{subfigure}{\linewidth}
    \centering
    \includegraphics[width=0.985\linewidth]{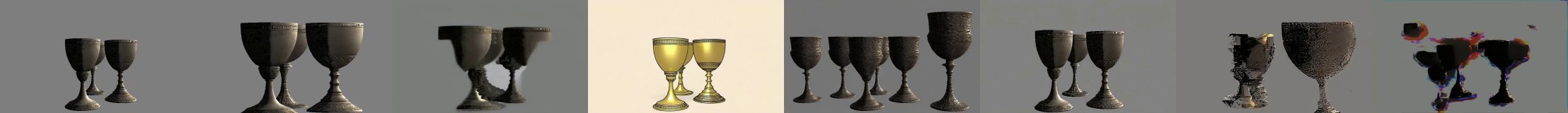}
    
  \end{subfigure}

  {%
    \fontsize{8pt}{9.6pt}\selectfont
    \vspace{-4pt}
    \begin{tabular}{@{}*{8}{>{\centering\arraybackslash}m{0.123\linewidth}}@{}}
      \minibox{HqEdit} & \minibox{InsV2V} & \minibox{InsVIE} & \minibox{LucyEdit} &
      \minibox{Qwen-Img-E} & \minibox{Se\~norita} & \minibox{Shape4Motion} & \minibox{ScaleVid (Ours)}
    \end{tabular}%
  }

  \vspace{-1pt}
  \begin{subfigure}{\linewidth}
    \centering
        \includegraphics[width=0.985\linewidth]{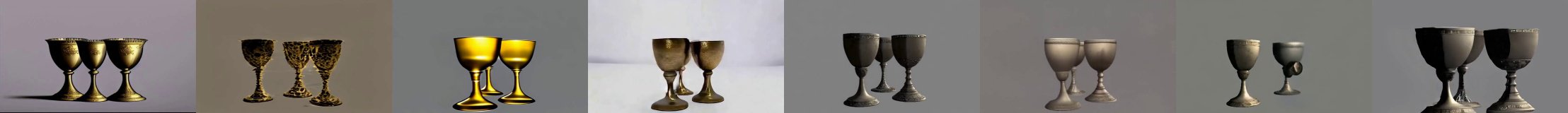}
  \end{subfigure}
  {%
    \fontsize{8pt}{9.6pt}\selectfont
    \begin{tabular}{@{}*{8}{>{\centering\arraybackslash}m{0.123\linewidth}}@{}}
      \minibox{Source} & \minibox{Ground Truth} & \minibox{DiffHandles} & \minibox{Ditto} &
      \minibox{Flux Fill} & \minibox{Flux Kontext} & \minibox{FreeFine} & \minibox{GeoDiffuser}
    \end{tabular}%
  }

  \vspace{-1pt}
  \begin{subfigure}{\linewidth}
    \centering
        \includegraphics[width=0.985\linewidth]{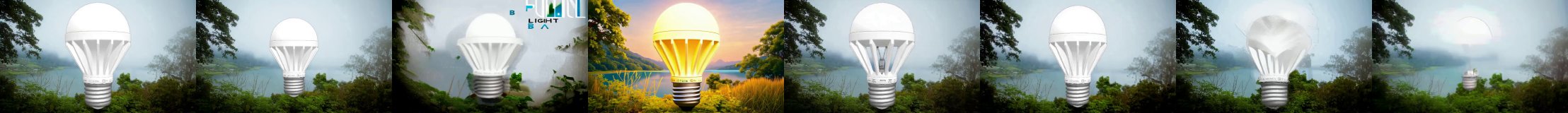}
  \end{subfigure}

  {%
    \fontsize{8pt}{9.6pt}\selectfont
    \vspace{-4pt}
    \begin{tabular}{@{}*{8}{>{\centering\arraybackslash}m{0.123\linewidth}}@{}}
      \minibox{HqEdit} & \minibox{InsV2V} & \minibox{InsVIE} & \minibox{LucyEdit} &
      \minibox{Qwen-Img-E} & \minibox{Se\~norita} & \minibox{Shape4Motion} & \minibox{ScaleVid (Ours)}
    \end{tabular}%
  }

  \vspace{-1pt}
  \begin{subfigure}{\linewidth}
    \centering
    \includegraphics[width=0.985\linewidth]{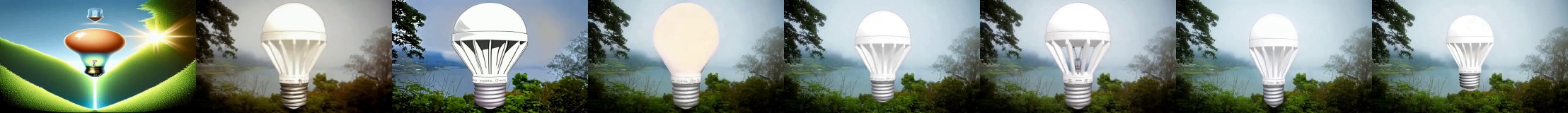}
  \end{subfigure}
  {%
    \fontsize{8pt}{9.6pt}\selectfont
    \begin{tabular}{@{}*{8}{>{\centering\arraybackslash}m{0.123\linewidth}}@{}}
      \minibox{Source} & \minibox{Mask} & \minibox{DiffHandles} & \minibox{Ditto} &
      \minibox{Flux Fill} & \minibox{Flux Kontext} & \minibox{FreeFine} & \minibox{GeoDiffuser}
    \end{tabular}%
  }

  \vspace{-1pt}
  \begin{subfigure}{\linewidth}
    \centering
    \includegraphics[width=0.985\linewidth]{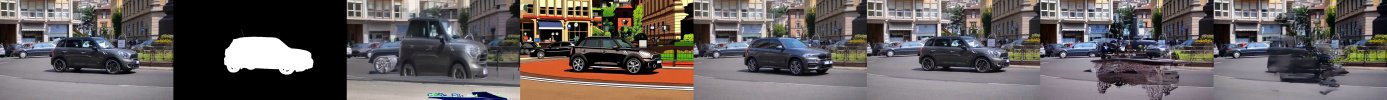}
  \end{subfigure}

  {%
    \fontsize{8pt}{9.6pt}\selectfont
    \vspace{-4pt}
    \begin{tabular}{@{}*{8}{>{\centering\arraybackslash}m{0.123\linewidth}}@{}}
      \minibox{HqEdit} & \minibox{InsV2V} & \minibox{InsVIE} & \minibox{LucyEdit} &
      \minibox{Qwen-Img-E} & \minibox{Se\~norita} & \minibox{Shape4Motion} & \minibox{ScaleVid (Ours)}
    \end{tabular}%
  }

  \vspace{-1pt}
  \begin{subfigure}{\linewidth}
    \centering
    \includegraphics[width=0.985\linewidth]{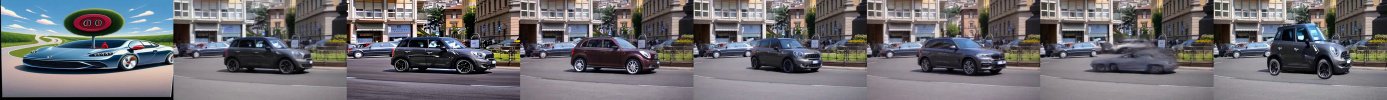}
  \end{subfigure}

  \caption{Comparison results on the Geometry Benchmark, Real-Background Benchmark, and Real-World Benchmark. For the {goblets}, ${s}=(1.80,1.50,0.80)$; for the {bulb}, ${s}=(0.75,0.75,0.75)$; and for the {car}, ${s}=(0.95,2.28,1.27)$.}
  
  \label{fig:video_compare}
\end{figure*}

\section{Experiment}







\subsection{Training Data}

\noindent\textbf{Main Model.}
Stage I is trained on a mixture of 1.8M filtered WebVid-10M videos, 900K synthetic videos generated by SelfForcing~\cite{huang2025self}, and 800K synthetic images produced by Stable Diffusion 3.5 Large~\cite{stabilityai_sd3_5_large_2024}. Stage II is finetuned on 180K Pexels videos.

\noindent\textbf{Deformer and Masker.}
We construct 1.5M paired videos from 300K object meshes by rendering five augmented pairs per mesh. Each pair contains the original and anisotropically scaled object under same configurations. Full training details are provided in the Supplementary.

\subsection{Benchmark Construction}

\begin{table*}[!t]
\centering
\small
\setlength{\tabcolsep}{1.5pt}
\renewcommand{\arraystretch}{0.95}
\newcommand{\cond}[1]{\scriptsize #1}

\begin{tabular}{@{}l|c|c|cccc||cc|ccccc@{}}
\toprule

\textbf{Method}
&
\textbf{Condition}
&
\textbf{Video}
&
\multicolumn{4}{c||}{\textbf{Background Preservation}}
&
\multicolumn{2}{c|}{\textbf{Scale Acc}}
&
\multicolumn{5}{c}{\textbf{Geometric Alignment}}
\\[-0.2ex]

&
&
\textbf{based}
&
MSE$\downarrow$
&
PSNR$\uparrow$
&
SSIM$\uparrow$
&
LPIPS$\downarrow$
&
IoU$\uparrow$
&
Area$\downarrow$
&
Yaw$\downarrow$
&
Pitch$\downarrow$
&
Dist$\downarrow$
&
Trans$\downarrow$
&
FIoU$\uparrow$
\\

\midrule


Ditto~\cite{bai2025scaling}
& \multirow{6}{*}{Text}
& \cmark
& 2028.51
& 17.79
& 0.645
& 0.312
& 0.442
& 0.759
& 0.495
& 0.383
& 0.403
& 0.296
& 0.797
\\

HqEdit~\cite{hui2025hqedit}
&
& \xmark
& 6819.12
& 10.33
& 0.456
& 0.482
& 0.261
& 1.375
& 0.287
& 0.713
& 0.650
& 0.600
& 0.555
\\

InsV2V~\cite{cheng2024consistent}
&
& \cmark
& 1291.92
& 18.59
& 0.703
& 0.259
& 0.408
& 0.709
& 0.268
& 0.624
& 0.647
& 0.524
& 0.627
\\

InsVIE~\cite{wu2025insvie}
&
& \cmark
& 2831.63
& 14.59
& 0.508
& 0.374
& 0.460
& 0.804
& 0.266
& 0.515
& 0.453
& 0.309
& 0.769
\\

LucyEdit~\cite{decartai2025lucyedit}
&
& \cmark
& 905.49
& 23.32
& 0.847
& 0.400
& 0.435
& 1.102
& 0.292
& 0.498
& 0.489
& 0.379
& 0.783
\\

Qwen-Image-E~\cite{wu2025qwen}
&
& \xmark
& 331.18
& \underline{27.41}
& \underline{0.904}
& 0.075
& 0.434
& 0.668
& 0.288
& 0.518
& 0.423
& 0.294
& 0.786
\\

\cmidrule(lr){1-14}


Flux Kontext~\cite{blackforestlabs2025fluxkontext}
& Image
& \xmark
& 989.54
& 19.68
& 0.652
& 0.150
& 0.425
& 0.675
& 0.292
& 0.377
& \underline{0.269}
& \underline{0.226}
& 0.771
\\

\cmidrule(lr){1-14}


Flux Fill~\cite{flux}
& \multirow{2}{*}{2D Mask}
& \xmark
& 343.32
& 25.59
& 0.902
& \underline{0.072}
& 0.378
& 0.706
& 0.260
& 0.712
& 0.596
& 0.444
& 0.661
\\

Se\~norita~\cite{zi2025se}
&
& \cmark
& 408.55
& 24.14
& 0.858
& 0.131
& 0.432
& 0.675
& 0.278
& 0.778
& 0.491
& 0.389
& 0.751
\\

\cmidrule(lr){1-14}


DiffHandles~\cite{pandey2024diffusion}
& \multirow{3}{*}{Depth}
& \xmark
& 1599.56
& 18.27
& 0.678
& 0.263
& \underline{0.604}
& 0.674
& 0.491
& \underline{0.313}
& 0.273
& 0.294
& \underline{0.812}
\\

FreeFine~\cite{zhu2025training}
&
& \xmark
& 520.58
& 22.46
& 0.766
& 0.356
& 0.523
& 0.489
& 0.258
& 0.616
& 0.409
& 0.395
& 0.740
\\

GeoDiffuser~\cite{sajnani2025geodiffuser}
&
& \xmark
& 333.92
& 24.97
& 0.821
& 0.120
& 0.544
& \underline{0.350}
& 0.304
& 0.552
& 0.439
& 0.308
& 0.741
\\

\cmidrule(lr){1-14}


Shape4Motion~\cite{liu2025shape}
& \cond{Point Cloud}
& \cmark
& \underline{250.31}
& 26.64
& 0.877
& 0.086
& 0.351
& 0.710
& \underline{0.253}
& 0.578
& 0.480
& 0.356
& 0.797
\\

\midrule


\textbf{ScaleVid (Ours)}
& 3D Scale
& \cmark
& \textbf{161.23}
& \textbf{30.58}
& \textbf{0.920}
& \textbf{0.049}
& \textbf{0.804}
& \textbf{0.227}
& \textbf{0.237}
& \textbf{0.205}
& \textbf{0.210}
& \textbf{0.178}
& \textbf{0.836}
\\

\bottomrule
\end{tabular}

\caption{
Quantitative evaluation of background preservation on the
Real-Background Benchmark, and scale accuracy and geometric
alignment on the Geometry Benchmark. Best and second-best results are
{bold} and {underlined}, respectively.
}
\label{tab:scale_geometry_background}
\end{table*}

\begin{table*}[t]
\centering
\small
\setlength{\tabcolsep}{2.2pt}
\renewcommand{\arraystretch}{0.90}

\begin{tabular}{@{}l|cccccc||cccc|cccc@{}}
\toprule

\textbf{Method}
&
\multicolumn{6}{c||}{\textbf{Foreground Fidelity}}
&
\multicolumn{4}{c|}{\textbf{Pexels}}
&
\multicolumn{4}{c}{\textbf{DAVIS}}
\\[-0.2ex]

&
MSE$\downarrow$
&
PSNR$\uparrow$
&
SSIM$\uparrow$
&
LPIPS$\downarrow$
&
DINO$\uparrow$
&
Dream$\uparrow$
&
Ewarp$\downarrow$
&
GPT$\uparrow$
&
Gemini$\uparrow$
&
User$\uparrow$
&
Ewarp$\downarrow$
&
GPT$\uparrow$
&
Gemini$\uparrow$
&
User$\uparrow$
\\

\midrule


Ditto
& 2362.73
& 16.06
& 0.782
& 0.220
& 0.536
& 0.821
& 0.441
& 0.14
& 0.06
& 0.08
& 2.156
& 0.40
& 0.60
& 0.09
\\

HqEdit
& 1575.52
& 17.96
& 0.797
& 0.216
& 0.506
& 0.827
& 9.475
& 0.14
& 0.34
& 0.03
& 9.128
& 0.14
& 0.08
& 0.05
\\

InsV2V
& 1008.17
& 20.50
& 0.812
& 0.205
& 0.641
& 0.874
& 0.344
& 2.46
& 3.64
& 0.23
& 1.529
& 2.34
& 3.03
& 0.19
\\

InsVIE
& 1724.08
& 17.47
& 0.793
& 0.216
& 0.556
& 0.831
& 0.329
& 1.78
& 2.72
& 0.25
& 2.772
& 1.57
& 1.62
& 0.10
\\

LucyEdit
& 1751.22
& 19.12
& 0.801
& 0.190
& 0.645
& 0.870
& 0.155
& 3.08
& 3.22
& 0.32
& 1.329
& 2.34
& 2.53
& 0.25
\\

Qwen-Image-E
& 666.06
& 22.41
& 0.825
& 0.204
& \underline{0.763}
& \underline{0.910}
& 1.799
& 2.22
& 2.60
& 0.13
& 2.204
& \underline{3.22}
& \underline{3.43}
& 0.14
\\

\cmidrule(lr){1-15}


Flux Kontext
& 653.20
& 22.59
& 0.825
& 0.207
& 0.761
& 0.907
& 0.387
& \underline{3.12}
& 3.06
& 0.30
& 1.485
& 2.01
& 2.62
& \underline{0.30}
\\

\cmidrule(lr){1-15}


Flux Fill
& 898.85
& 21.63
& 0.816
& 0.217
& 0.561
& 0.843
& 2.307
& 1.46
& 1.74
& 0.06
& 2.064
& 1.27
& 2.06
& 0.05
\\

Se\~norita
& 988.02
& 21.41
& 0.825
& 0.218
& 0.577
& 0.855
& \textbf{0.097}
& 0.94
& 1.02
& 0.08
& \textbf{0.886}
& 2.01
& 2.05
& 0.23
\\

\cmidrule(lr){1-15}


DiffHandles
& \underline{387.18}
& \underline{25.09}
& \underline{0.852}
& \underline{0.154}
& 0.726
& 0.882
& 4.586
& 0.56
& 0.66
& 0.09
& 6.316
& 0.63
& 0.80
& 0.06
\\

FreeFine
& 688.98
& 22.72
& 0.812
& 0.180
& 0.611
& 0.862
& 1.566
& 1.38
& 1.46
& 0.04
& 2.897
& 1.18
& 1.62
& 0.02
\\

GeoDiffuser
& 581.98
& 23.03
& 0.843
& 0.170
& 0.650
& 0.860
& 1.738
& 0.36
& 0.46
& 0.08
& 2.528
& 1.10
& 1.02
& 0.03
\\

\cmidrule(lr){1-15}


Shape4Motion
& 540.86
& 22.83
& 0.846
& 0.201
& 0.631
& 0.861
& \underline{0.122}
& 1.78
& 2.04
& \underline{0.37}
& \underline{0.943}
& 1.24
& 0.45
& 0.17
\\

\midrule


\textbf{ScaleVid (Ours)}
& \textbf{329.73}
& \textbf{25.33}
& \textbf{0.867}
& \textbf{0.133}
& \textbf{0.850}
& \textbf{0.934}
& 0.133
& \textbf{3.74}
& \textbf{4.22}
& \textbf{0.81}
& 1.102
& \textbf{4.14}
& \textbf{4.41}
& \textbf{0.80}
\\

\bottomrule
\end{tabular}

\caption{
Quantitative evaluation of foreground fidelity on the Geometry
Benchmark and real-world editing performance on Pexels and DAVIS.
Ewarp is reported in the range of $1\times10^{-3}$. Best and
second-best results are {bold} and {underlined},
respectively.
}
\label{tab:foreground_video_results}

\end{table*}

We collect 48 high-quality meshes from Poly Haven~\cite{polyhaven} and render each under manually configured settings. All benchmark meshes are excluded from our training set. Scaling factors $\mathbf{s}=(s_x,s_y,s_z)\in(0.3,3.0)^3$ are randomly sampled and applied to form strictly aligned video pairs.

We establish three complementary benchmarks. \textbf{(1) Geometry Benchmark.}
Objects are rendered on a uniform gray canvas to isolate geometric accuracy and foreground fidelity. For each object, we generate four synchronized outputs: the original video $V_{\text{ori}}$, the original mask $M_{\text{ori}}$, the scaled video $V_{\text{scl}}$, and the scaled mask $M_{\text{scl}}$, resulting in 48 video groups. \textbf{(2) Real-Background Benchmark.} Meshes are composited into real-world background videos collected from the Pexels dataset, yielding another 48 video groups. \textbf{(3) Real-World Benchmark.} To assess overall performance in realistic scenarios, we evaluate on two real-world datasets, including 50 videos from the Pexels~\cite{pexels} dataset and 91 videos with fast-motion from the DAVIS dataset~\cite{perazzi2016benchmark}. Since paired 3D-scaled ground truth is unavailable for real videos, real-world evaluation focuses on perceptual quality and temporal consistency.

\subsection{Evaluation Metrics}

To evaluate controllable object scaling in videos, we adopt four categories of metrics that measure editing quality. 

\noindent\textbf{Background and Foreground Evaluation.}
We evaluate the background and foreground regions, separated by the ground-truth mask $M_{\text{gt}}$, using MSE, PSNR, SSIM, and LPIPS~\cite{zhang2018unreasonable}. Foreground fidelity is further assessed by DINOv2~\cite{oquab2023dinov2} and DreamSim~\cite{fu2023dreamsim} for identity and perceptual similarities.

\noindent\textbf{Scale Accuracy and Geometric Alignment.}
We first evaluate the mask of model outputs and measure 2D mask accuracy using mask IoU and area error:
$$
\text{IoU}=\frac{|M_{\rm pred}\cap M_{\rm gt}|}{|M_{\rm pred}\cup M_{\rm gt}|},\,\text{Area} =
\frac{\left|\,|M_{\text{pred}}| - |M_{\text{gt}}|\,\right|}
{|M_{\text{gt}}|},
$$
where $M_{\text{pred}}$ is the predicted foreground mask by SAM2.

To further assess 3D geometric alignment, we adopt a mesh-based fitting evaluation inspired by Ctrl\&Shift~\cite{ruan2026ctrl}. Given the ground-truth scaled mesh and the predicted foreground mask, we optimize the camera and pose parameters $  (y, p, d, t_x, t_y)  $ — corresponding to yaw angle, pitch angle, camera distance, and image-plane translations — to best align the projected mesh silhouette with the predicted mask. Let $  (y^*, p^*, d^*, t_x^*, t_y^*)  $ denote the corresponding ground-truth values. We then report the following errors:
\begin{gather*}
\text{Yaw} = {|y - y^*|}, \,
\text{Pitch} = {|p - p^*|}, \,
\text{Dist} = {|d - d^*|}/d^*, \\
\text{Trans} = \sqrt{(t_x - t_x^*)^2 + (t_y - t_y^*)^2},
\end{gather*}
along with the silhouette fitting IoU (FIoU) between the rendered mesh silhouette and the predicted mask. These metrics collectively quantify how well the edited object preserves geometric alignment with the underlying 3D structure.

\noindent\textbf{Temporal and Global Perceptual Quality.}
We evaluate temporal consistency using Ewarp~\cite{lai2018learning}.
For global perceptual quality, GPT-5~\cite{openai_gpt5_2025} and
Gemini-2.5 Pro~\cite{comanici2025gemini} score the Real-World Benchmark based on scale accuracy, background preservation,
transformation correctness, structure consistency, and appearance
preservation. We further conduct a human preference study with
29 participants using 10 multiple-choice questions. Details are provided in the Supplementary.

\subsection{Qualitative Results}
As shown in Fig.~\ref{fig:2d_3d}, planar 2D scaling cannot
recover out-of-frame regions or perspective changes, causing broken
airplane wings, misaligned bull legs, and implausible newly exposed
safe faces. For a fair comparison, the planar baseline uses the same
in-plane scaling factors $(s_x,s_y)$ as ScaleVid. Moreover, it ignores object-scene
interactions, leading to unnatural shadows and boundary transitions around the laptop. In contrast, ScaleVid produces geometry-consistent results by learning transformation priors from 3D supervision.

In Fig.~\ref{fig:video_compare}, image-based methods show noticeable flickering. Text-controlled methods struggle to accurately interpret specific scale values. HqEdit, Ditto, InsV2V, and InsVIE show limited ability to preserve the identity of the edited object. Señorita shows similar artifacts due to its Flux Fill
first-frame guidance. Additional visual results and videos
are provided in the Technical and Media Supplementary.

\subsection{Quantitative Results}
As shown in Table~\ref{tab:scale_geometry_background}, \abbr{} achieves
the best scale accuracy, geometric alignment, and background preservation. Although
textual prompts can explicitly specify numerical scaling factors, text-driven
editing methods often fail to translate these values into precise and
consistent geometric transformations. In contrast, geometry-aware methods achieve noticeably better scale accuracy and geometric alignment by explicitly incorporating geometric priors. However, training-free and depth-guided approaches still preserve fine-grained object structure less effectively than mesh-based methods.

Table~\ref{tab:foreground_video_results} further shows that our method
achieves the best foreground fidelity on all metrics, including pixel-level error and perceptual similarity. On the real-world Pexels and DAVIS
datasets, it obtains the highest GPT, Gemini, and user-preference scores,
demonstrating superior perceptual quality and overall realism, together with competitive Ewarp.

Inference cost, baseline details, extensive quantitative results and more ablations are provided in Supplementary.

\subsection{Ablation Studies}
\subsubsection{Effectiveness of Progressive Training}
Table~\ref{tab:ablation_scale_geometry} evaluates the geometric contribution of each training stage on the Geometry Benchmark. The Deformer performs competitively due to its closely matched rendered training distribution. Stage~II substantially improves over Stage~I, confirming that 3D deformation guidance is essential for geometric control. Combining both stages achieves competitive overall performance, showing that planar pretraining provides a beneficial initialization for subsequent geometry-aware finetuning.

\begin{table}[t]
\centering
\small
\setlength{\tabcolsep}{2.8pt}
\renewcommand{\arraystretch}{0.9}

\begin{tabular}{l|cc|ccccc}
\toprule
\textbf{Method}
& IoU$\uparrow$
& Area$\downarrow$
& Yaw$\downarrow$
& Pitch$\downarrow$
& Dist$\downarrow$
& Trans$\downarrow$
& FIoU$\uparrow$ \\
\midrule

Deformer
& \underline{0.776}
& \textbf{0.222}
& \textbf{0.267}
& \underline{0.241}
& \underline{0.259}
& 0.230
& \underline{0.821} \\

Stage I
& 0.525
& 0.813
& 0.447
& 0.365
& 0.576
& 0.339
& 0.762 \\

Stage II
& 0.746
& 0.345
& 0.294
& 0.303
& 0.280
& \underline{0.223}
& 0.804 \\

Stage I+II
& \textbf{0.804}
& \underline{0.227}
& \textbf{0.237}
& \textbf{0.205}
& \textbf{0.210}
& \textbf{0.178}
& \textbf{0.836} \\

\bottomrule
\end{tabular}
\caption{Stage-wise ablation on scale \& geometric alignment.}

\label{tab:ablation_scale_geometry}
\end{table}

\subsubsection{Effectiveness of the Bidirectional Loss for Deformer}

As shown in Table~\ref{tab:dual}, moderate bidirectional loss consistently improves scale accuracy and geometric alignment over the unidirectional setting ($\lambda=0$). A moderate sampling probability performs best: $\lambda=0.2$ achieves the highest IoU and FIoU and the lowest errors across all geometric metrics, while larger values gradually degrade performance. This indicates that bidirectional supervision is beneficial, but overly frequent reverse constraints restrict deformation flexibility.

\begin{table}[t]
\centering
\small
\setlength{\tabcolsep}{4.4pt}
\renewcommand{\arraystretch}{0.9}
\begin{tabular}{l|cc|ccccc}
\toprule
\boldmath$\lambda$
& IoU$\uparrow$
& Area$\downarrow$
& Yaw$\downarrow$
& Pitch$\downarrow$
& Dist$\downarrow$
& Trans$\downarrow$
& FIoU$\uparrow$ \\
\midrule
1.0 & 0.616 & 0.579 & 0.475 & 0.371 & 0.347 & 0.234 & 0.710 \\
0.5 & \underline{0.716} & \underline{0.345} & \underline{0.412} & \underline{0.308} & \underline{0.269} & \underline{0.221} & \underline{0.801} \\
0.2 & \textbf{0.742} & \textbf{0.280} & \textbf{0.402} & \textbf{0.287} & \textbf{0.230} & \textbf{0.216} & \textbf{0.817} \\
0 & 0.702 & 0.420 & \underline{0.412} & 0.339 & 0.325 & 0.222 & 0.797 \\
\bottomrule
\end{tabular}

\caption{Ablation study on loss design of Deformer.}

\label{tab:dual}
\end{table}





\section{Conclusion}

We present \textbf{ScaleVid}, a geometry-aware framework for controllable
video object scaling without explicit 3D reconstruction at inference.
Through progressive pseudo-source construction with real-video targets,
ScaleVid decouples geometric transformation from video synthesis and
supports geometry-aware scaling directly in video domain. We also introduce
complementary benchmarks for evaluating geometric alignment, foreground
fidelity, background preservation, and overall quality. Extensive
experiments demonstrate strong and practical performance.

\bibliography{arxiv}


\newpage

\appendix


\section{Extensive Related Works}

Image inpainting has been extensively studied over the past years~\cite{ju2024brushnet, flux, powerpaint_zhuang2023task, rombach2022high, li2025rorem, liu2024prefpaintaligningimageinpainting, xie2025turbofilladaptingfewsteptexttoimage, bld, zhang2025ultrahighresolutionimageinpainting, podell2023sdxlimprovinglatentdiffusion}. Stable Diffusion Inpainting is a representative approach that concatenates masked latents with noisy latents to predict the original latent representation. BrushNet~\cite{ju2024brushnet} adopts a decomposed dual-branch diffusion architecture, leading to improved inpainting performance. PowerPaint~\cite{powerpaint_zhuang2023task} is a versatile inpainting framework that supports text-guided object editing across a wide range of tasks. Flux-Fill~\cite{flux} is built upon the Flux generative model and further extends its capabilities for image inpainting.

Video inpainting has advanced rapidly in recent years~\cite{zhang2024avid, zi2025minimax, bian2025videopainter, jiang2025vace, yang2025mtv, li2025diffueraser, wang2023videocomposer, propainter_zhou2023propainter, floed_gu2024advanced}. AVID~\cite{zhang2024avid} is the first method to introduce text-guided video inpainting, employing sparse control to modify object appearance and proposing a strategy for long video generation. COCOCO~\cite{zi2025cococo} improves consistency and controllability by incorporating damped global attention and enhanced textual cross-attention within motion blocks. MiniMax-Remover~\cite{zi2025minimax} focuses on object removal, leveraging a min–max optimization framework to prevent undesired object regeneration within masked regions. VideoPainter~\cite{bian2025videopainter} proposes a plug-and-play module that supports video inpainting of arbitrary length. Finally, VACE~\cite{jiang2025vace} presents an all-in-one video editing framework capable of performing video inpainting via a ControlNet-based architecture.

\section{Implementation Details}

\begin{algorithm}[t]
\caption{Progressive Training of the Main Model}
\label{alg:main_training}
\begin{algorithmic}[1]
\Require Video $V$, object mask $M$, Main Model $\theta$

\State Extract foreground $F=V\odot M$
\State Obtain the complete bg $B$ using Minimax Remover

\If{Stage I}
    \State Sample planar scaling factors $(s_x,s_y)$
    \State Apply bounding box $M^{\mathrm{box}}=\operatorname{BBox}(M)$
    \State Transform foreground $F$ to obtain
    $F^{\mathrm{src}}$
    \State Construct
    $B^{\mathrm{tgt}}
    =B\odot(1-M^{\mathrm{box}})$
    \State Set
    $\mathbf{c}
    =
    \{F^{\mathrm{src}},
    B^{\mathrm{tgt}},
    M^{\mathrm{box}}\}$
\Else
    \State Sample anisotropic scaling factors
    $\mathbf{s}=(s_x,s_y,s_z)$
    \State Generate pseudo source fg
    $F^{\mathrm{src}}
    =D_{\phi}(F,\mathbf{s})$
    \State Predict pseudo source mask
    $M^{\mathrm{src}}
    =G_{\psi}(F^{\mathrm{src}})$
    \State Construct
    $B^{\mathrm{src}}
    =B\odot(1-M^{\mathrm{src}})$
    \State \parbox[t]{0.88\linewidth}{
Generate target-aligned guidance:\par
\centering
$\displaystyle
F^{\mathrm{tgt}}
=
D_{\phi}\!\left(F^{\mathrm{src}},\mathbf{s}^{-1}\right)$
}
    \State Set
    $\mathbf{c}
    =
    \{F^{\mathrm{tgt}},
    B^{\mathrm{src}},
    M^{\mathrm{src}}\}$
\EndIf

\State Encode the original complete sample:
$z_V=\mathcal{E}_{\mathrm{VAE}}(V)$
\State Compute
$\mathcal{L}_{\mathrm{main}}
=
\mathcal{L}_{\mathrm{FM}}
(f_{\theta};z_V,\mathbf{c})$
\State Update $\theta$

\end{algorithmic}
\end{algorithm}

\subsection{Training Details}
\textbf{Training Details of Main Model}
During training, we randomly resize and downsample frames. In addition, we randomly drop the conditioning information with probability 0.1 by replacing the foreground video with an all-gray video, so that classifier-free guidance can be applied at inference time. The global batch size is 256 in Stage I and 64 in Stage II, with constant learning rate of 1e-5.

\noindent \textbf{Training Details of Deformer and Masker}
The training procedure for the Deformer and Masker are similar. We use DMD2 to distill the Masker from 20 sampling steps to 3 steps. We didn't perform random drop during the training stage of Masker. For Deformer, we randomly drop the $\textbf{s}$ with probability 0.1 by replacing it with a learnable negative embedding. Table~\ref{tab:refacade_detail} summarizes the key hyperparameters of Main Model, Deformer and Masker.

\begin{table}
\scriptsize
\setlength{\tabcolsep}{1.5pt}
\centering
\begin{tabular}{l|ccccc}
\toprule
\multirow{3}{*}{\raisebox{-2.0ex}{Config}} & \multicolumn{5}{c}{\textbf{Model}} \\
\cmidrule(lr){2-6}
 & \multicolumn{2}{c}{Main Model} & \textbf{Deformer} & \multicolumn{2}{c}{\textbf{Masker}} \\
\cmidrule(lr){2-3}\cmidrule(lr){4-4}\cmidrule(lr){5-6}
 & Stage I & Stage II & Stage I & Stage I & Distill \\
\midrule
Batch Size / GPU  & 4 & 2 & 4 & 4 & 2 \\
Accumulation Step  & \multicolumn{2}{c}{1} & 1 & \multicolumn{2}{c}{1} \\
Gradient Ckpt & \multicolumn{2}{c}{True} & True & \multicolumn{2}{c}{True} \\
Optimizer  & \multicolumn{2}{c}{AdamW} & AdamW & \multicolumn{2}{c}{AdamW} \\
Learning Rate  & \multicolumn{2}{c}{$1\times10^{-5}$} & $1\times10^{-5}$ & $1\times10^{-5}$ & $5\times10^{-6}$ \\
LR Schedule  & \multicolumn{2}{c}{Constant} & Constant & \multicolumn{2}{c}{Constant} \\
Timestep Sampling  & \multicolumn{2}{c}{Uniform} & Uniform & \multicolumn{2}{c}{Uniform} \\
Num GPUs & 64 & 32 & 32 & 32 & 8 \\
Training Steps & 24000 & 6000 & 23000 & 23000 & 50 \\
Training Hours & 128 & 71 & 100 & 100 & 1 \\
\midrule
Num Main Layers & \multicolumn{2}{c}{30} & 30 & \multicolumn{2}{c}{10} \\
Token Dimension  & \multicolumn{2}{c}{1536} & 1536 & \multicolumn{2}{c}{1536} \\
Parameters  & \multicolumn{2}{c}{1.2869B} & 1.4225B & \multicolumn{2}{c}{0.3871B} \\
Control Layer Indices & \multicolumn{2}{c}{0,4,8,12} & - & \multicolumn{2}{c}{-} \\
Pre-trained Model & \multicolumn{2}{c}{Minimax Remover} & Wan2.1-1.3B & \multicolumn{2}{c}{Wan2.1-1.3B} \\
\midrule
Sample Steps & \multicolumn{2}{c}{20} & 20 & 20 & 3 \\
Sampler & \multicolumn{2}{c}{Flow Euler} & Flow Euler & \multicolumn{2}{c}{Flow Euler} \\
Input Resolution(s) 
  & \multicolumn{2}{c}{Multi-resolution} 
  & Multi-resolution 
  & \multicolumn{2}{c}{Multi-resolution} \\
\bottomrule
\end{tabular}
\caption{Hyperparameter and training details of Main Model, Deformer and Masker.}
\label{tab:refacade_detail}
\end{table}

\subsection{Inference Details of \abbr{}}

\begin{figure}[t]
    \centering
    \includegraphics[width=0.985\linewidth]{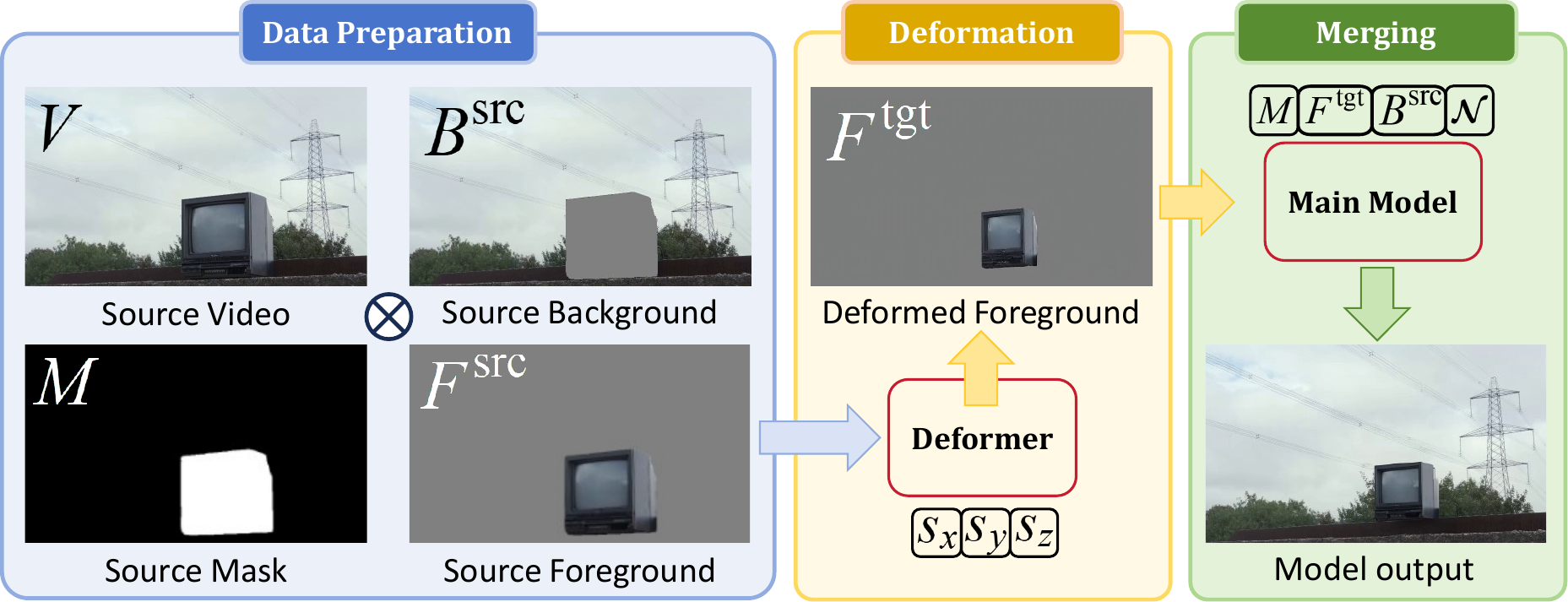}
    \caption{Inference pipeline of \abbr{}. }
    \label{fig:inference}
\end{figure}

Inference pipeline is shown in Fig.~\ref{fig:inference}. The user first provides a source video $V$ and obtains the source mask $M$ using SAM2, based on which the source foreground $F^{\rm src}$ and source background $B^{\rm src}$ are separated.  Following Minimax Remover~\cite{zi2025minimax}, we dilate the mask for 3-5 pixels to avoid boundary leakage. The source foreground $F^{\rm src}$ is then deformed by the Deformer under the guidance of the scaling factor $\mathbf{s}=(s_x, s_y, s_z)$, yielding the deformed foreground $F^{\rm tgt}$. Finally, $M$, $F^{\rm tgt}$, $B^{\rm src}$, and the noise $\mathcal{N}$ are concatenated and fed into the Main Model, and the output is decoded by a pretrained VAE decoder. Note that the Masker and object remover are not needed during inference stage.

\subsection{Benchmark Construction}
To construct our benchmark, we collect 48 high-quality object meshes from publicly available websites. The selected objects cover several semantic categories, including furniture, decorative objects, industrial items, appliances, natural objects, electronics, and tools. All videos are rendered using Kaolin~\cite{jatavallabhula2019kaolin}.

Each object is used to render one video group. Specifically, each group contains four aligned components: the original video, the original mask, the scaled foreground video, and the scaled foreground mask. The masks are directly provided by the renderer, and therefore serve as accurate pixel-level silhouettes for evaluation.

To cover diverse motion patterns, the 48 rendered groups are divided into three subsets, including 15 translation sequences, 17 rotation sequences, and 16 collision-deformation sequences. For each object, we randomly sample anisotropic 3D scaling factors $\mathbf{s}=(s_x,s_y,s_z)\in(0.3,3.0)^3$ and apply them to the object geometry.

\begin{figure}[t]
  \centering
  \renewcommand{\arraystretch}{1.0}
  \setlength{\tabcolsep}{0pt}
  {%
    \fontsize{8pt}{9.6pt}\selectfont
    \begin{tabular}{@{}*{4}{>{\centering\arraybackslash}m{0.25\linewidth}}@{}}
      \minibox{Original Video} & \minibox{Original Mask} & \minibox{Scaled Video} & \minibox{Scaled Mask} \\
    \end{tabular}%
  }

  \vspace{-1pt}

    \centering
    \includegraphics[width=0.985\linewidth]{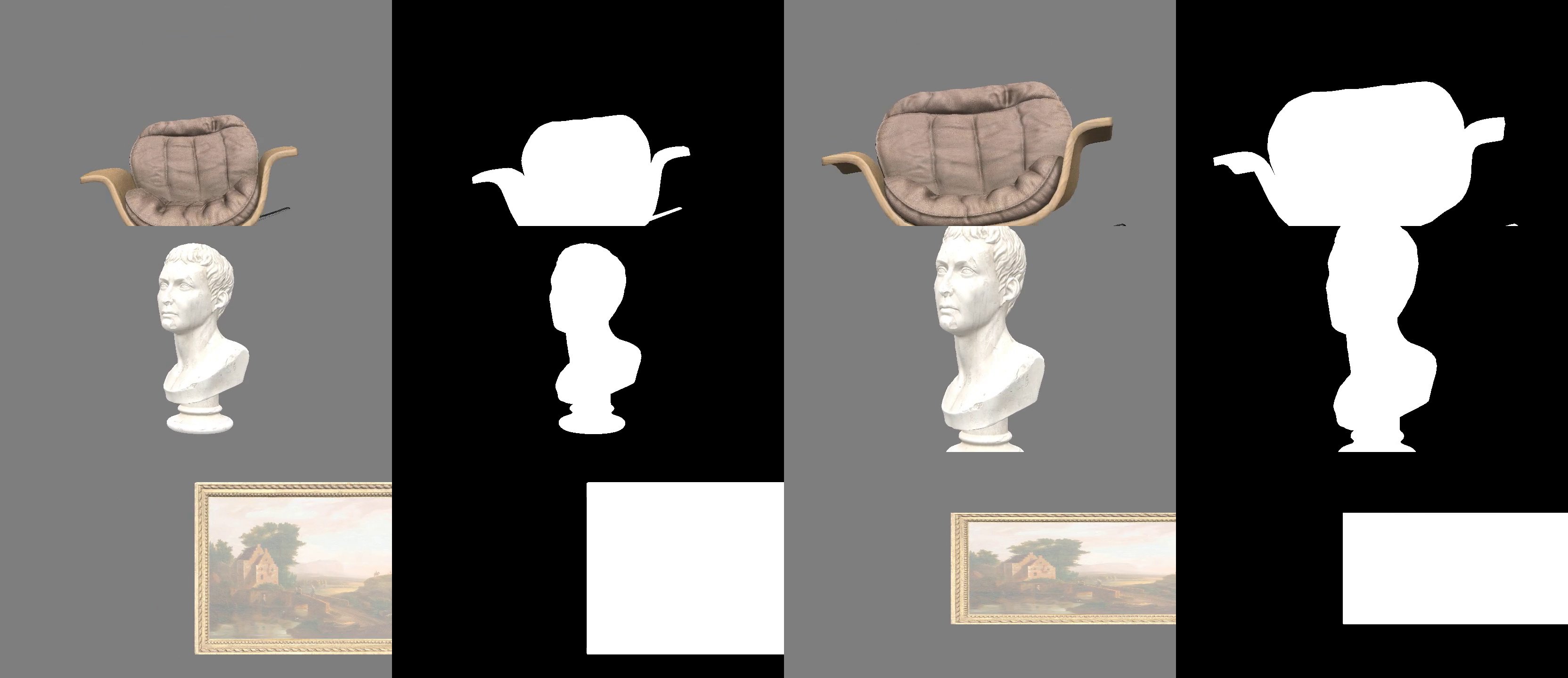}

  \caption{Visualization of our Geometry Benchmark.}
  \label{fig:image_video_compare}
\end{figure}

To ensure strict pairwise comparability, the original video and the scaled foreground video in each group are rendered with exactly the same motion trajectory, lighting setup, and camera parameters. Therefore, the only controlled variation comes from the applied 3D scaling, which enables reliable quantitative evaluation of both geometric alignment and visual fidelity. All rendered videos are generated at a resolution of $480\times832$ with 33 frames. The scaled foreground video contains only the transformed foreground object, without background content. This is shown in Fig.~\ref{fig:image_video_compare}.

\begin{figure}[t]
  \centering
  \renewcommand{\arraystretch}{1.0}
  \setlength{\tabcolsep}{0pt}
  {%
    \fontsize{8pt}{9.6pt}\selectfont
    \begin{tabular}{@{}*{5}{>{\centering\arraybackslash}m{0.20\linewidth}}@{}}
      \minibox{Background} & \minibox{Src Vid} & \minibox{Src Mask} & \minibox{Scaled Vid} & \minibox{Scaled Mask} \\
    \end{tabular}%
  }

  \vspace{-1pt}

    \centering
    \includegraphics[width=\linewidth]{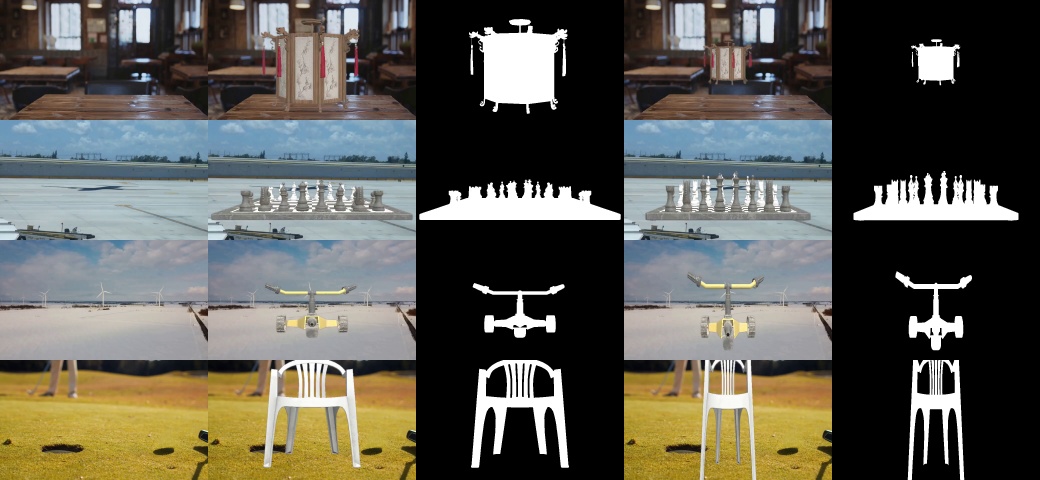}

  \caption{Visualization of our Real-Background Benchmark.}
  \label{fig:bg}
\end{figure}

Since the background in our Geometry Benchmark is uniformly gray, the evaluation of background preservation may appear relatively simplified. 
To address this concern, we further build an additional benchmark by replacing the gray background in the original synthetic pipeline with realistic open-scene video backgrounds, such as beaches and other spacious natural environments, as shown in Fig.~\ref{fig:bg}. 
We then render the foreground mesh onto these backgrounds to construct paired data under the same controllable setup. 
Compared with the original benchmark, this new benchmark introduces richer scene content and more realistic background variations, thereby providing a more convincing evaluation of background preservation.

\begin{figure}[t]
  \centering
  \renewcommand{\arraystretch}{1.0}
  \setlength{\tabcolsep}{0pt}

  {%
    \fontsize{8pt}{9.6pt}\selectfont
    \begin{tabular}{@{}*{4}{>{\centering\arraybackslash}m{0.25\linewidth}}@{}}
      \minibox{Original Video}
      &
      \minibox{$\mathbf{s}=(1,1,1.5)$}
      &
      \minibox{$\mathbf{s}=(1,1.5,1)$}
      &
      \minibox{$\mathbf{s}=(1.5,1,1)$}
    \end{tabular}%
  }

  \vspace{-1pt}

  \includegraphics[width=\linewidth]{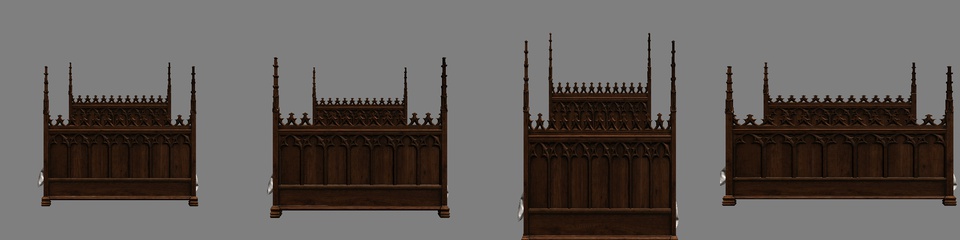}

  \vspace{2pt}

  {%
    \fontsize{8pt}{9.6pt}\selectfont
    \begin{tabular}{@{}*{4}{>{\centering\arraybackslash}m{0.25\linewidth}}@{}}
      \minibox{Original Video}
      &
      \minibox{$\mathbf{s}=(1,1,1.5)$}
      &
      \minibox{$\mathbf{s}=(1,1.5,1)$}
      &
      \minibox{$\mathbf{s}=(1.5,1,1)$}
    \end{tabular}%
  }

  \vspace{-1pt}

  \includegraphics[width=\linewidth]{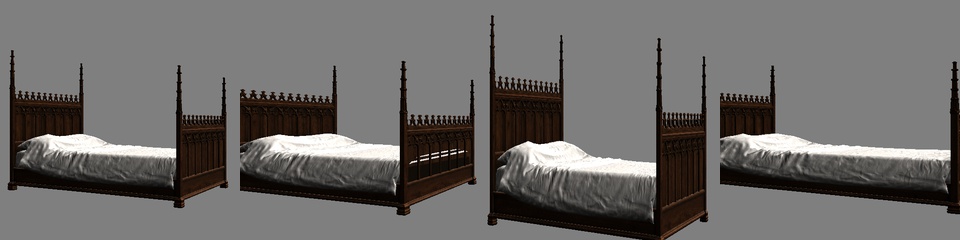}

  \caption{
  Pose-consistent axis-wise controllability of the constructed
  geometry benchmark.
  Top: axis-wise scaling of the original mesh.
  Bottom: the same scaling operations after rotating the original
  mesh by $60^\circ$ around the $y$-axis.
  In each case, one canonical axis is independently enlarged by
  $1.5\times$, while the remaining two axes are unchanged.
  }
  \label{fig:axis_control}
\end{figure}

To verify the controllability of the constructed geometric transformations,
we visualize axis-wise scaling examples in Fig.~\ref{fig:axis_control}.
Starting from the original object, we independently scale one canonical
axis by $1.5\times$ while keeping the other two scale factors fixed to
one. The results demonstrate that the proposed canonical alignment
produces disentangled transformations along different object-centric
directions.

\subsection{Inference Details of Baselines}
\begin{tcolorbox}[colback=white,colframe=black!75!white,title=Prompt Template for Text-guided Methods,fonttitle=\bfseries,breakable]
\small
Rescale the \{object\} by $\{s_x\}$ times in width, $\{s_y\}$ times in height, and $\{s_z\}$ times in depth. Keep background unchanged.
\end{tcolorbox}

\noindent\textbf{Implementation Details of DiffusionHandles.} This method is depth-guided, we use $\textbf{s}$ to edit the depth map. Pretrained MiDaS~\cite{birkl2023midas} and SD2-Depth model are used for inference. We use default configurations: $\text{bg\_weight}=1.25$, $\text{fg\_weight}=1.5$, and perform 50 inference steps on $512\times 512$ images frame by frame.

\noindent\textbf{Implementation Details of Ditto.}
We use a pretrained LoRA with Wan2.1-VACE-14B. Inference is performed at a resolution of \(480 \times 832\) with 33 frames, conditioned on the instructive prompt, while keeping all other settings at their default configuration. The generated videos are resized back to the original resolution.

\noindent\textbf{Implementation Details of FreeFine.} 
We use pretrained SV3D~\cite{voleti2024sv3d}, SD v1.5 and Depth Anything model~\cite{yang2024depth} for inference. We set the translation and rotation as zero, and use $\textbf{s}$ for inference. Moreover, we set CFG scale as 3.5, performing 50 denoising steps on $512\times 512$ images frame by frame.

\noindent\textbf{Implementation Details of GeoDiffuser.} 
We use the pretrained MiDaS~\cite{birkl2023midas} and SD v1.5 for inference. We use $\textbf{s}$ for object rescaling task. We set CFG scale as 3.0, with 50 DDIM denoising steps. Other configurations remain default value.

\noindent\textbf{Implementation Details of InsV2V.}
We use the pretrained InsV2V checkpoint for evaluation. Input videos are resized to a resolution of \(384 \times 384\) and truncated to 33 frames. We set \texttt{text\_cfg} to 7.5 and \texttt{img\_cfg} to 1.2, while keeping all other parameters at their default settings. The generated videos are resized back to the original resolution.

\noindent\textbf{Implementation Details of InsVIE.}
We use the pretrained InsVIE checkpoint together with the CogVideoX-2B~\cite{yang2024cogvideox} base model. Input videos are resized to a resolution of \(480 \times 720\) and truncated to 49 frames. The model is conditioned on the instructive prompt, with a negative prompt of ``bad quality'', while all other parameters follow the default configuration.

\noindent\textbf{Implementation Details of LucyEdit.}
We use the Lucy-Edit-1.1-Dev model for evaluation. Input videos are resized to a resolution of \(480 \times 832\) and truncated to 33 frames. We set CFG to 5.0 and condition the model on the instructive prompt, using empty prompt as the negative prompt. All other settings follow the default configuration. Finally, the generated videos are resized back to the original resolution.

\noindent\textbf{Implementation Details of Se\~norita.}
We use the pretrained Se\~norita checkpoint with the CogVideoX-5b-I2V base model for evaluation. Input videos are resized to \(448 \times 768\) and truncated to 33 frames. The first frame is edited by Flux-Fill and then used as the starting frame for generation. We set CFG to 4.0 and perform 50 denoising steps, conditioning the model on the instructive prompt, while keeping all other parameters at their default settings.

\noindent\textbf{Implementation Details of Flux-Fill.}
We use the FLUX.1-Fill-dev model and perform inference at the original image resolution.
The pipeline takes the source image, its corresponding mask and a descriptive prompt as input.
We set the CFG scale to 30.0 and use 50 steps for inference.

\noindent\textbf{Implementation Details of Flux-Kontext.}
We use the FLUX.1-Kontext-dev model conditioned on the instructive prompt and reference image.
Inference is performed at the original image resolution with 28 denoising steps and CFG is set to 3.0.

\noindent\textbf{Implementation Details of HQ-Edit.}
We use the released pretrained checkpoint of HQ-Edit.
Input images are resized to resolution $512 \times 512$ before inference. We set the CFG to 7.0, perform 30 denoising steps and set \texttt{image\_guidance\_scale} to 1.5 while conditioning on the instructive prompt.
Finally, the generated images are resized back to the original resolution for comparison.

\noindent\textbf{Implementation Details of Qwen-Image-Edit.}
For Qwen-Image-Edit, we perform inference at the original resolution of each input image. 
We run 50 denoising steps with \texttt{true\_cfg\_scale} set to 4.0, conditioning the model on the instruction prompt.

\noindent\textbf{Implementation Details of Shape for Motion.}
We use their pretrained weight and SV3D for inference. We follow the six-step inference: (1) preprocess the frames of input video by extracting the depth and mask; (2) reconstructing the object; (3) test the optimized model and save the canonical mesh; (4) manually editing the point cloud guided by the same $\mathbf{s}$: following the official implementation, we import the reconstructed
canonical point cloud into Blender and apply the exact anisotropic
scaling factors $(s_x,s_y,s_z)$ using Blender's numerical scaling
operator; (5) propagate the editing from one frame to all other frames; (6) generative rendering.

\subsection{Evaluation Metrics Implementation}
\noindent\textbf{Implementation Details of Background Evaluation.} We first dilate original mask by 16 pixels to mitigate mask inaccuracy. We then compute the average MSE, PSNR, SSIM, and LPIPS over the remaining background region. For videos, these metrics are computed on a per-frame basis and then averaged over all frames of all videos.

\noindent\textbf{Implementation Details of Foreground Evaluation.} As discussed in Sec. 4.3 of the main text, we use DINO, LPIPS, and DreamSim for foreground evaluation. Specifically, we first extract the foreground regions from the videos. For DINO and DreamSim, we use their corresponding base models to extract features from both the generated videos and the ground truth, and then compute the cosine similarity between the two feature vectors, where a larger value indicates higher similarity in material, color, and structure.

\noindent\textbf{Implementation Details of LLM Evaluation.}
Given the source image or videos and the output of one method, we ask GPT-5~\citep{openai_gpt5_2025} and Gemini-2.5-Pro~\citep{comanici2025gemini} to assign a score. The instruction is as follows:

\begin{tcolorbox}[colback=white,colframe=black!75!white,title=Template for LLM Evaluation on Geometry Benchmark,fonttitle=\bfseries,breakable]
\small
    You will receive four images:
    
    A, B: the first and middle frames of the ground-truth video.
    
    C, D: the first and middle frames of the generated video.
    
    Please evaluate generated against ground truth from the following four aspects:
    
    1) Object scale accuracy: whether the target object's size matches the ground truth.
    
    2) Background preservation: whether the background is preserved correctly.
    
    3) Transformation correctness: whether the object is scaled around its center, without unwanted translation or rotation.
    
    4) Structure and appearance preservation: whether the object's structure and visual appearance are preserved.
    
    Scoring rule:
    
    - Give 1 point for each satisfied aspect.
    
    - Total score must be an integer from 0 to 4.
    
    - Return ONLY the integer score.
\end{tcolorbox}

\begin{tcolorbox}[colback=white,colframe=black!75!white,title=Template for LLM Evaluation on Real-World Benchmark,fonttitle=\bfseries,breakable]
\small
You are asked to evaluate an object scaling result.

The target object is: $\{object\_name\}$.

The task is to scale the specified object according to the given 3D scaling factors:
sx = ${s_x}$, sy = ${s_y}$, sz = ${s_z}$.

sx, sy, and sz control object-centric width/length, height/upright, and depth/thickness, respectively.

You will receive four images:
A, B: the input images before editing.
C, D: the edited images after object scaling.

Please compare the input image and the edited image, and rate the edited result from 0 to 5 on the following aspects:

1. Scale Accuracy: whether the target object is scaled according to the instruction.
2. Transformation Correctness: whether the scaling is centered on the object without undesired translation or rotation.
3. Background Preservation: whether the background remains unchanged and free of artifacts.
4. Overall Quality: the overall realism and editing quality.
5. Appearance Preservation: whether the appearance of original object is preserved.

Scoring rule:
- If one aspect is satisfied, add one point.
- The total score must be an integer from 0 to 5.
- Please only return the integer score.
- Do not return any explanation, JSON, punctuation, or extra text.
\end{tcolorbox}

\noindent\textbf{Implementation Details of Camera Pose Estimation.}
To estimate camera pose from the predicted foreground mask, we adopt a two-stage mesh-fitting procedure. First, the ground-truth mesh is centered at its centroid and uniformly normalized to a canonical scale. We use a perspective camera with fixed vertical field of view $60^\circ$ and optimize five parameters $(y,p,d,t_x,t_y)$, corresponding to yaw, pitch, camera distance, and image-plane translations. Here, the yaw angle $y \in (-\pi,\pi]$ describes the horizontal rotation of the object, the pitch angle $p \in [-\frac{\pi}{2},\frac{\pi}{2}]$ describes the vertical rotation, $d$ represents the camera-object distance, and $(t_x,t_y)\in[-1,1]^2$ denotes the normalized image-plane translation.

For each frame, the predicted mask is binarized and resized to two resolutions, i.e., $240\times 416$ and $480\times 832$. The low-resolution mask is used for coarse initialization, while the high-resolution mask is used for final refinement. We compute the centroid of the low-resolution mask and convert it into normalized image-plane coordinates, which provides an initialization for $(t_x,t_y)$.

In the coarse stage, we randomly sample $M=256$ camera candidates, with yaw sampled from $(-\pi,\pi]$, pitch from $[-\pi/2,\pi/2]$, distance from $[0.5,4.0]$, and translations sampled around the centroid-based initialization. Silhouette rendering is then performed, and candidates with fitting IoU larger than 0.35 are retained. This process is repeated until 64 valid candidates are collected or the maximum number of sampling rounds is reached.

In the fine stage, the retained candidates are further optimized by Adam for 150 iterations with learning rate 0.05. The optimization objective is the silhouette IoU loss between the rendered mesh mask and the predicted foreground mask. We early-stop the optimization once the fitting IoU exceeds 0.96. In addition, when processing videos frame by frame, we use the optimized result from the previous frame as initialization for the current frame whenever available, which improves temporal stability and reduces optimization difficulty. We report the optimized pose parameters and the final silhouette fitting IoU for evaluation.

\begin{figure}[t]
    \centering
    \includegraphics[width=0.985\linewidth]{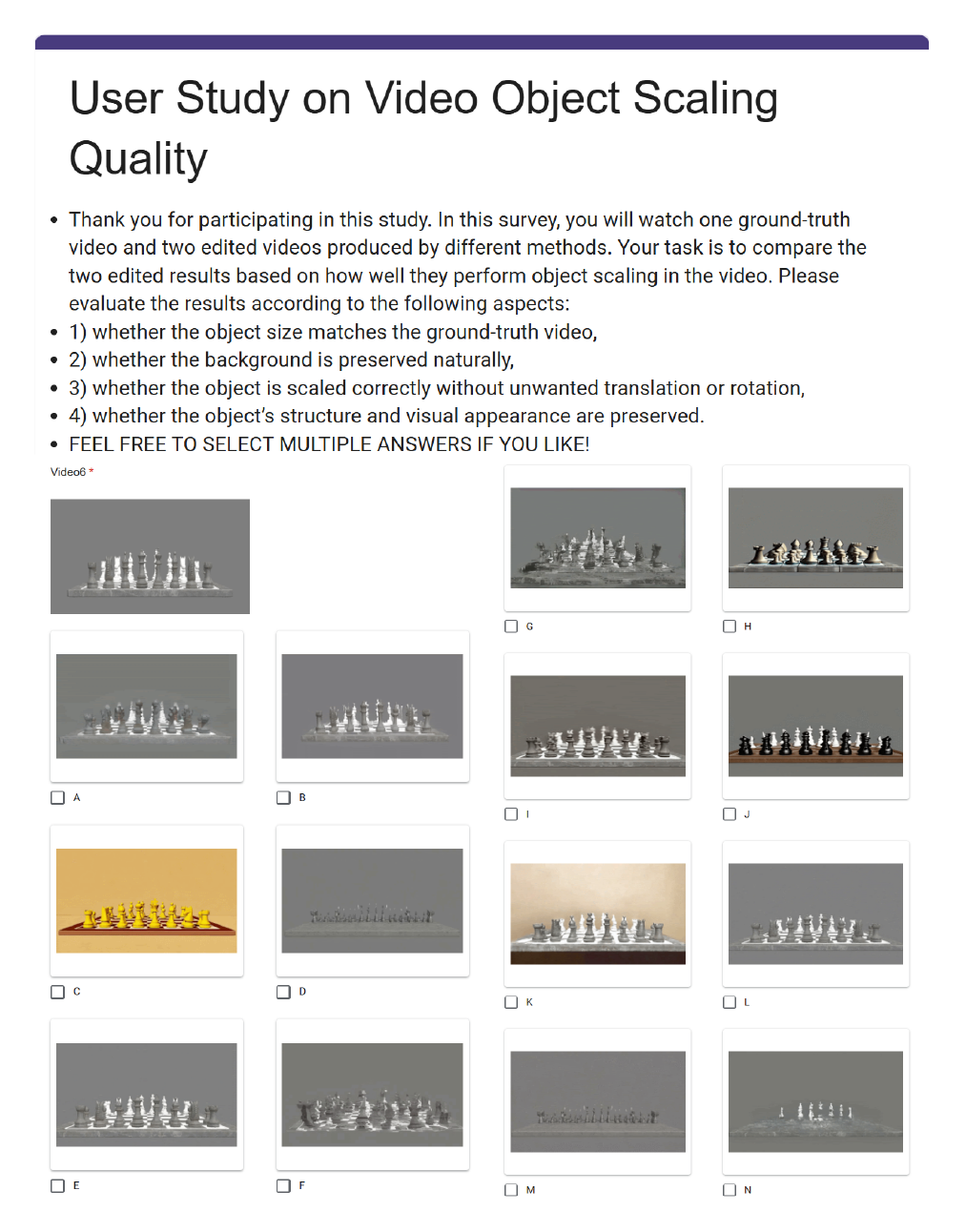}
    \caption{User study on Geometry Benchmark.}
    \label{fig:questionnaire}
\end{figure}

\begin{figure}[t]
    \centering
    \includegraphics[width=0.985\linewidth]{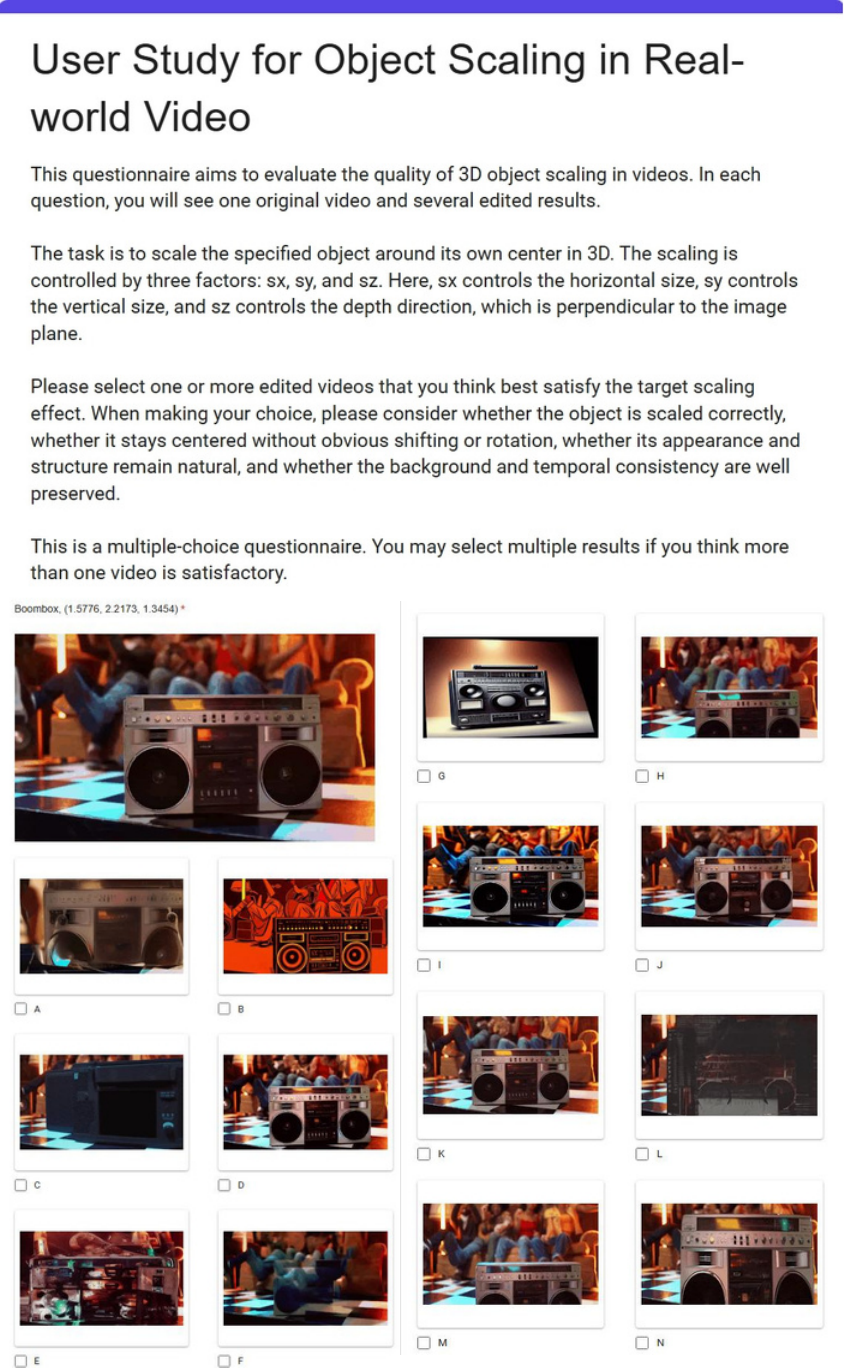}
    \caption{User study on the Real-World Benchmark.}
    \label{fig:questionnaire2}
\end{figure}

\noindent\textbf{Implementation Details of User Preference.}
To evaluate human preferences over different editing methods, we design a questionnaire that presents the
results of various image and video editing approaches. Participants are asked to assess the outputs from
multiple aspects and select all options they find satisfactory. The questionnaire instructions are as in Figures~\ref{fig:questionnaire}, \ref{fig:questionnaire2}.

For a method $m$, its user preference score is computed as
\begin{equation}
\operatorname{UserPref}(m)
=
\frac{N_m}{N_{\mathrm{q}}\times N_{\mathrm{item}}},
\end{equation}
where $N_m$ denotes the total number of times method $m$ is selected,
$N_{\mathrm{q}}$ is the number of valid questionnaires, and
$N_{\mathrm{item}}$ is the number of questions in each questionnaire.

\section{Extensive Quantitative Results}
\subsection{Consistency between LLM and Human Evaluation}

Following~\citep{zi2025minimax}, we compare the evaluation results
of human evaluators and LLM evaluators, including GPT-5 and Gemini-2.5
Pro. We randomly sampled 90 method outputs from the Geometry Benchmark
and asked 10 human evaluators to score them using the same four criteria.
To further investigate the consistency between LLM-based and human
evaluation, we compute the mean absolute error (MAE) and Spearman rank
correlation between LLM scores and human scores.

As shown in Table~\ref{tab:llm_human_average}, both GPT-5 and
Gemini-2.5 Pro demonstrate strong agreement with human evaluation,
achieving Spearman correlations of 0.766 and 0.738, respectively.
Gemini-2.5 Pro achieves a closer score calibration with human
evaluators, obtaining a lower MAE of 0.278, while GPT-5 shows slightly
stronger ranking consistency and a more conservative scoring tendency.
These results indicate that LLM-based evaluation can provide
consistent assessments with human judgments for our video editing
task.
\begin{table}[t]
  \centering
  \small
  \begin{tabular}{l|ccc}
    \toprule
    Evaluator & Average Score & MAE$\downarrow$ & Spearman $\rho\uparrow$ \\
    \midrule
    Human & 3.09 & -- & -- \\
    GPT-5 & 2.86 & 0.344 & 0.766 \\
    Gemini-2.5 Pro & 3.01 & 0.278 & 0.738 \\
    \bottomrule
  \end{tabular}
  \caption{Consistency analysis between human and LLM-based
  evaluations over 90 samples. MAE measures the score deviation
  from human judgments, while Spearman correlation measures the
  ranking consistency.}
  \label{tab:llm_human_average}
\end{table}

\subsection{Inference Cost} 
We report the wall-clock inference time and peak GPU memory usage under the same setting: all methods are evaluated on videos of resolution \textbf{512$\times$512} with \textbf{33} frames, using their default inference steps on a single A800 GPU. The results are summarized in Table~\ref{tab:infer_video}. Image-based methods require performing VAE encoding and decoding for each frame of the video, which introduces substantial computational overhead. In addition, DiffusionHandles requires two complete forward passes for each edited image, further increasing the inference cost. GeoDiffuser further incurs per-frame optimization cost. Shape-for-Motion requires estimating camera parameters and relies on manual external mesh editing, which also leads to considerable computational overhead.

\begin{table}[t]
\centering
\small
\setlength{\tabcolsep}{1.2pt}
\begin{tabular}{lcccc}
\toprule
Method 
& DiffHandles & Ditto & Flux-Fill & Flux-Ktx \\
\midrule
Time$\downarrow$
& 33m33s & 6m29s & 4m41s & 15m40s \\
GPU (GB)$\downarrow$
& 13.08 & 42.97 & 34.41 & 35.85 \\
Steps
& 50 & 50 & 50 & 28 \\
\midrule
Method 
& FreeFine & GeoDiffuser & HqEdit & InsV2V \\
\midrule
Time$\downarrow$
& 6m12s & 23min58s & 48.51s & 2m29s \\
GPU (GB)$\downarrow$
& 11.74 & 23.62 & \textbf{3.48} & 16.14 \\
Steps
& 50 & 50 & 30 & 20 \\
\midrule
Method 
& InsVIE & LucyEdit & Qwen-Img-E & Se\~norita \\
\midrule
Time$\downarrow$
& 1m33s & \underline{22.17s} & 41m29s & 1m58s \\
GPU (GB)$\downarrow$
& 31.05 & 27.66 & 59.67 & 30.04 \\
Steps
& 50 & 50 & 50 & 50 \\
\midrule
Method 
& Shape4Motion & \textbf{Ours(E2E)} & Deformer & Masker \\
\midrule
Time$\downarrow$
& 56m58s & \textbf{21.78s} & 6.18s & 0.16s \\
GPU (GB)$\downarrow$
& 23.06 & \underline{7.61} & 7.61 & 7.61 \\
Steps
& 6 & 20 & 20 & 3 \\
\bottomrule
\end{tabular}
\caption{Inference cost on videos.}
\label{tab:infer_video}
\end{table}

\subsection{Foreground Fidelity under Realistic Backgrounds}
Paired geometry-aware scaling targets are generally unavailable
for in-the-wild videos. To approximate realistic evaluation while
retaining accurate paired supervision, our Real-Background
Benchmark composites controllably rendered foregrounds with
background videos collected from real-world scenes. This hybrid
paired setting bridges the gap between fully controlled synthetic
evaluation and unpaired in-the-wild video evaluation. As shown in
Table~\ref{tab:foreground_fidelity}, ScaleVid achieves the best
results on five of the six metrics, including MSE, PSNR, SSIM,
DINO and DreamSim, while obtaining the second-best LPIPS.
These results demonstrate that the foreground-fidelity advantage
of ScaleVid is not limited to the uniform gray background of the
Geometry Benchmark, but remains consistent under realistic scene
textures, colors, and background complexity.
\begin{table}[t]
\centering
\small
\setlength{\tabcolsep}{1.0pt}
\renewcommand{\arraystretch}{0.95}

\begin{tabular}{@{}l|cccccc@{}}
\toprule
\textbf{Method}
&
MSE$\downarrow$
&
PSNR$\uparrow$
&
SSIM$\uparrow$
&
LPIPS$\downarrow$
&
DINO$\uparrow$
&
Dream$\uparrow$
\\
\midrule

DiffHandles
& \underline{660.55}
& \underline{22.90}
& 0.822
& 0.155
& 0.697
& \underline{0.885} \\

Ditto
& 2373.49
& 16.02
& 0.779
& 0.214
& 0.542
& 0.819 \\

Flux Fill
& 1221.53
& 19.90
& 0.815
& 0.178
& 0.560
& 0.834 \\

Flux Kontext
& 944.86
& 20.90
& 0.818
& 0.161
& \underline{0.718}
& 0.885 \\

FreeFine
& 846.85
& 21.21
& 0.803
& 0.156
& 0.652
& 0.872 \\

GeoDiffuser
& 758.17
& 22.23
& 0.848
& \textbf{0.136}
& 0.691
& 0.883 \\

HqEdit
& 1906.73
& 17.12
& 0.789
& 0.212
& 0.468
& 0.809 \\

InsV2V
& 969.61
& 20.34
& 0.820
& 0.176
& 0.670
& 0.871 \\

InsVIE
& 1584.23
& 18.28
& 0.795
& 0.188
& 0.610
& 0.847 \\

LucyEdit
& 1291.69
& 19.55
& 0.803
& 0.180
& 0.621
& 0.859 \\

Qwen-Image-E
& 983.23
& 20.58
& 0.820
& 0.164
& 0.713
& 0.884 \\

Se\~norita
& 1194.44
& 20.04
& \underline{0.823}
& 0.177
& 0.577
& 0.845 \\

Shape4Motion
& 1187.29
& 19.78
& 0.812
& 0.188
& 0.578
& 0.842 \\

\midrule

\textbf{ScaleVid (Ours)}
& \textbf{430.87}
& \textbf{24.98}
& \textbf{0.864}
& \underline{0.140}
& \textbf{0.834}
& \textbf{0.927} \\

\bottomrule
\end{tabular}
\caption{Quantitative evaluation of foreground fidelity on the Real-Background Benchmark.}
\label{tab:foreground_fidelity}
\end{table}

\subsection{Comparison with 2D Affine Scaling}
We further compare ScaleVid with a deterministic 2D affine-scaling baseline on the Geometry Benchmark. We do not report foreground-fidelity metrics for this comparison because the affine baseline directly transforms the ground-truth foreground pixels in the image plane, giving it privileged access to the original appearance and making pixel-level or perceptual foreground comparisons inherently unfair. Background-preservation metrics are also omitted, since the Geometry Benchmark uses a uniform gray canvas and is designed primarily to isolate geometric transformation quality rather than realistic background reconstruction. We therefore focus on scale accuracy and geometric alignment. As shown in Table~\ref{tab:affine_2d_comparison}, ScaleVid consistently outperforms 2D affine scaling, demonstrating that directly resizing foreground pixels cannot adequately reproduce the perspective-dependent shape changes and newly exposed surfaces induced by object-centric 3D scaling.
\begin{table}[t]
\centering
\small
\setlength{\tabcolsep}{3.0pt}
\renewcommand{\arraystretch}{1.08}
\begin{tabular}{l|ccccccc}
\toprule
Method
& IoU$\uparrow$
& Area$\downarrow$
& Yaw$\downarrow$
& Pitch$\downarrow$
& Dist$\downarrow$
& Trans$\downarrow$
& FIoU$\uparrow$ \\
\midrule
Affine2D
& 0.514
& 0.830
& 0.496
& 0.401
& 0.603
& 0.369
& 0.730 \\

ScaleVid
& 0.804
& 0.227
& 0.237
& 0.205
& 0.210
& 0.178
& 0.836 \\
\bottomrule
\end{tabular}
\caption{Quantitative comparison with planar affine scaling on the Geometry Benchmark.}
\label{tab:affine_2d_comparison}
\end{table}

\subsection{Geometry Consistency between Deformer Guidance and Main Model Output}

To examine whether the Main Model preserves the geometry encoded by
the Deformer guidance, we compare the foreground masks of the Deformer
condition, the final Main Model output, and the ground truth.
Let $M_{\rm def}$, $M_{\rm main}$, and $M_{\rm gt}$ denote the
foreground masks of the Deformer output, Main Model output, and
ground-truth target, respectively. Masks are extracted by SAM2.
For two masks $M_a$ and $M_b$, where $M_b$ is treated as the reference,
we compute
\begin{equation}
\operatorname{AreaErr}(M_a,M_b)
=
\frac{\left|\,|M_a|-|M_b|\,\right|}{|M_b|}.
\end{equation}

\begin{table}[t]
\centering
\small
\setlength{\tabcolsep}{5.5pt}
\renewcommand{\arraystretch}{1.05}
\begin{tabular}{ll|cc}
\toprule
Evaluated Mask & Reference Mask
& IoU$\uparrow$
& Area Error$\downarrow$ \\
\midrule
$M_{\rm def}$  & $M_{\rm gt}$  & 0.776 & 0.222 \\
$M_{\rm main}$ & $M_{\rm gt}$  & 0.804 & 0.227 \\
$M_{\rm main}$ & $M_{\rm def}$ & \textit{0.924} & \textit{0.169} \\
\bottomrule
\end{tabular}
\caption{Geometry consistency between the Deformer guidance and the
Main Model output on the Geometry Benchmark. The first two rows measure
target-geometry accuracy, while the last row directly measures how well
the Main Model preserves the geometry encoded by the Deformer condition.}
\label{tab:deformer_main_geometry_consistency}
\end{table}
As shown in Table~\ref{tab:deformer_main_geometry_consistency},
the Main Model output remains highly consistent with the Deformer
guidance, achieving an IoU of 0.924 and an area error of 0.169.
This high overlap indicates that the Main Model largely preserves the
geometry specified by its foreground condition, supporting our design
assumption that the condition geometry determines the desired output
geometry.

Meanwhile, the consistency is not expected to be perfect. As shown in
Fig.~\ref{fig:err_cor}, the Deformer may occasionally produce blurry or
imprecise foreground boundaries, whereas the Main Model can refine the
object structure and compensate for such deformation artifacts.
Consequently, the Main Model does not simply copy the Deformer guidance
pixel by pixel, but preserves its overall target geometry while improving
foreground quality and boundary accuracy. This is also reflected in the
comparison with the ground truth: the Main Model improves the mask IoU
from 0.776 to 0.804, while maintaining a comparable area error
(0.227 versus 0.222).

\subsection{Safe-Region Background Preservation}

For in-the-wild videos, the valid background region is
ambiguous because object enlargement may cover previously
visible background, whereas object shrinkage reveals regions
that are unobserved in the source video. Therefore, we do not
evaluate pixel fidelity in regions close to either the source or
edited object. Let $M_{\mathrm{src}}$ and
$M_{\mathrm{out}}^{(k)}$ denote the source-object mask and
the output-object mask of method $k$, respectively. We define
a shared conservative exclusion region as
\begin{equation}
M_{\mathrm{unsafe}}
=
\operatorname{Dilate}
\left(
M_{\mathrm{src}}
\cup
\bigcup_k M_{\mathrm{out}}^{(k)},
r
\right),
\end{equation}
where the dilation radius is set to $r=16$ pixels for all
methods. We then evaluate background fidelity only on
\begin{equation}
M_{\mathrm{safe}} = 1 - M_{\mathrm{unsafe}}.
\end{equation}
We report MSE, PSNR, SSIM, and LPIPS within this safe
region in Tab.~\ref{tab:davis_bg}. This measures unintended changes in the
far background, rather than the quality of newly revealed
background regions or local object--scene interactions.

\begin{table}[t]
\centering
\small
\setlength{\tabcolsep}{4pt}
\renewcommand{\arraystretch}{1.08}
\begin{tabular}{l|cccc}
\toprule
Method
& MSE$\downarrow$
& PSNR$\uparrow$
& SSIM$\uparrow$
& LPIPS$\downarrow$ \\
\midrule

DiffHandles
& 1340.13
& 18.24
& 0.598
& 0.242 \\

Ditto
& 1882.43
& 17.51
& 0.678
& 0.424 \\

Flux Fill
& 181.72
& 28.78
& \underline{0.874}
& \underline{0.060} \\

Flux Kontext
& 1147.84
& 18.84
& 0.488
& 0.133 \\

FreeFine
& 476.07
& 23.57
& 0.705
& 0.111 \\

GeoDiffuser
& 377.18
& 24.93
& 0.746
& 0.129 \\

HqEdit
& 7098.23
& 10.88
& 0.276
& 0.371 \\

InsV2V
& 2008.26
& 16.73
& 0.506
& 0.288 \\

InsVIE
& 3179.37
& 13.78
& 0.349
& 0.330 \\

LucyEdit
& 515.51
& 25.95
& 0.792
& 0.116 \\

Qwen-Image-E
& 345.02
& 29.12
& \textbf{0.876}
& \underline{0.060} \\

Se\~norita
& 255.74
& 26.34
& 0.795
& 0.138 \\

Shape4Motion
& \underline{86.10}
& \textbf{30.55}
& 0.863
& 0.074 \\

\midrule

\textbf{ScaleVid (Ours)}
& \textbf{79.11}
& \underline{30.37}
& 0.871
& \textbf{0.054} \\

\bottomrule
\end{tabular}
\caption{Quantitative comparison of safe-region background
preservation on the DAVIS subset of our Real-World Benchmark.
The best results
are \textbf{boldfaced}, and the second-best results are
\underline{underlined}.}
\label{tab:davis_bg}
\end{table}

\subsection{Comparison with FlowDrag}

FlowDrag is a geometry-aware image editing method based on
point-drag control. Although it does not natively support global
object-centric anisotropic scaling specified by continuous factors
$(s_x,s_y,s_z)$, we include it as an additional geometry-aware
baseline for completeness.

To adapt FlowDrag to our task, we derive drag constraints from the
target scaling transformation and apply the method frame by frame.
Since sparse local drag control cannot fully specify the global
shape changes induced by anisotropic 3D scaling, this comparison
should be regarded as a complementary reference rather than a
directly matched control setting.

As reported in Tab.~\ref{tab:flowdrag}, FlowDrag attains a relatively high fitting IoU, indicating that
sparse drag control can partially match the target silhouette.
However, its substantially worse mask IoU, area error, and camera-
pose deviations show that local drag constraints do not reliably
recover the intended global anisotropic scaling.
\begin{table}[t]
\centering
\small
\setlength{\tabcolsep}{0.8pt}

\begin{tabular}{@{}cccccccc@{}}
\toprule

\multicolumn{4}{c|}{%
    \textbf{Gray Background}%
}
&
\multicolumn{4}{c}{%
    \textbf{Real Background}%
} \\
\cmidrule(lr){1-4}
\cmidrule(lr){5-8}

MSE$\downarrow$
& PSNR$\uparrow$
& SSIM$\uparrow$
& \multicolumn{1}{c|}{LPIPS$\downarrow$}
& MSE$\downarrow$
& PSNR$\uparrow$
& SSIM$\uparrow$
& LPIPS$\downarrow$ \\
\midrule

268.14
& 26.38
& 0.944
& \multicolumn{1}{c|}{0.104}
& 1425.78
& 19.11
& 0.698
& 0.286 \\

\midrule

\multicolumn{6}{c|}{%
    \textbf{Foreground Fidelity}%
}
&
\multicolumn{2}{c}{%
    \textbf{LLM Eval}%
} \\
\cmidrule(lr){1-6}
\cmidrule(lr){7-8}

MSE$\downarrow$
& PSNR$\uparrow$
& SSIM$\uparrow$
& LPIPS$\downarrow$
& DINO$\uparrow$
& \multicolumn{1}{c|}{Dream$\uparrow$}
& GPT$\uparrow$
& Gemini$\uparrow$ \\
\midrule

697.63
& 21.87
& 0.826
& 0.224
& 0.635
& \multicolumn{1}{c|}{0.791}
& 1.833
& 2.146 \\

\midrule

\multicolumn{7}{c|}{%
    \textbf{Scale Accuracy and Geometric Alignment}%
}
&
\multicolumn{1}{c}{%
    \textbf{Temporal}%
} \\
\cmidrule(lr){1-7}
\cmidrule(lr){8-8}

IoU$\uparrow$
& Area$\downarrow$
& Yaw$\downarrow$
& Pitch$\downarrow$
& Dist$\downarrow$
& Trans$\downarrow$
& \multicolumn{1}{c|}{FIoU$\uparrow$}
& Ewarp$\downarrow$ \\
\midrule

0.449
& 0.837
& 0.460
& 0.256
& 0.403
& 0.375
& \multicolumn{1}{c|}{0.830}
& 13.738 \\

\bottomrule
\end{tabular}
\caption{Quantitative results of FlowDrag on our three benchmarks. Ewarp is at the range of $1\times 10^{-3}$.}
\label{tab:flowdrag}
\end{table}

\subsection{Overall Performance Evaluation on Geometry Benchmark}
Table~\ref{tab:main_quantitative} reports temporal consistency and perceptual quality on the Geometry Benchmark, where all videos are rendered on a uniform gray background to eliminate the influence of background variation. Under this controlled setting, the evaluation focuses solely on the quality of geometric editing and temporal coherence. Our method achieves the highest GPT, Gemini, and user preference scores, demonstrating that it produces the most realistic and perceptually faithful scaling results. Although Se~norita achieves the lowest Ewarp, it receives substantially lower perceptual scores, indicating that low temporal warping error alone does not necessarily translate into better visual quality. These results suggest that our method strikes a better balance between temporal consistency and faithful geometry-aware editing.
\begin{table}[t]
\centering
\small
\setlength{\tabcolsep}{4pt}
\renewcommand{\arraystretch}{1.08}
\begin{tabular}{l|cccc}
\toprule
Method
& Ewarp$\downarrow$
& GPT$\uparrow$
& Gemini$\uparrow$
& User Pref.$\uparrow$ \\
\midrule

DiffHandles
& 8.1166
& 1.6522
& 1.8867
& 0.0379 \\

Ditto
& 4.7164
& 1.1875
& 0.8667
& 0.0517 \\

Flux Fill
& 25.3799
& 1.4375
& 1.6341
& 0.0517 \\

Flux Kontext
& 4.9252
& 2.5625
& \underline{2.4359}
& 0.0724 \\

FreeFine
& 9.0984
& 1.0638
& 1.9756
& 0.0552 \\

GeoDiffuser
& 5.6394
& 1.4167
& 1.6809
& \underline{0.1138} \\

HqEdit
& 147.4057
& 0.6875
& 0.5882
& 0.0276 \\

InsV2V
& 4.8068
& 1.8958
& 1.3659
& 0.0517 \\

InsVIE
& 3.8690
& 1.3958
& 2.1591
& 0.0345 \\

LucyEdit
& 3.2542
& 1.5625
& 1.4773
& 0.0241 \\

Qwen-Image-E
& 3.3570
& \underline{2.7083}
& 2.1951
& 0.0345 \\

Se\~norita
& \textbf{0.5659}
& 1.4043
& 1.5500
& 0.0207 \\

Shape4Motion
& \underline{0.6178}
& 1.2632
& 1.9091
& 0.0345 \\

\midrule

\textbf{ScaleVid (Ours)}
& 0.7771
& \textbf{2.7542}
& \textbf{3.1471}
& \textbf{0.7793} \\

\bottomrule
\end{tabular}
\caption{Quantitative comparison of temporal consistency on Geometry Benchmark and perceptual quality.
The best results are \textbf{boldfaced} and the second-best results are
\underline{underlined}. Ewarp is at the range of $1\times 10^{-3}$.}
\label{tab:main_quantitative}
\end{table}

\section{Extensive Ablation Studies}
\subsubsection{Impact of Stage I Pretraining}
We extend the experiments in \textbf{[Main text, Table 3]}, as shown in Table~\ref{tab:ablation_all}. As shown in Table~\ref{tab:ablation_all}, the two training stages provide complementary benefits. Stage~I mainly improves background preservation, while Stage~II enhances temporal consistency and foreground fidelity through explicit deformation modeling. Stage I+II achieves the best foreground fidelity and global perceptual quality while retaining strong background preservation, achieving the highest Gemini score and the best results across all foreground fidelity metrics. Although the standalone Deformer obtains the lowest Ewarp and competitive semantic scores, it cannot directly preserve the real-world background. These results demonstrate that Stage~I provides reliable background reconstruction, whereas Stage~II introduces effective geometry-aware deformation, and their combination produces more faithful editing results.

\begin{table}[t]
\centering
\small
\setlength{\tabcolsep}{1pt}
\renewcommand{\arraystretch}{1.08}

\begin{tabular}{l|ccccccc}
\toprule

\multicolumn{8}{c}{%
    \textbf{Background Preservation and Temporal Consistency}%
} \\
\midrule

\multirow{2}{*}{Method}
& \multicolumn{4}{c|}{Real-world Background}
& \multicolumn{3}{c}{Overall Quality} \\
\cmidrule(lr){2-5}
\cmidrule(lr){6-8}

& MSE$\downarrow$
& PSNR$\uparrow$
& SSIM$\uparrow$
& \multicolumn{1}{c|}{LPIPS$\downarrow$}
& Ewarp$\downarrow$
& GPT$\uparrow$
& Gemini$\uparrow$ \\
\midrule

Deformer
& --
& --
& --
& \multicolumn{1}{c|}{--}
& \textbf{0.524}
& \textbf{2.813}
& \underline{2.711} \\

StageI
& \underline{130.95}
& 31.53
& 0.935
& \multicolumn{1}{c|}{0.042}
& 0.862
& 1.833
& 1.951 \\

StageII
& 181.17
& 28.62
& 0.862
& \multicolumn{1}{c|}{0.065}
& 1.342
& 2.506
& 2.211 \\

StageI+II
& 161.23
& \underline{30.58}
& \underline{0.920}
& \multicolumn{1}{c|}{\underline{0.049}}
& \underline{0.777}
& \underline{2.754}
& \textbf{3.147} \\

\midrule

\multicolumn{8}{c}{%
    \textbf{Foreground Fidelity}%
} \\
\midrule

Method
& MSE$\downarrow$
& PSNR$\uparrow$
& SSIM$\uparrow$
& LPIPS$\downarrow$
& DINO$\uparrow$
& \multicolumn{2}{c}{Dream$\uparrow$} \\
\midrule

Deformer
& \underline{401.90}
& \underline{24.37}
& \underline{0.838}
& \underline{0.137}
& \underline{0.838}
& \multicolumn{2}{c}{\underline{0.924}} \\

StageI
& 533.87
& 23.53
& 0.829
& 0.180
& 0.718
& \multicolumn{2}{c}{0.875} \\

StageII
& 544.73
& 22.85
& 0.823
& 0.161
& 0.739
& \multicolumn{2}{c}{0.897} \\

StageI+II
& \textbf{329.73}
& \textbf{25.33}
& \textbf{0.867}
& \textbf{0.133}
& \textbf{0.850}
& \multicolumn{2}{c}{\textbf{0.934}} \\

\bottomrule
\end{tabular}
\caption{Ablation study of different training stages.}
\label{tab:ablation_all}
\end{table}

\subsection{Impact of Masker Distillation Steps}

\begin{figure}[t]
    \centering
    \begin{subfigure}[t]{0.49\linewidth}
        \centering
        \includegraphics[width=\linewidth]{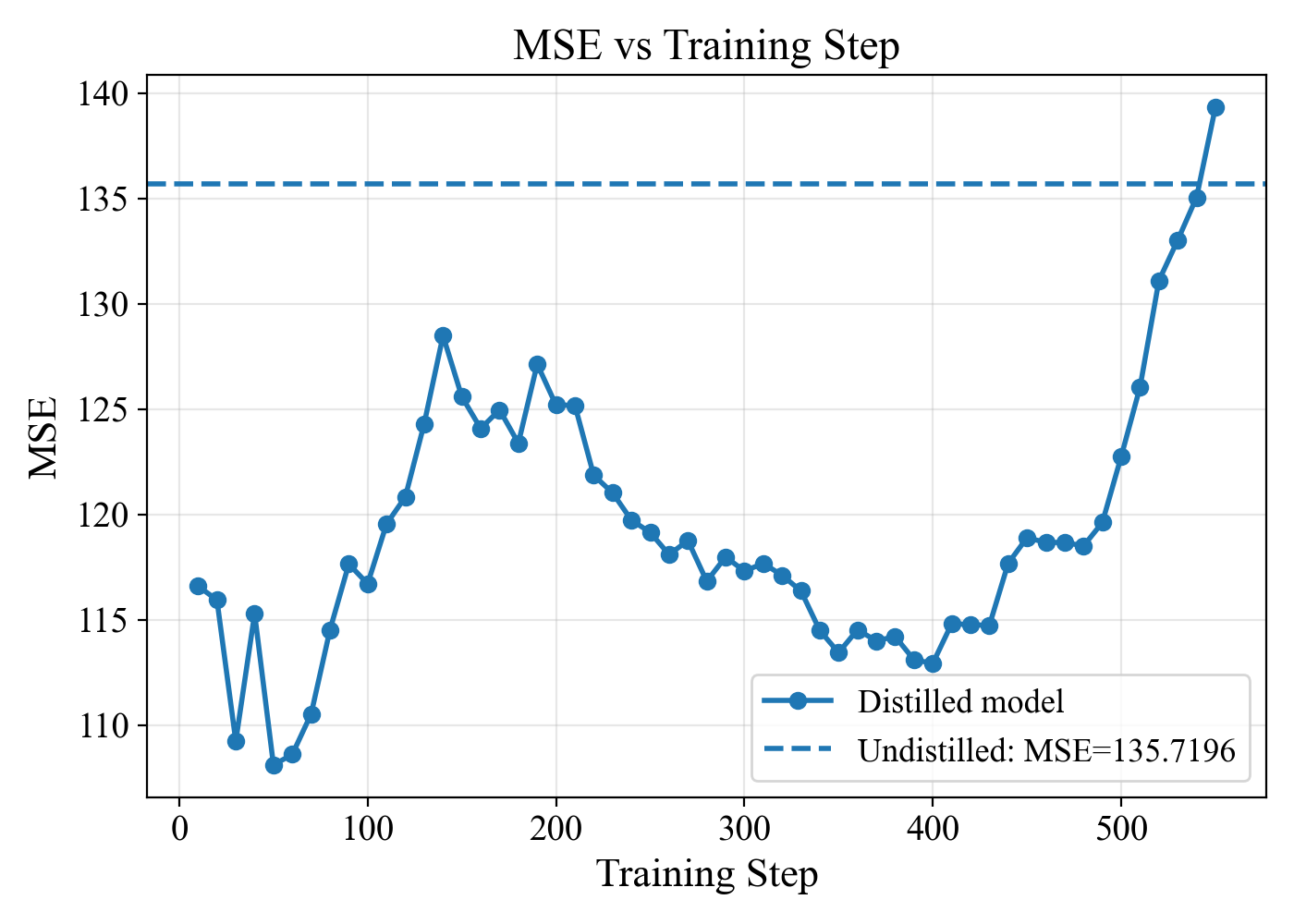}
        \caption{MSE versus training step. }
        \label{fig:mse_vs_step}
    \end{subfigure}
    \hfill
    \begin{subfigure}[t]{0.49\linewidth}
        \centering
        \includegraphics[width=\linewidth]{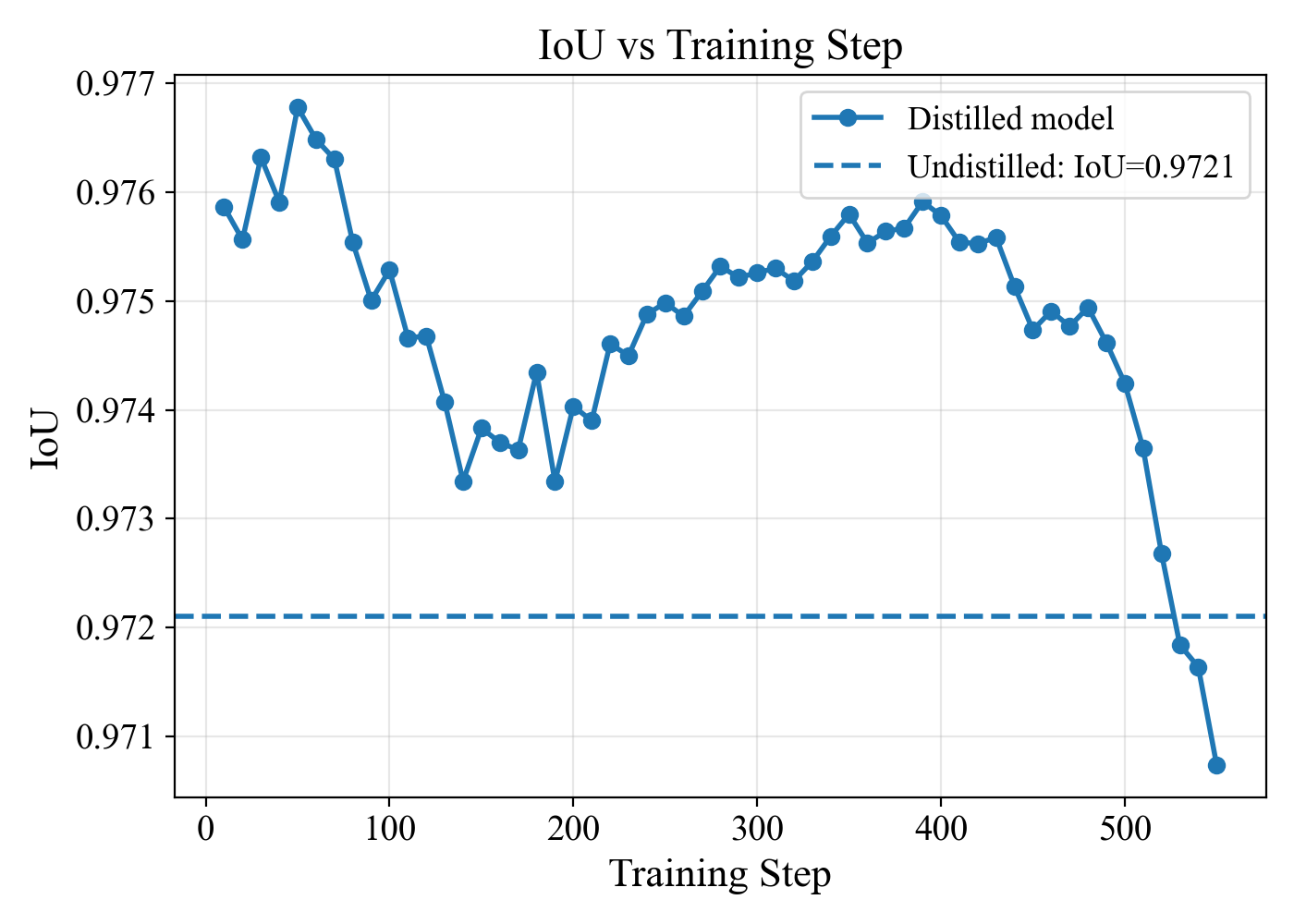}
        \caption{IoU versus training step. }
        \label{fig:iou_vs_step}
    \end{subfigure}
    \caption{Performance of the distilled Masker across training steps. We report MSE and IoU, and compare them with the undistilled baseline (step 0) under 20 denoising steps.}
    \label{fig:masker_step_curves}
\end{figure}

We analyze the influence of DMD training steps on the distilled Masker, where the student performs 3-step inference and the baseline is the undistilled 20-step model. As shown in Fig.~\ref{fig:masker_step_curves}, distillation improves both IoU and MSE across a wide range of training steps, indicating that the distilled Masker can preserve, and even enhance, mask quality under much fewer inference steps. Notably, the best results are obtained around 50 training steps, where the model achieves the highest IoU and the lowest MSE. After that, the performance fluctuates and gradually declines, with a clear degradation in the late stage of training. Therefore, we adopt 50 DMD training steps in the final model, as it yields the most favorable balance between efficiency and segmentation accuracy.

\subsection{Impact of Deformer Distillation Steps}
\begin{figure}[t]
    \centering
    \begin{subfigure}[t]{0.49\linewidth}
        \centering
        \includegraphics[width=\linewidth]{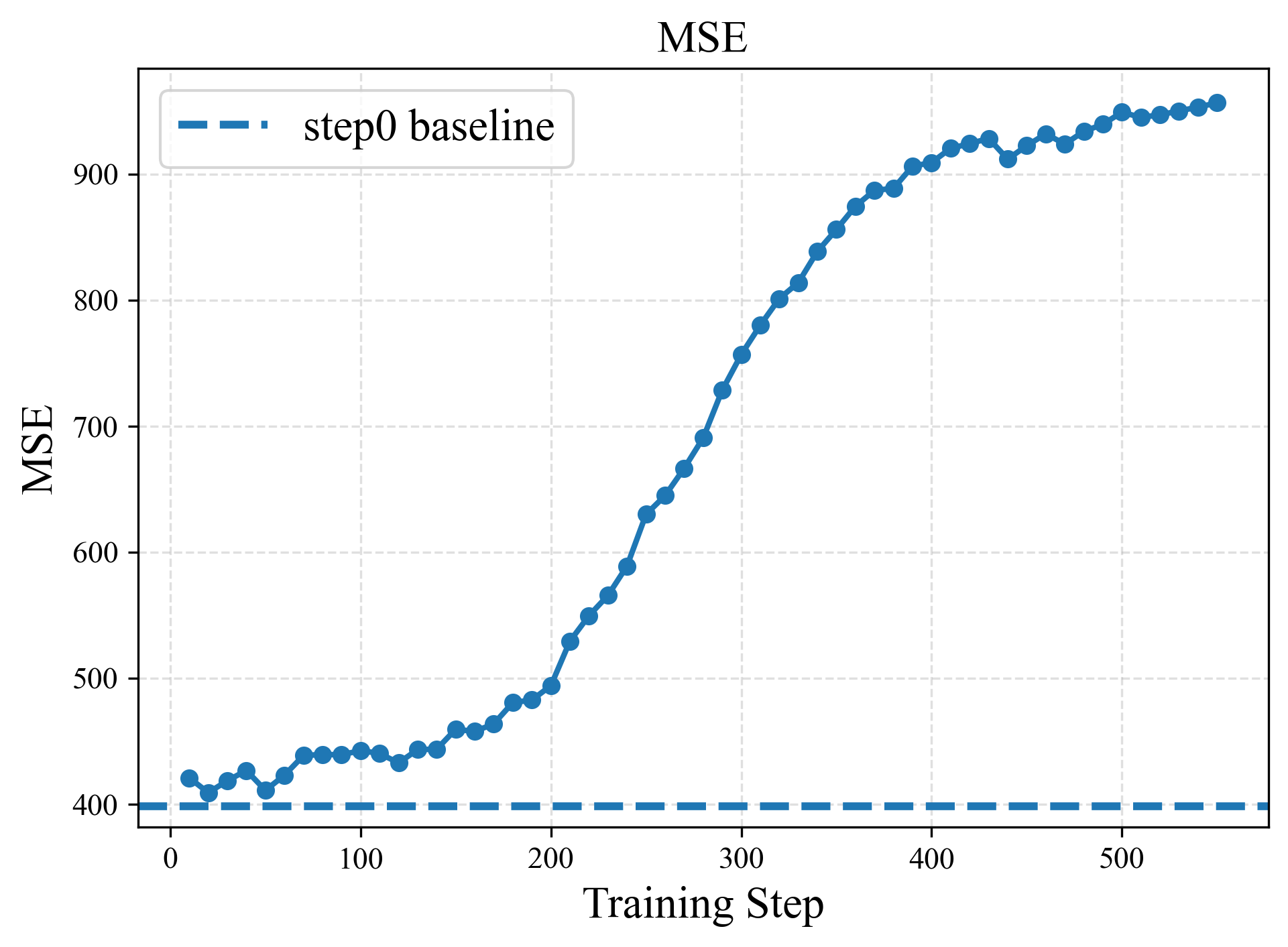}
        \caption{MSE versus training step.}
        \label{fig:mse1_vs_step}
    \end{subfigure}
    \hfill
    \begin{subfigure}[t]{0.49\linewidth}
        \centering
        \includegraphics[width=\linewidth]{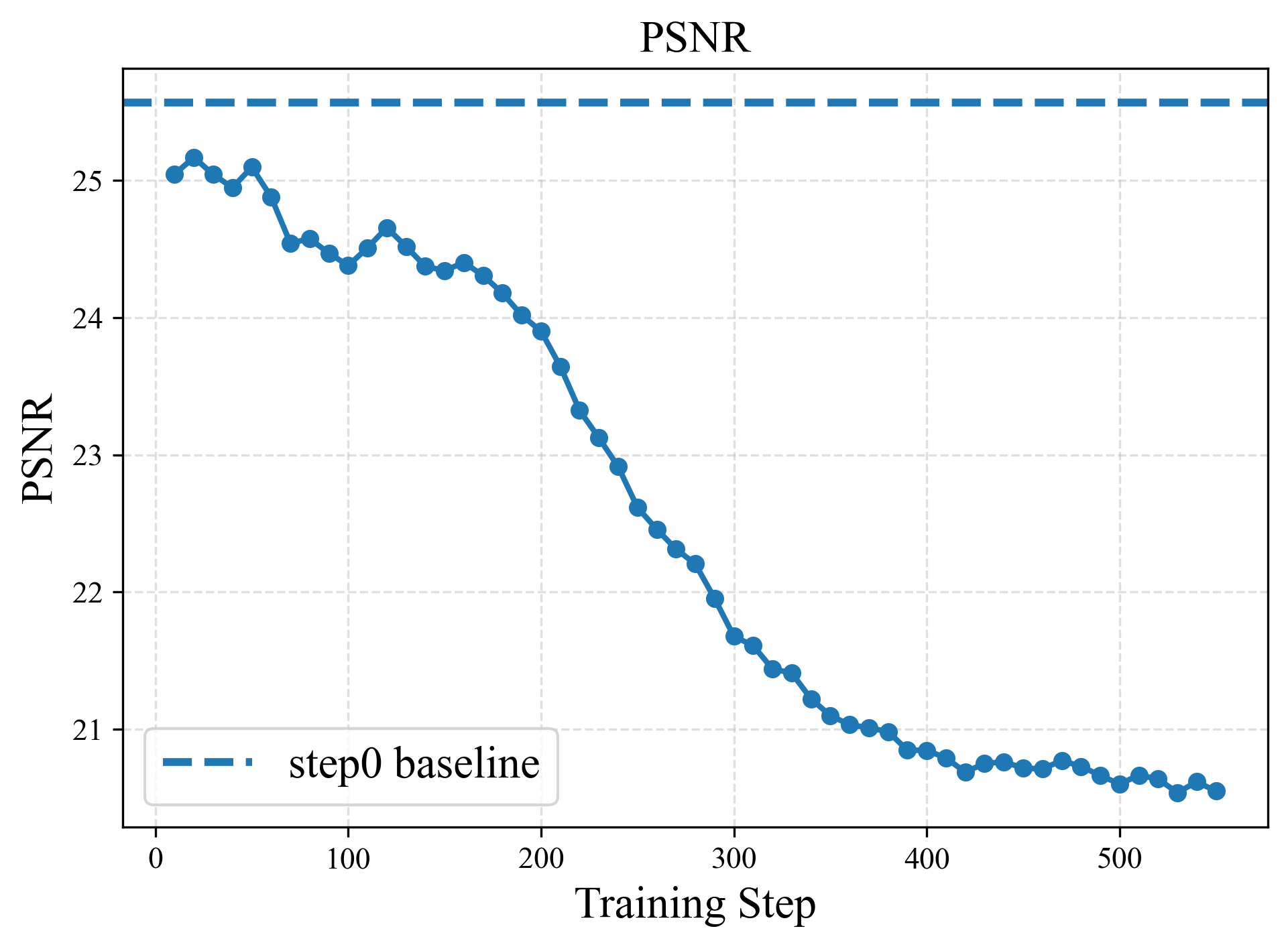}
        \caption{PSNR versus training step.}
        \label{fig:psnr_vs_step}
    \end{subfigure}

    \vspace{4pt}

    \begin{subfigure}[t]{0.49\linewidth}
        \centering
        \includegraphics[width=\linewidth]{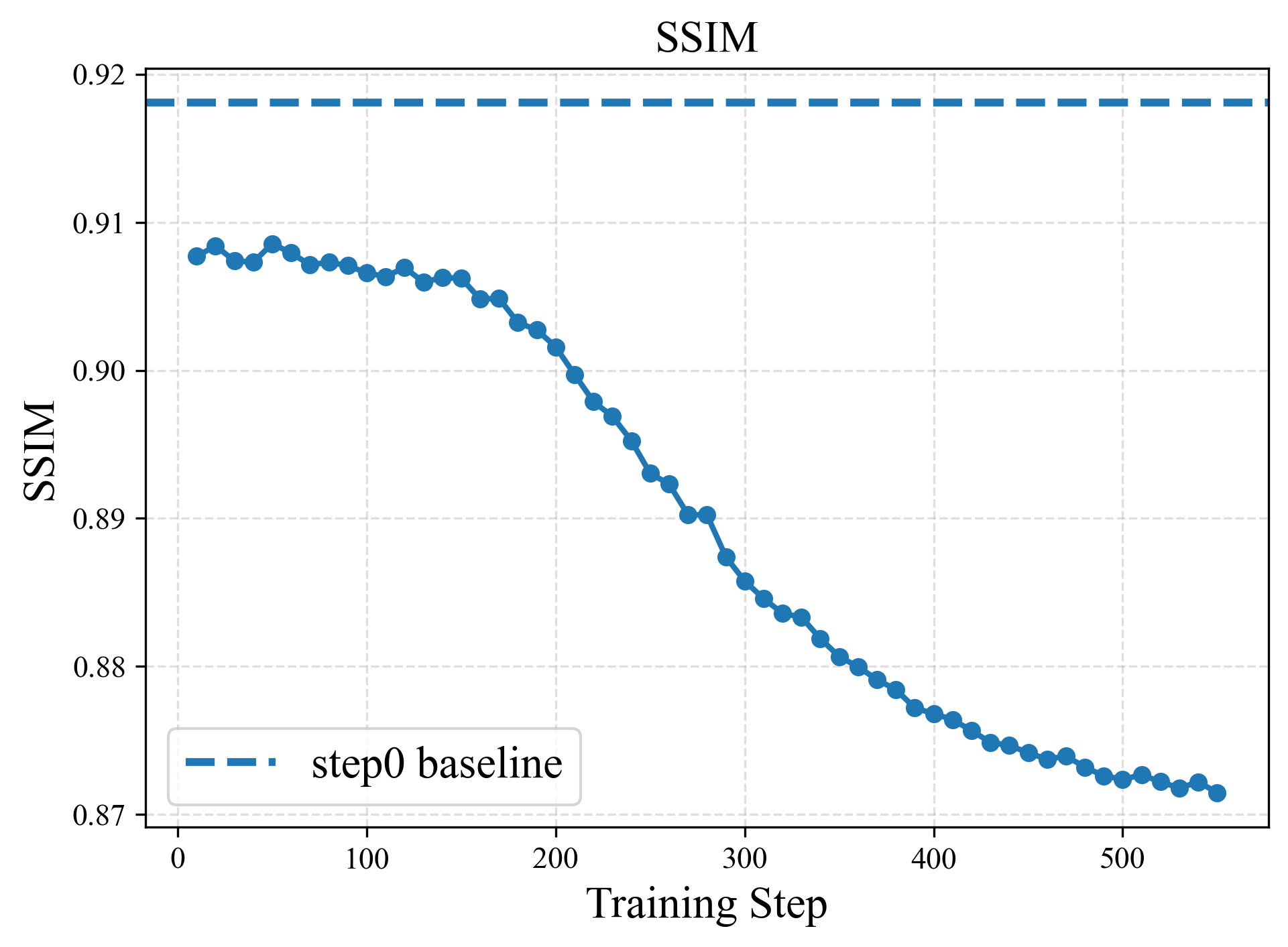}
        \caption{SSIM versus training step.}
        \label{fig:ssim_vs_step}
    \end{subfigure}
    \hfill
    \begin{subfigure}[t]{0.49\linewidth}
        \centering
        \includegraphics[width=\linewidth]{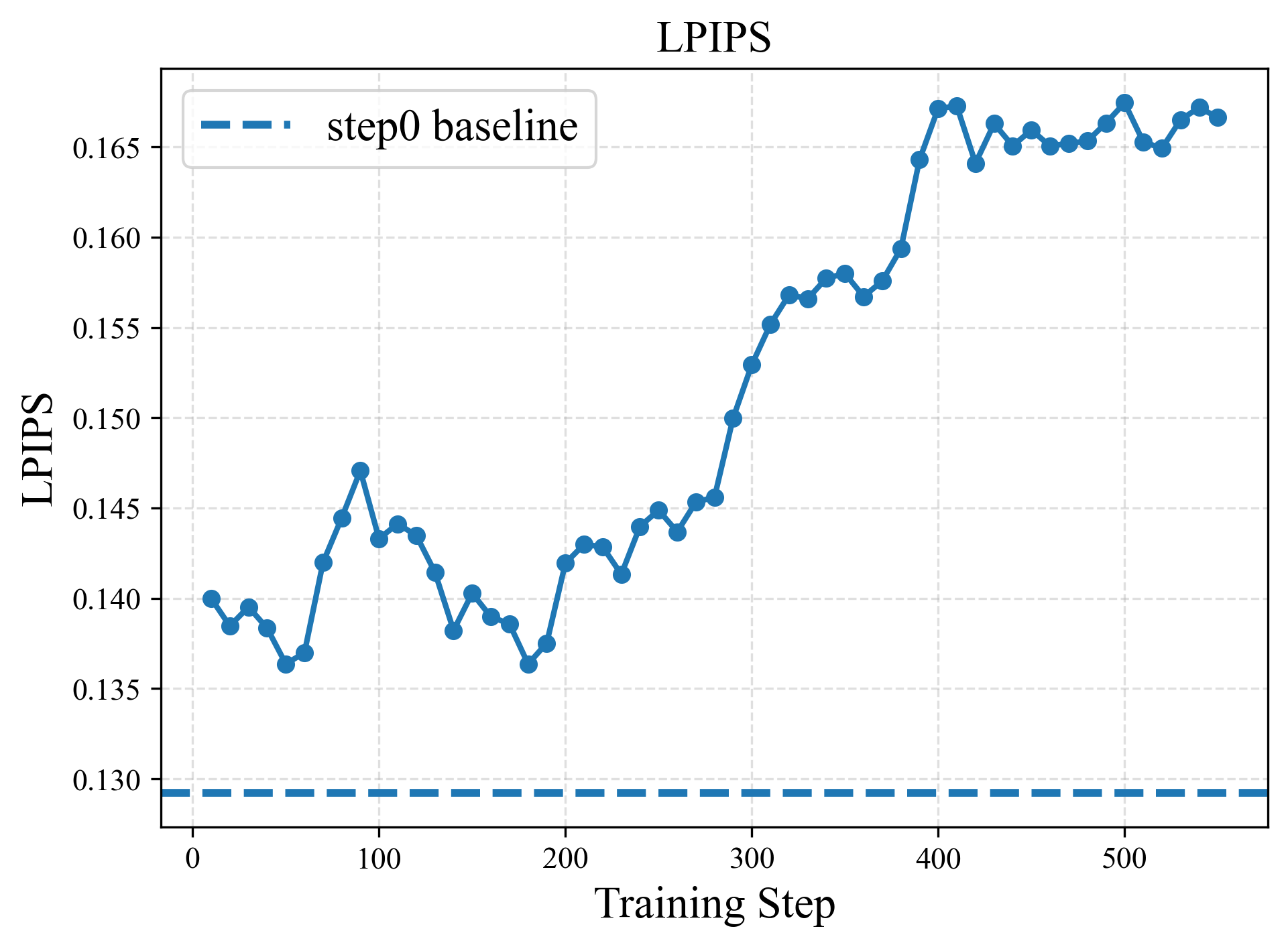}
        \caption{LPIPS versus training step.}
        \label{fig:lpips_vs_step}
    \end{subfigure}

    \caption{Performance across different distillation training steps. Step 0 denotes the undistilled baseline with 20 inference steps and is shown as a dashed reference line.}
    \label{fig:metric_step_curves}
\end{figure}

We also attempted to distill the Deformer into a 6-step model using DMD, and evaluated it every 10 distillation steps. 
We compare the full output video of the distilled 6-step Deformer against the ground-truth video using global MSE, PSNR, SSIM, and LPIPS. 
The results are shown in Fig.~\ref{fig:metric_step_curves}, where the dashed line denotes the performance of the original undistilled Deformer with 20-step inference. 
As shown in the figure, the distilled Deformer fails to achieve performance comparable to the undistilled model. 
Furthermore, increasing the number of distillation steps causes the model to deviate progressively from the desired distribution. 
We conjecture that the deformation task requires relatively accurate multi-step refinement, which is difficult to preserve under aggressive step reduction. 
Therefore, in training the Main Model, we continue to use the full 20-step Deformer at every training step.

\subsection{Impact of Scale Sampling Strategies}
As shown in Table~\ref{tab:ablation_scale_sampling}, log-uniform (Log-uni) sampling achieves better results than uniform sampling on all metrics. This is mainly because the scale range \((0.4, 2.5)\) is asymmetric in the original scale domain: uniform sampling places more probability mass on enlargement cases, while log-uniform sampling makes shrinking and enlargement more balanced. Consequently, the model trained with uniform sampling performs worse on shrinking cases at inference time, leading to inferior foreground fidelity and geometric alignment. In contrast, log-uniform sampling provides a more balanced training distribution and thus yields more robust overall performance.

\begin{table}[t]
\centering
\small
\setlength{\tabcolsep}{1.16pt}
\renewcommand{\arraystretch}{1.08}

\begin{tabular}{l|ccccccc}
\toprule

\multicolumn{8}{c}{%
    \textbf{Foreground Fidelity}%
} \\
\midrule

Method
& MSE$\downarrow$
& PSNR$\uparrow$
& SSIM$\uparrow$
& LPIPS$\downarrow$
& DINO$\uparrow$
& \multicolumn{2}{c}{Dream$\uparrow$} \\
\midrule

Uniform
& 560.79
& 22.92
& 0.822
& 0.164
& 0.788
& \multicolumn{2}{c}{0.910} \\

Log-uni
& \textbf{329.73}
& \textbf{25.33}
& \textbf{0.867}
& \textbf{0.133}
& \textbf{0.850}
& \multicolumn{2}{c}{\textbf{0.934}} \\

\midrule

\multicolumn{8}{c}{%
    \textbf{Scale Accuracy and Geometric Alignment}%
} \\
\midrule

Method
& IoU$\uparrow$
& \multicolumn{1}{c|}{Area$\downarrow$}
& Yaw$\downarrow$
& Pitch$\downarrow$
& Dist$\downarrow$
& Trans$\downarrow$
& FIoU$\uparrow$ \\
\midrule

Uniform
& 0.677
& \multicolumn{1}{c|}{0.397}
& 0.280
& 0.263
& 0.401
& 0.291
& 0.784 \\

Log-uni
& \textbf{0.804}
& \multicolumn{1}{c|}{\textbf{0.227}}
& \textbf{0.237}
& \textbf{0.205}
& \textbf{0.210}
& \textbf{0.178}
& \textbf{0.836} \\

\bottomrule
\end{tabular}
\caption{Comparison of different scale sampling strategies on the Geometry Benchmark.}
\label{tab:ablation_scale_sampling}
\end{table}

\subsection{Impact of Inverse Deformation}
We quantitatively evaluate the impact of inverse deformation on the
Real-Background Benchmark. As shown in Table~\ref{tab:inverse_deformation},
inverse deformation consistently improves both background preservation
and foreground fidelity.
Without inverse deformation, the Main Model is trained with clean
foreground conditions but receives Deformer-generated guidance during
inference, resulting in a training--inference gap. Therefore, it is
less capable of handling deformation artifacts and imperfect object
boundaries introduced by the Deformer.
By applying inverse deformation during training, the Main Model is
exposed to target-aligned but imperfect foreground conditions, which
better match the inference scenario. This enables the model to refine
the deformed foreground while maintaining high-quality foreground
appearance and seamless composition with the background. The improved mask IoU and area error further indicate that the
Main Model better follows the target-scale foreground guidance,
rather than improving appearance by restoring the object toward
its original geometry.

\begin{table}[t]
\centering
\small
\setlength{\tabcolsep}{1.5pt}

\begin{tabular}{@{}lcccccc@{}}
\toprule

\textbf{Method}
&
\multicolumn{4}{c|}{\textbf{Background Preservation}}
&
\multicolumn{2}{c}{\textbf{Scale Accuracy}}
\\
\cmidrule(lr){2-5}
\cmidrule(lr){6-7}

&
MSE$\downarrow$
& PSNR$\uparrow$
& SSIM$\uparrow$
& \multicolumn{1}{c|}{LPIPS$\downarrow$}
& IoU$\uparrow$
& Area$\downarrow$
\\
\midrule

w/o Inv. Def.
& 358.63
& 25.01
& 0.864
& \multicolumn{1}{c|}{0.102}
& 0.754
& 0.269
\\

Ours
& 161.23
& 30.58
& 0.920
& \multicolumn{1}{c|}{0.049}
& 0.799
& 0.232
\\

\midrule

\textbf{Method}
&
\multicolumn{6}{c}{\textbf{Foreground Fidelity}}
\\
\cmidrule(lr){2-7}

&
MSE$\downarrow$
& PSNR$\uparrow$
& SSIM$\uparrow$
& LPIPS$\downarrow$
& DINO$\uparrow$
& Dream$\uparrow$
\\
\midrule

w/o Inv. Def.
& 703.56
& 21.84
& 0.827
& \multicolumn{1}{c|}{0.170}
& 0.818
& 0.890
\\

Ours
& 430.87
& 24.98
& 0.864
& \multicolumn{1}{c|}{0.144}
& 0.834
& 0.927
\\

\bottomrule
\end{tabular}

\caption{
Quantitative ablation of inverse deformation on the
Real-Background Benchmark. We evaluate background preservation,
scale accuracy, and foreground fidelity.
}
\label{tab:inverse_deformation}
\end{table}

\subsection{Impact of CFG on Main Model}

As shown in Fig.~\ref{fig:cfg_main}, Tables~\ref{tab:cfg_main} and~\ref{tab:scale_geo_metrics}, the CFG~\cite{ho2022classifier} scale introduces a clear trade-off between foreground fidelity and geometric control. 
When the CFG scale is set to 1.0, the model achieves competitive foreground reconstruction quality on most appearance metrics, including MSE, PSNR, SSIM, LPIPS, DINO, and DreamSim, indicating that weaker guidance is beneficial for preserving object appearance. 
However, its geometric accuracy is not consistently optimal. 
As the CFG scale increases to 2.0 and 3.0, the model obtains more balanced performance, with clear improvements on several geometry-related metrics such as area error, yaw/pitch error, and distance error, while maintaining competitive foreground quality. 
In contrast, an excessively large CFG scale, such as 5.0, degrades foreground fidelity noticeably and does not further improve the overall geometric alignment in a stable manner. 
Therefore, we find that a moderate CFG scale around 2.0 provides the best trade-off between appearance preservation and controllable 3D-aware scaling, and is thus more suitable for the Main Model.

\begin{table}[t]
\centering
\small
\setlength{\tabcolsep}{4pt}
\renewcommand{\arraystretch}{1.08}

\begin{tabular}{lcccccc}
\toprule
CFG
& MSE$\downarrow$
& PSNR$\uparrow$
& SSIM$\uparrow$
& LPIPS$\downarrow$
& DINO$\uparrow$
& Dream$\uparrow$ \\
\midrule

1.0
& \textbf{333.43}
& \textbf{25.01}
& \textbf{0.853}
& \textbf{0.137}
& \textbf{0.850}
& \textbf{0.933} \\

2.0
& \underline{374.72}
& \underline{24.59}
& \underline{0.843}
& \underline{0.141}
& \underline{0.831}
& \underline{0.933} \\

3.0
& 404.26
& 24.25
& 0.834
& 0.144
& 0.832
& 0.928 \\

5.0
& 423.50
& 23.96
& 0.823
& 0.148
& 0.793
& 0.921 \\

\bottomrule
\end{tabular}
\caption{Foreground fidelity evaluation on the Geometry Benchmark of Main Model.}
\label{tab:cfg_main}
\end{table}

\begin{table}[t]
\centering
\small
\setlength{\tabcolsep}{3.5pt}
\renewcommand{\arraystretch}{1.05}

\begin{tabular}{l|cc|ccccc}
\toprule
CFG
& IoU$\uparrow$
& Area$\downarrow$
& Yaw$\downarrow$
& Pitch$\downarrow$
& Dist$\downarrow$
& Trans$\downarrow$
& FIoU$\uparrow$ \\
\midrule

1.0
& 0.793
& 0.238
& 0.252
& 0.256
& \underline{0.210}
& 0.212
& 0.821 \\

2.0
& 0.784
& 0.241
& \textbf{0.239}
& 0.274
& 0.217
& \underline{0.184}
& \textbf{0.832} \\

3.0
& \underline{0.799}
& \textbf{0.187}
& \underline{0.241}
& \textbf{0.198}
& \textbf{0.204}
& 0.188
& \underline{0.829} \\

5.0
& \textbf{0.802}
& \underline{0.197}
& 0.266
& \underline{0.247}
& 0.219
& \textbf{0.165}
& 0.813 \\

\bottomrule
\end{tabular}
\caption{Scale accuracy and geometric alignment evaluation of the Geometry Benchmark on Main Model.}
\label{tab:scale_geo_metrics}
\end{table}

\subsection{Impact of CFG on Deformer}
As shown in Fig.~\ref{fig:cfg_deformer}, Table~\ref{tab:cfg_deformer}, applying CFG to the Deformer does not consistently improve overall performance. 
When CFG is disabled (i.e., CFG$=1.0$), the Deformer achieves the best results on most reconstruction metrics, including MSE, PSNR, SSIM, LPIPS, DINO, and DreamSim, indicating that unguided sampling already provides strong foreground fidelity. 
Although CFG$=2.0$ remains competitive on several perceptual metrics, the improvement is limited and does not justify the additional inference cost. 
As the CFG scale increases further, the Deformer output quality degrades noticeably across almost all metrics, showing that overly strong guidance is harmful in this setting. 
Therefore, considering both efficiency and performance, we adopt CFG$=1.0$ in training and inference by default. 
This choice yields low latency while remaining comparable to CFG$=2.0$ in quality.

\begin{table}[t]
\centering
\small
\setlength{\tabcolsep}{4pt}
\renewcommand{\arraystretch}{1.08}

\begin{tabular}{lcccccc}
\toprule
CFG
& MSE$\downarrow$
& PSNR$\uparrow$
& SSIM$\uparrow$
& LPIPS$\downarrow$
& DINO$\uparrow$
& Dream$\uparrow$ \\
\midrule

1.0
& \textbf{401.90}
& \textbf{24.37}
& \textbf{0.838}
& \textbf{0.137}
& \textbf{0.838}
& \textbf{0.924} \\

2.0
& 495.34
& 23.52
& 0.825
& 0.129
& 0.865
& 0.935 \\

3.0
& 667.12
& 22.24
& 0.811
& 0.146
& 0.821
& 0.918 \\

4.0
& 934.97
& 20.61
& 0.803
& 0.176
& 0.705
& 0.873 \\

\bottomrule
\end{tabular}
\caption{Impact of CFG on the Geometry Benchmark of Deformer.}
\label{tab:cfg_deformer}
\end{table}

\begin{figure}[t]
  \centering
  \renewcommand{\arraystretch}{1.0}

  {%
    \fontsize{8pt}{9.6pt}\selectfont
    \begin{tabular}{@{}*{3}{>{\centering\arraybackslash}m{0.30\linewidth}}@{}}
      \minibox{Source} & \minibox{Ground Truth} & \minibox{CFG=1.0} \\
    \end{tabular}%
  }

  \vspace{-1pt}

  \includegraphics[width=\linewidth]{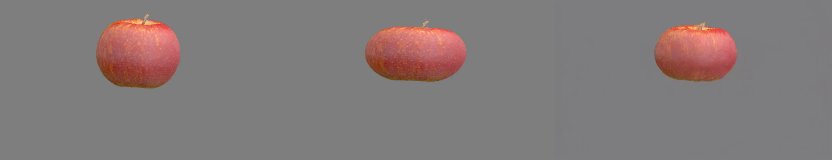}

  \vspace{2pt}

  {%
    \fontsize{8pt}{9.6pt}\selectfont
    \begin{tabular}{@{}*{3}{>{\centering\arraybackslash}m{0.30\linewidth}}@{}}
      \minibox{CFG=2.0} & \minibox{CFG=3.0} & \minibox{CFG=5.0}\\
    \end{tabular}%
  }

  \vspace{-1pt}

  \includegraphics[width=\linewidth]{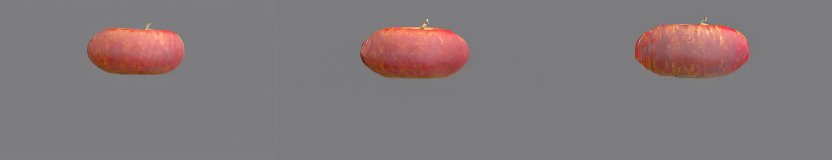}

  \caption{Visualization of Main Model under various CFG scales on the Geometry Benchmark.}
  \label{fig:cfg_main}
\end{figure}

\begin{figure}[t]
  \centering
  \renewcommand{\arraystretch}{1.0}

  {%
    \fontsize{8pt}{9.6pt}\selectfont
    \begin{tabular}{@{}*{3}{>{\centering\arraybackslash}m{0.30\linewidth}}@{}}
      \minibox{Source} & \minibox{Ground Truth} & \minibox{CFG=1.0} \\
    \end{tabular}%
  }

  \vspace{-1pt}

  \includegraphics[width=\linewidth]{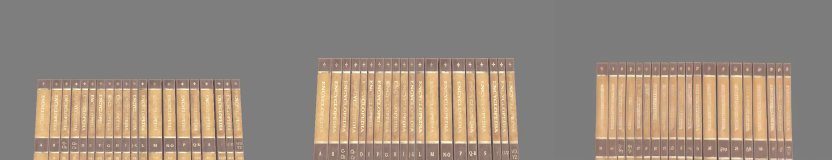}

  \vspace{2pt}

  {%
    \fontsize{8pt}{9.6pt}\selectfont
    \begin{tabular}{@{}*{3}{>{\centering\arraybackslash}m{0.30\linewidth}}@{}}
      \minibox{CFG=2.0} & \minibox{CFG=3.0} & \minibox{CFG=5.0}\\
    \end{tabular}%
  }

  \vspace{-1pt}

  \includegraphics[width=\linewidth]{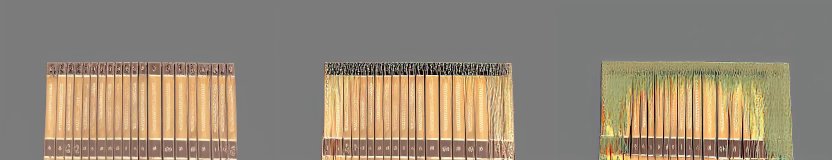}

  \caption{Visualization of Deformer under various CFG scales on the Geometry Benchmark.}
  \label{fig:cfg_deformer}
\end{figure}

\section{Extensive Qualitative Results}
\subsection{Main Model Can Compensate for Deformer Errors}
\begin{figure}[t]
  \centering
  \renewcommand{\arraystretch}{1.0}
  \setlength{\tabcolsep}{0pt}
  {%
    \fontsize{8pt}{9.6pt}\selectfont
    \begin{tabular}{@{}*{3}{>{\centering\arraybackslash}m{0.333\linewidth}}@{}}
      \minibox{Source} & \minibox{Main Model} & \minibox{Deformer}\\
    \end{tabular}%
  }

  \vspace{-1pt}

    \centering
    \includegraphics[width=\linewidth]{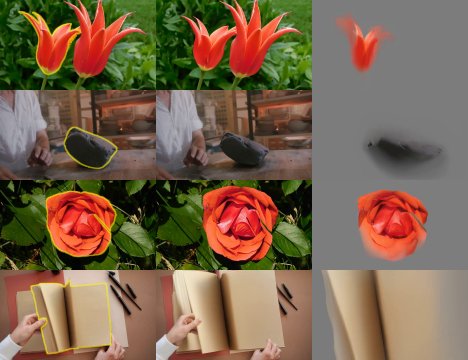}

  \caption{Visualization of Error-correction mechanism of Main Model. Though Deformer may provide blurred outputs, Main Model can still fix this.}
  \label{fig:err_cor}
\end{figure}

As shown in Fig.~\ref{fig:err_cor}, the Deformer does not always generate accurate foreground videos. Nevertheless, the Main Model can effectively mitigate the resulting blur artifacts. We believe this property arises from the bidirectional deformation strategy during training: although the conditioning inputs are always provided by the Deformer, the reconstruction target is consistently the real video. Consequently, discrepancies between imperfect deformed conditions and ground-truth videos are naturally introduced during training, which encourages the Main Model to learn to compensate for Deformer failures.

\subsubsection{Impact of the Bidirectional Loss for Deformer}

\begin{figure}[t]
  \centering
  \renewcommand{\arraystretch}{1.0}
  \setlength{\tabcolsep}{0pt}
  {%
    \fontsize{8pt}{9.6pt}\selectfont
    \begin{tabular}{@{}*{4}{>{\centering\arraybackslash}m{0.25\linewidth}}@{}}
      \minibox{Ground Truth} & \minibox{Mask} & \minibox{w/ Bid Loss} & \minibox{w/o Bid Loss} \\
    \end{tabular}%
  }

  \vspace{-1pt}

    \centering
    \includegraphics[width=\linewidth]{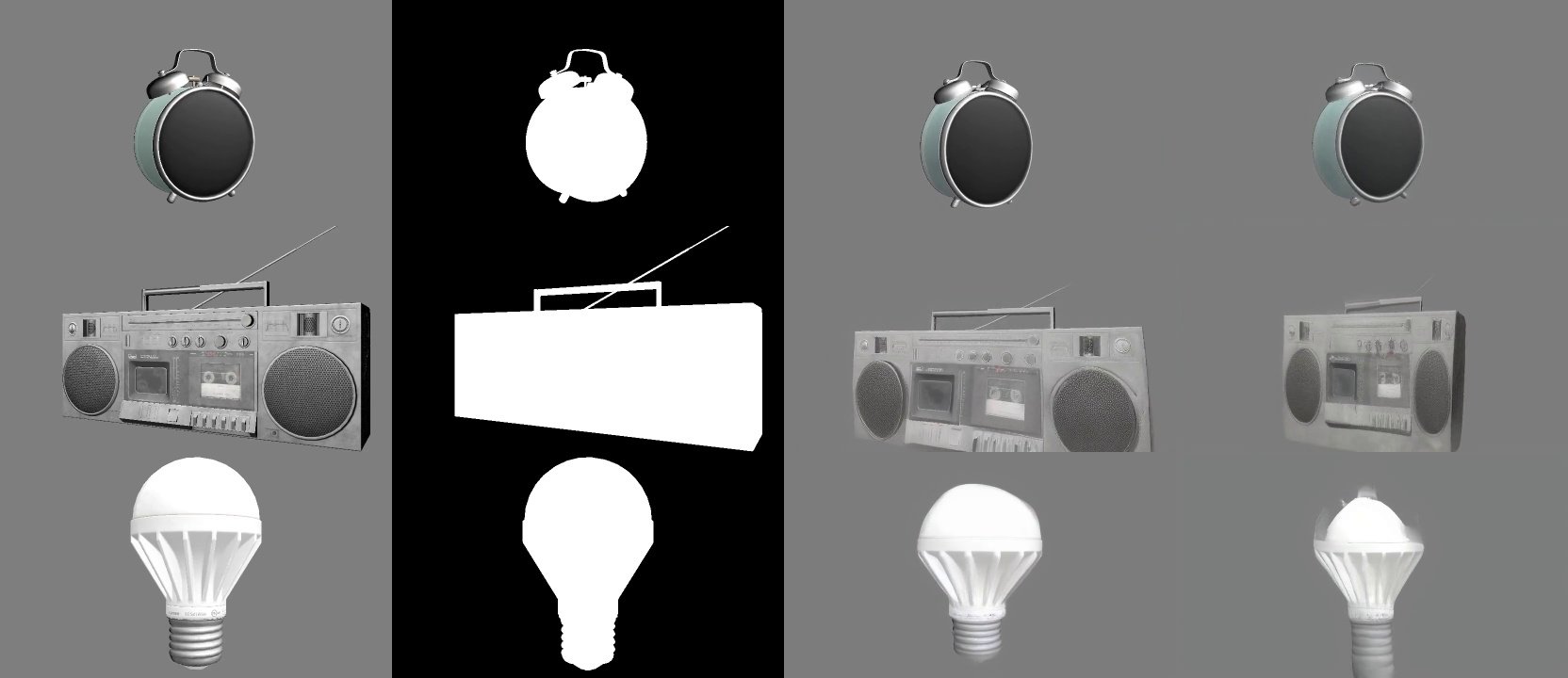}

  \caption{Effectiveness of the Bidirectional Loss. }
  \label{fig:dual}
\end{figure}

As shown in Fig.~\ref{fig:dual}, the bidirectional loss brings clear improvements to the Deformer in geometric alignment. This suggests that the bidirectional formulation effectively constrains the deformation trajectory and mitigates accumulated spatial drift.  We attribute this improvement to the fact that the bidirectional objective encourages the model to preserve transformation consistency under forward and reverse scaling, thereby learning a more geometrically faithful deformation field.

\subsubsection{Why Inverse Deformation is Needed}
\begin{figure}[t]
  \centering
  \renewcommand{\arraystretch}{1.0}
  \setlength{\tabcolsep}{0pt}
  {%
    \fontsize{8pt}{9.6pt}\selectfont
    \begin{tabular}{@{}*{4}{>{\centering\arraybackslash}m{0.25\linewidth}}@{}}
      \minibox{Source} & \minibox{Model Output} & \minibox{Source} & \minibox{Model Output} \\
    \end{tabular}%
  }

  \vspace{-1pt}

    \centering
    \includegraphics[width=\linewidth]{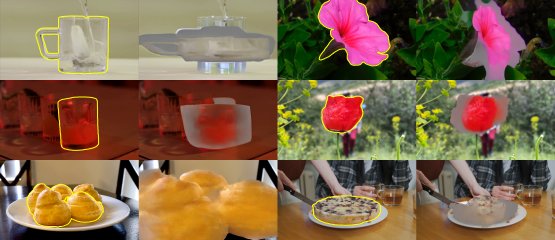}

  \caption{Inference results without inverse deformation during Main Model training. }
  \label{fig:bad}
\end{figure}

As shown in Fig.~\ref{fig:bad}, without inverse deformation the model suffers from a training--inference mismatch: during training it always receives the ground-truth foreground as input, while at inference time only the Deformer output is available. This encourages shortcut learning, where the model tends to copy and paste the foreground appearance instead of adapting to deformation artifacts. As a result, when the Deformer output is inaccurate, the model lacks the ability to correct such errors. Inverse deformation alleviates this issue by making the training input distribution more consistent with inference.

\subsection{Axis-wise Controllability of the Deformer}
\begin{figure}[t]
  \centering
  \renewcommand{\arraystretch}{1.0}
  \setlength{\tabcolsep}{0pt}

  \begin{tabular}{
    @{}
    >{\centering\arraybackslash}m{0.065\linewidth}
    >{\centering\arraybackslash}m{0.935\linewidth}
    @{}
  }

    &
    {%
      \fontsize{8pt}{9.6pt}\selectfont
      \begin{tabular}{
        @{}*{4}{>{\centering\arraybackslash}m{0.25\linewidth}}@{}
      }
        \minibox{$\mathbf{s}=(1,1,1)$}
        &
        \minibox{$\mathbf{s}=(0.7,1,1)$}
        &
        \minibox{$\mathbf{s}=(1,0.7,1)$}
        &
        \minibox{$\mathbf{s}=(1,1,0.7)$}
      \end{tabular}%
    }
    \\[-1pt]

    \rotatebox[origin=c]{90}{%
      \fontsize{8pt}{9.6pt}\selectfont
      \textbf{GT}
    }
    &
    \includegraphics[width=\linewidth]{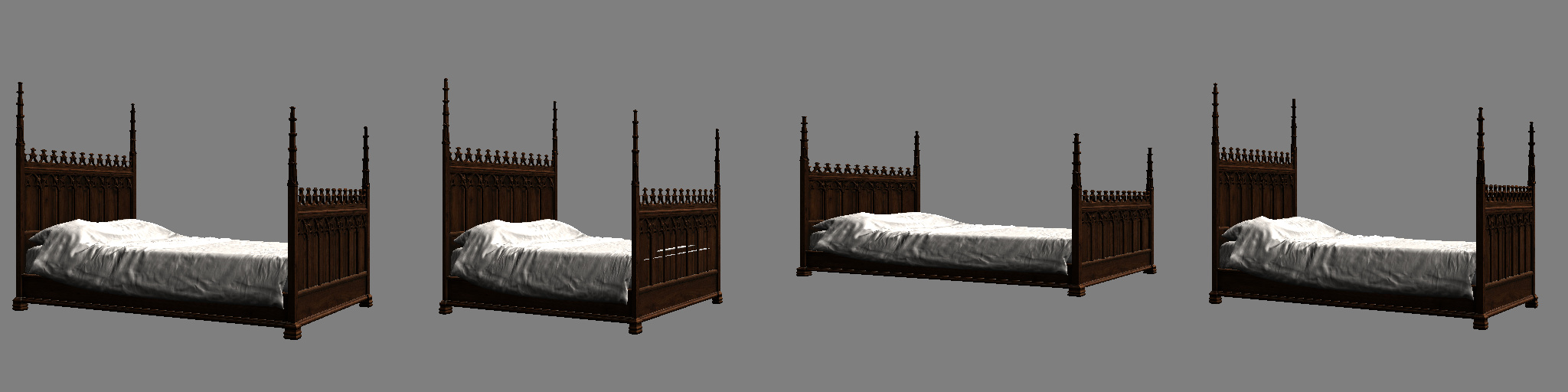}
    \\[2pt]

    \rotatebox[origin=c]{90}{%
      \fontsize{8pt}{9.6pt}\selectfont
      \textbf{Deformer}
    }
    &
    \includegraphics[width=\linewidth]{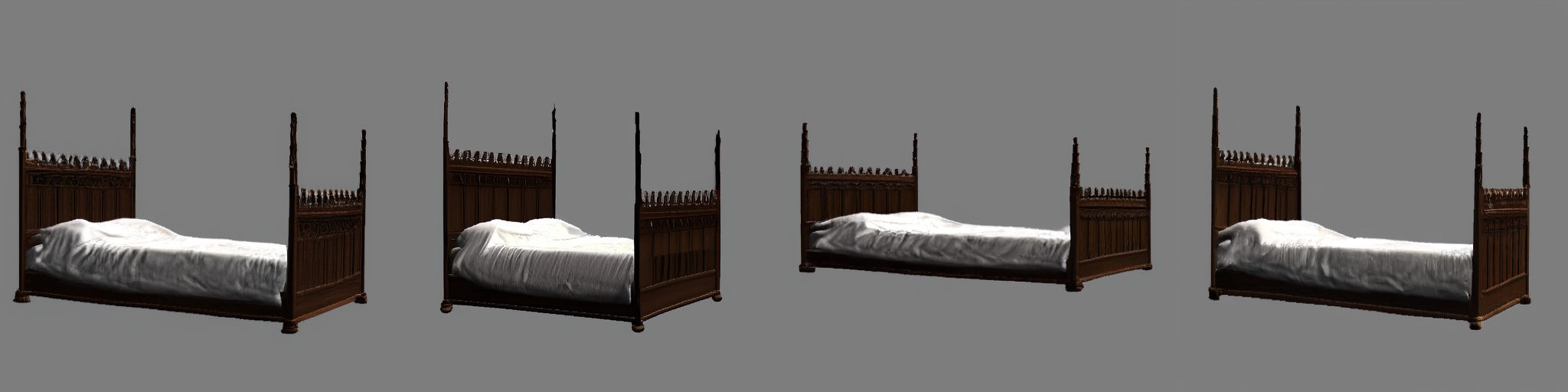}

  \end{tabular}

  \caption{
    Axis-wise scaling results. The top row shows mesh-rendered
    ground truth, and the bottom row shows Deformer outputs under
    identical scaling factors.
  }
  \label{fig:xyz}
\end{figure}

Figure~\ref{fig:xyz} qualitatively evaluates whether
the Deformer can distinguish the projected directions of the
canonical object axes. We independently shrink one axis to $0.7$
while keeping the other two scaling factors fixed to $1$. Compared
with the corresponding mesh-rendered ground truth, the Deformer
produces consistent axis-specific changes in object extent 
and visible structure. These results indicate that the Deformer can
respond differently to $s_x$, $s_y$, and $s_z$, rather than treating
them as equivalent image-plane scaling controls.

\subsection{Effect of Mask Dilation on Boundary Leakage}

\begin{figure}[t]
    \centering
    \includegraphics[width=0.985\linewidth]{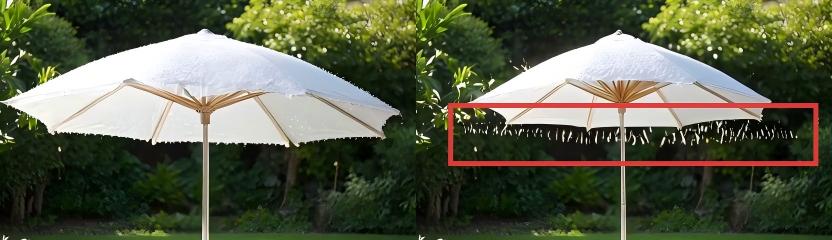}
    \caption{Effect of missing mask dilation. Left: the original frame. Right: the
edited result obtained without dilating the SAM2 mask.}
    \label{fig:mask_dilation}
\end{figure}

Figure~\ref{fig:mask_dilation} illustrates the importance of mask
dilation during preprocessing. SAM2 occasionally produces masks that
slightly under-segment the target object~\cite{zi2025minimax,huang2026refacade}. Without dilation, a narrow
band of the original object boundary remains in the estimated
background and is therefore treated as valid background content by the
Main Model.

As shown in Figure~\ref{fig:mask_dilation}, the edited result still
exhibits plausible perspective changes induced by 3D scaling, including
the exposure of additional umbrella ribs. However, multiple white
streaks appear below the umbrella canopy. These artifacts spatially
correspond to the original umbrella boundary, indicating that they are
caused by residual foreground pixels leaking into the background
condition rather than by the geometric transformation itself.

We therefore dilate the SAM2 mask during both training and inference to
remove uncertain boundary pixels and reduce foreground leakage into the
background condition.

\subsection{Visual Results of Deformer and Masker}
\begin{figure*}[t]
  \centering
  \renewcommand{\arraystretch}{1.0}
  {%
    \fontsize{8pt}{9.6pt}\selectfont
    \begin{tabular}{@{}*{6}{>{\centering\arraybackslash}m{0.147\linewidth}}@{}}
      \minibox{Source Foreground} & \minibox{Ground Truth} & \minibox{Model Output} & \minibox{Source Foreground} & \minibox{Ground Truth} & \minibox{Model Output} \\
    \end{tabular}%
  }

  \vspace{-1pt}

    \centering
    \includegraphics[width=\linewidth]{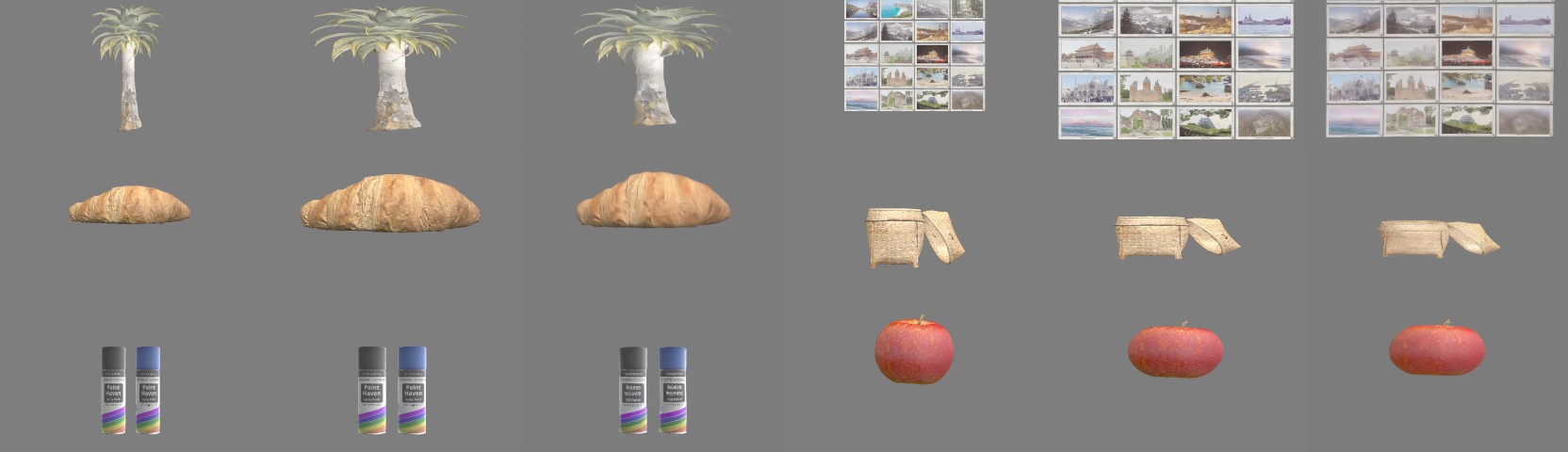}

  \caption{Visualization of our Deformer.}
  \label{fig:deformer}
\end{figure*}

\begin{figure*}[t]
  \centering
  \renewcommand{\arraystretch}{1.0}
  {%
    \fontsize{8pt}{9.6pt}\selectfont
    \begin{tabular}{@{}*{6}{>{\centering\arraybackslash}m{0.147\linewidth}}@{}}
      \minibox{Source Foreground} & \minibox{Ground Truth} & \minibox{Model Output} & \minibox{Source Foreground} & \minibox{Ground Truth} & \minibox{Model Output} \\
    \end{tabular}%
  }

  \vspace{-1pt}

    \centering
    \includegraphics[width=\linewidth]{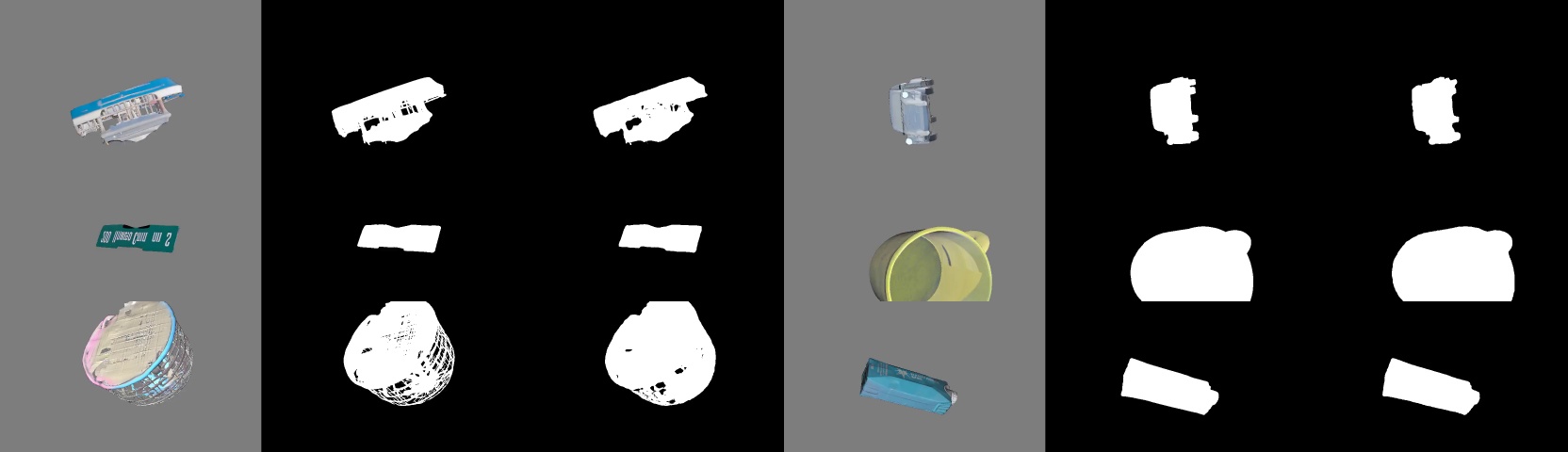}

  \caption{Visualization of our Masker.}
  \label{fig:masker}
\end{figure*}

Figures~\ref{fig:deformer},~\ref{fig:masker} present qualitative results of the Deformer and Masker, respectively. 
The Masker is able to predict accurate masks for regular objects, while being slightly weaker at capturing fine hollow structures in more complex objects. 
We attribute this limitation to the lightweight architecture design applied to the Masker for reducing training and inference latency, which reduces its model capacity. 
Nevertheless, the subsequent Main Model is able to tolerate and further correct such imperfect masks during video synthesis. 

\subsection{Comparison on Real-World Videos}
While the main benchmark used in this work is synthetic, it does not fully cover the distribution of real-world videos. 
To further assess the practical object scaling ability of our method against other baselines, we conduct additional qualitative experiments on real videos from the Pexels~\cite{pexels} and DAVIS~\cite{perazzi2016benchmark} datasets. 

Pexels mainly contains relatively stable objects and high-quality scenes, whereas DAVIS includes faster object motion and more severe deformation, making the scaling task more challenging. 
As illustrated in Fig.~\ref{fig:davis_compare} and Fig.~\ref{fig:pexels_compare}, our method consistently produces more accurate scaling results, with object transformations that better align with the target deformation parameters $\mathbf{s}$.

\section{Discussion}
\begin{figure}[t]
  \centering
  \renewcommand{\arraystretch}{1.0}
  {%
    \fontsize{8pt}{9.6pt}\selectfont
    \begin{tabular}{@{}*{4}{>{\centering\arraybackslash}m{0.210\linewidth}}@{}}
      \minibox{Source} & \minibox{Target} & \minibox{Source} & \minibox{Target} \\
    \end{tabular}%
  }

  \vspace{-1pt}

    \centering
    \includegraphics[width=\linewidth]{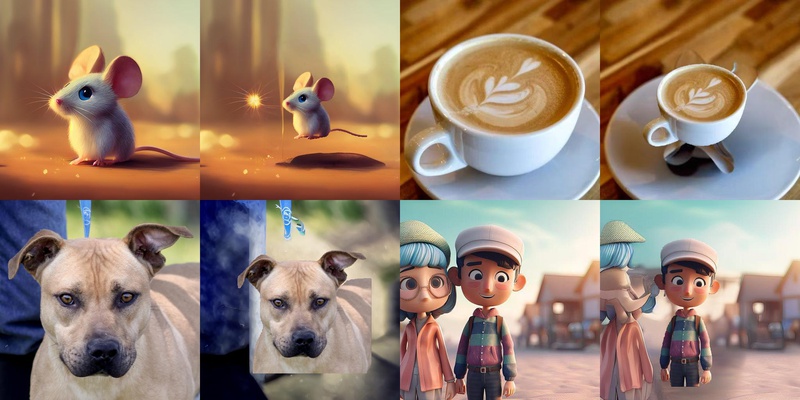}

  \caption{Visualization of Geobench~\cite{zhu2025training}.}
  \label{fig:geobench}
\end{figure}

\begin{figure}[t]
  \centering
  \renewcommand{\arraystretch}{1.0}
  \setlength{\tabcolsep}{0pt}
  {%
    \fontsize{8pt}{9.6pt}\selectfont
    \begin{tabular}{@{}*{3}{>{\centering\arraybackslash}m{0.33\linewidth}}@{}}
      \minibox{Source} & \minibox{2D Planar Scaling} & \minibox{Our Benchmark}\\
    \end{tabular}%
  }

  \vspace{-1pt}

    \centering
    \includegraphics[width=\linewidth]{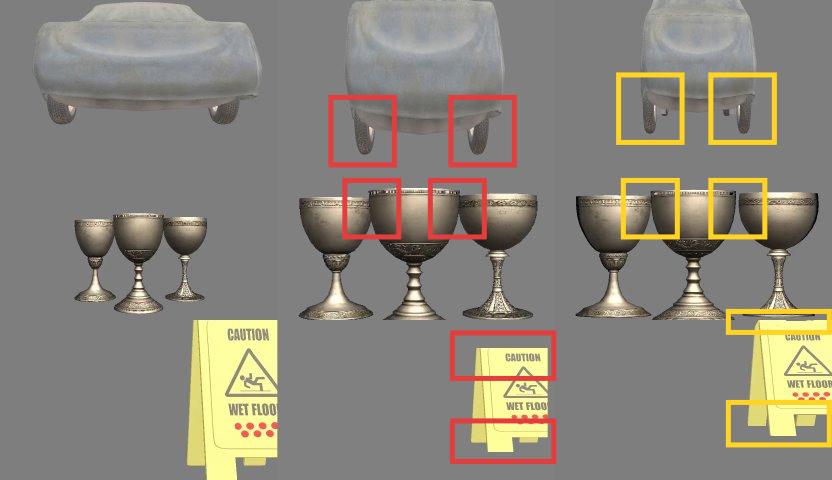}

  \caption{Visual comparison between 2D scaling and our benchmark. 2D scaling shares the same $(s_x,s_y)$.}
  \label{fig:bench_2d_3d}
\end{figure}

\paragraph{Why do we evaluate on a synthetic benchmark?}
The main reason is that real videos generally do not provide geometry-aware object scaling pairs: in most cases, only the source video is available, while the corresponding target video after controlled 3D-aware scaling does not exist. 
If one attempts to construct such pairs from real data, the common practice is to resize the foreground object in the 2D image plane and then composite it back into the background. 
However, this process cannot guarantee physically plausible lighting, consistent surface appearance, or object structures that conform to true 3D geometric transformations, as the Geobench~\cite{zhu2025training} shown in Fig.~\ref{fig:geobench}. 
Figure~\ref{fig:bench_2d_3d} presents a visual comparison between 2D planar scaling and our benchmark, where the 2D baseline shares the same $(s_x, s_y)$. Unlike 2D planar scaling, our benchmark captures the perspective changes caused by 3D transformations. For instance, when $s_x<1$, the rear wheel of the car becomes visible, which cannot be reproduced by a purely 2D operation. In addition, because scaling is applied to all selected objects jointly, transformations in multi-object scenes may also lead to new occlusion relationships.

In contrast, our synthetic benchmark is built from mesh-rendered paired videos and is strictly paired under controlled transformations. 
Beyond object motion and shape variation, it also preserves geometry-aware surface material changes and lighting consistency induced by 3D scaling, making it much more suitable for quantitative evaluation. 
Therefore, although synthetic data cannot fully cover the distribution of real videos, it provides a controlled and reliable testbed for measuring scale accuracy, geometric alignment, and appearance preservation. 
To complement this controlled benchmark, we further provide qualitative comparisons on real datasets, where our method also demonstrates strong object scaling capability in practical scenarios.

\paragraph{Role of the Real-Background Benchmark.}
The Real-Background Benchmark serves as a semi-realistic paired
evaluation between fully controlled synthetic testing and unpaired
real-world evaluation. It combines controllably rendered foregrounds
with real video backgrounds, preserving exact geometric targets while
introducing realistic scene textures, motion, and background
complexity. It therefore enables quantitative evaluation of foreground
fidelity and background preservation under more realistic conditions.

\paragraph{How is illumination consistency supported?}
In our framework, illumination consistency is encouraged by the Deformer
and the Main Model in complementary ways.
For the Deformer, it is learned from rendered training pairs in which only
the mesh vertices are rescaled, while the camera, motion, and scene
configurations remain unchanged. In particular, the lighting direction and
intensity are fixed across each pair. This provides controlled supervision
for learning geometry-dependent appearance changes under consistent
illumination.
For the Main Model, illumination consistency is learned implicitly from
real-video supervision. During training, the complete real video is always
used as the target, while the source-side conditions are constructed from
the foreground, mask, and background. This encourages the model to refine
the deformed foreground and recover an appearance that is compatible with
the observed real scene. Overall, the Deformer provides geometry-aware
guidance under controlled illumination, while the Main Model restores
realistic appearance through real-video targets.

\paragraph{Why do we use multiple modules?}
Our framework uses two modules at inference time: the Deformer and the Main Model. 
A natural question is why we do not merge the Deformer into the Main Model and train a single end-to-end model. 
The main reason is the lack of strictly paired real-video training data. 
If the Deformer were absorbed into the Main Model, the whole system could only be trained on mesh-rendered paired videos, since only such synthetic data provide controllable geometry-aware scaling pairs. 
However, real videos do not have strictly aligned target pairs under 3D-aware scaling. 
As a result, a unified model trained purely on rendered pairs would likely struggle to generalize to real videos: it may fail to preserve the real background and may also produce foregrounds that remain biased toward the rendered domain rather than the real pixel domain.

By separating the two modules, we assign them different roles. 
The Deformer is responsible for providing object structure that matches the target scaling parameter $\mathbf{s}$, while the Main Model focuses on foreground-background fusion and realism enhancement. 
This decoupled design allows the geometric transformation to be learned from controllable rendered data, while the final video synthesis is learned in the real-video domain. 
Therefore, our method can perform inference on real videos without suffering from a direct domain gap between rendered supervision and real-world outputs.

\paragraph{Relationship between Training Objective and Inference Geometry.}
A potential concern is whether the reconstruction objective of the
Main Model would restore the object to its original geometry during
training, since the original video is used as the reconstruction target.
However, the Main Model is not trained to infer or reverse the scaling
operation. Instead, the geometric configuration of its foreground
condition defines the desired output geometry.
During training, the inverse deformation step is introduced only to
align the transformed foreground with the original video target while
retaining the appearance artifacts produced by the Deformer. Therefore,
the Main Model learns geometry-preserving appearance refinement and
foreground-background composition rather than inverse scaling.
At inference time, replacing the training foreground condition with a
scaled foreground changes the target geometry specified to the model,
while the learned refinement capability remains unchanged.

\section{Limitations and Failure Cases}
\subsection{Limitations}
The main limitation of \abbr{} comes from the Deformer. 
Its training data are entirely constructed from mesh-rendered video pairs, which still exhibit a distribution gap from real pixel-space videos. 
Although the Main Model can partially bridge this gap through learned appearance synthesis and foreground-background fusion, the overall performance of \abbr{} is still bounded by the quality of the Deformer outputs. 
As a result, our method remains limited in reconstructing very fine object details in real videos, especially when the deformation results are imperfect.
Moreover, the OBB-based canonical axis definition can be ambiguous for geometrically symmetric objects, such as spheres and cylinders; for example, a cylinder does not provide a unique distinction between its two radial directions, making the semantics of $s_x$ and $s_z$ underdetermined.

\subsection{Failure Cases}

\begin{figure}[t]
    \centering
    \includegraphics[width=0.985\linewidth]{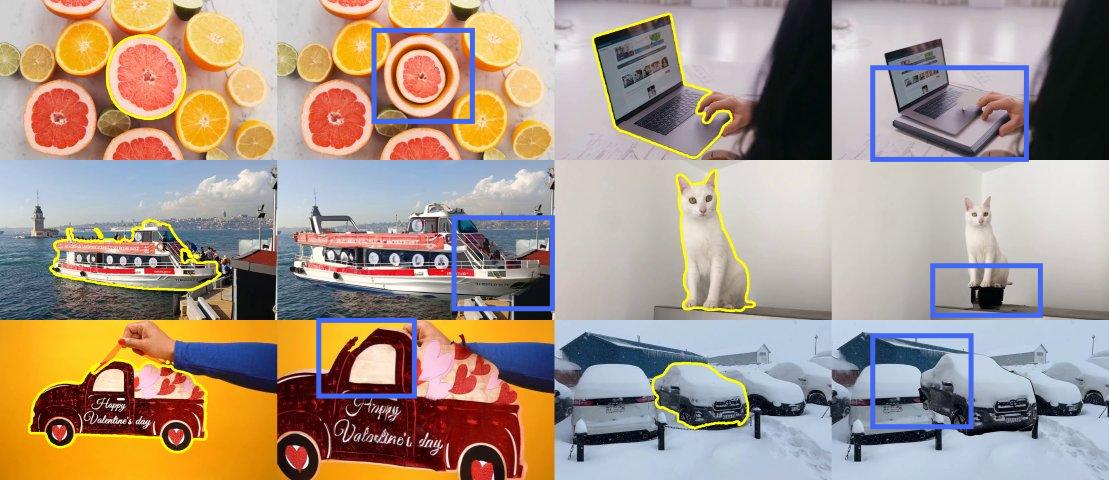}
    \caption{Visualization of failure cases.}
    \label{fig:failure}
\end{figure}

We show four representative failure cases in Fig.~\ref{fig:failure}. 
In the boat example, the target object is globally enlarged with a very large scaling parameter $\mathbf{s}$, causing the bow to extend onto the shore and resulting in an unrealistic collision artifact. 
In the other examples, the target objects are globally shrunk. Although the main objects become smaller as intended, undesired content appears around their original silhouettes. Moreover, the reconstructed object depends on the quality of original mask, a visible gap can be observed on the roof of the car.

We attribute these failures to two factors. 
First, though our method uses real videos as targets, it still lacks sufficient modeling of object--scene interaction, especially under large geometric changes. 
Second, object shrinking inevitably reveals previously occluded background regions, making the final result highly dependent on object removal quality. 
In our pipeline, this capability is inherited from Minimax Remover as the Main Model is initialized by it, which is not perfect and may introduce spurious content or inconsistent background completion. 
As a result, such artifacts can be propagated to the final edited videos.

\section{Future Work}
In the future, we plan to improve \abbr{} from four aspects. 
First, a more powerful Deformer trained with more realistic data or synthetic-to-real adaptation may further reduce the distribution gap between mesh-rendered supervision and real videos. 
Second, explicitly modeling object--scene interaction could help handle challenging cases involving collision, occlusion, or large geometric changes. 
Third, improving object removal and background completion would likely enhance performance in shrinking scenarios, where previously occluded regions need to be plausibly revealed. Fourth, the current framework could be extended to more general object
manipulation tasks, including translation and rotation. Following the
nine-dimensional transformation representation used in GeoBench~\cite{zhu2025training}, the
current scaling vector $\mathbf{s}\in\mathbb{R}^{3}$ could be generalized
to
\begin{equation}
\mathbf{u}
=
(t_x,t_y,t_z,r_x,r_y,r_z,s_x,s_y,s_z)
\in\mathbb{R}^{9},
\end{equation}
where the first three dimensions parameterize translation, the next
three parameterize rotation, and the last three retain anisotropic
scaling control.


\clearpage
\begin{figure*}[!htp]
  \centering
  \renewcommand{\arraystretch}{1.0}
  \setlength{\tabcolsep}{0pt}

  \vspace{3pt}

  {%
    \fontsize{8pt}{9.6pt}\selectfont 
    \begin{tabular}{@{}*{8}{>{\centering\arraybackslash}m{0.125\linewidth}}@{}}
      \minibox{Source} & \minibox{Ground Truth} & \minibox{DiffHandles} & \minibox{Ditto} &
      \minibox{Flux-Fill} & \minibox{Flux-Kontext} & \minibox{FreeFine} & \minibox{GeoDiffuser}\\
    \end{tabular}%
  }

  \vspace{-1pt}
  \begin{subfigure}{\linewidth}
    \centering
    \includegraphics[width=\linewidth]{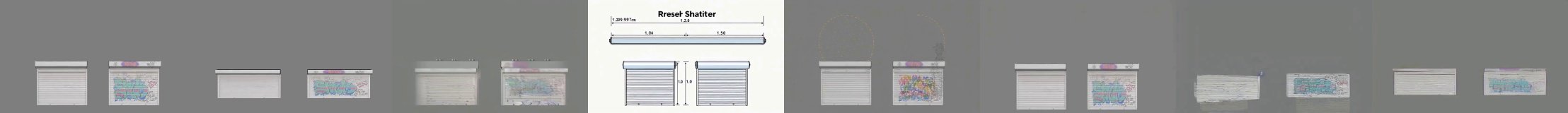}
  \end{subfigure}

  {%
    \fontsize{8pt}{9.6pt}\selectfont
    \vspace{-4pt}
    \begin{tabular}{@{}*{8}{>{\centering\arraybackslash}m{0.125\linewidth}}@{}}
      \minibox{HqEdit} & \minibox{InsV2V} & \minibox{InsVIE} & \minibox{Lucy-Edit} & \minibox{Qwen-Img-E}& \minibox{Señorita} &
      \minibox{Shape4Motion} & \minibox{ScaleVid (Ours)} \\
    \end{tabular}%
  }

  \vspace{-1pt}
  \begin{subfigure}{\linewidth}
    \centering
    \includegraphics[width=\linewidth]{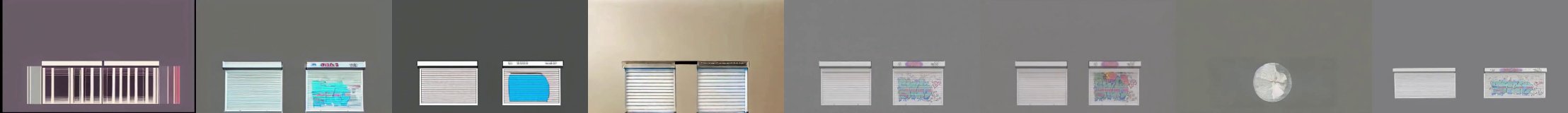}
  \end{subfigure}

    \vspace{3pt}
    {\captionsetup{font=footnotesize}
    \fbox{%
      \parbox{0.985\linewidth}{%
        \textbf{Prompt:} \textit{Rescale the \textbf{two rollers} by $1.2500$ times in width, $0.6500$ times in height, and $1.0000$ times in depth. Keep background unchanged.}
      }%
    }}

  {%
    \fontsize{8pt}{9.6pt}\selectfont 
    \begin{tabular}{@{}*{9}{>{\centering\arraybackslash}m{0.125\linewidth}}@{}}
      \minibox{Source} & \minibox{Ground Truth} & \minibox{DiffHandles} & \minibox{Ditto} &
      \minibox{Flux-Fill} & \minibox{Flux-Kontext} & \minibox{FreeFine} & \minibox{GeoDiffuser}\\
    \end{tabular}%
  }

  \vspace{-1pt}
  \begin{subfigure}{\linewidth}
    \centering
    \includegraphics[width=\linewidth]{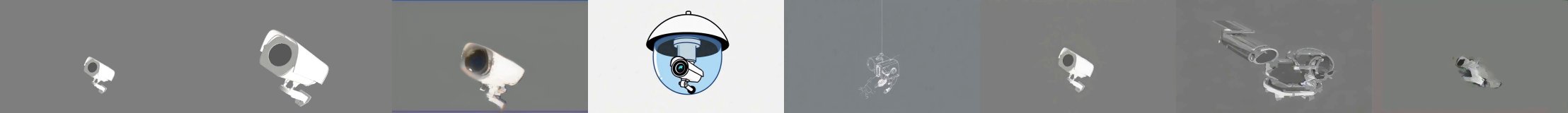}
  \end{subfigure}

  {%
    \fontsize{8pt}{9.6pt}\selectfont
    \vspace{-4pt}
    \begin{tabular}{@{}*{9}{>{\centering\arraybackslash}m{0.125\linewidth}}@{}}
      \minibox{HqEdit} & \minibox{InsV2V} & \minibox{InsVIE} & \minibox{Lucy-Edit} & \minibox{Qwen-Img-E}& \minibox{Señorita} &
      \minibox{Shape4Motion} & \minibox{ScaleVid (Ours)} \\
    \end{tabular}%
  }

  \vspace{-1pt}
  \begin{subfigure}{\linewidth}
    \centering
    \includegraphics[width=\linewidth]{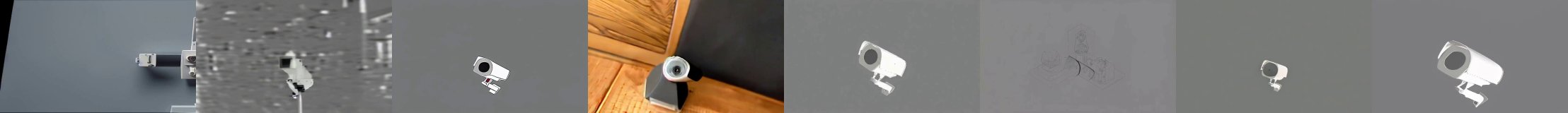}    
  \end{subfigure}

    \vspace{3pt}
    {\captionsetup{font=footnotesize}
    \fbox{%
      \parbox{0.985\linewidth}{%
        \textbf{Prompt:} \textit{Rescale the \textbf{camera} by $2.1700$ times in width, $2.0000$ times in height, and $2.1000$ times in depth. Keep background unchanged.}
      }%
    }}

  {%
    \fontsize{8pt}{9.6pt}\selectfont 
    \begin{tabular}{@{}*{9}{>{\centering\arraybackslash}m{0.125\linewidth}}@{}}
      \minibox{Source} & \minibox{Ground Truth} & \minibox{DiffHandles} & \minibox{Ditto} &
      \minibox{Flux-Fill} & \minibox{Flux-Kontext} & \minibox{FreeFine} & \minibox{GeoDiffuser}\\
    \end{tabular}%
  }

  \vspace{-1pt}
  \begin{subfigure}{\linewidth}
    \centering
    \includegraphics[width=\linewidth]{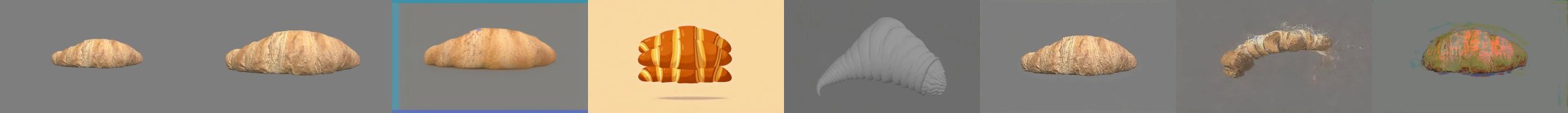}    
  \end{subfigure}

  {%
    \fontsize{8pt}{9.6pt}\selectfont
    \vspace{-4pt}
    \begin{tabular}{@{}*{9}{>{\centering\arraybackslash}m{0.125\linewidth}}@{}}
      \minibox{HqEdit} & \minibox{InsV2V} & \minibox{InsVIE} & \minibox{Lucy-Edit} & \minibox{Qwen-Img-E}& \minibox{Señorita} &
      \minibox{Shape4Motion} & \minibox{ScaleVid (Ours)} \\
    \end{tabular}%
  }

  \vspace{-1pt}
  \begin{subfigure}{\linewidth}
    \centering
    \includegraphics[width=\linewidth]{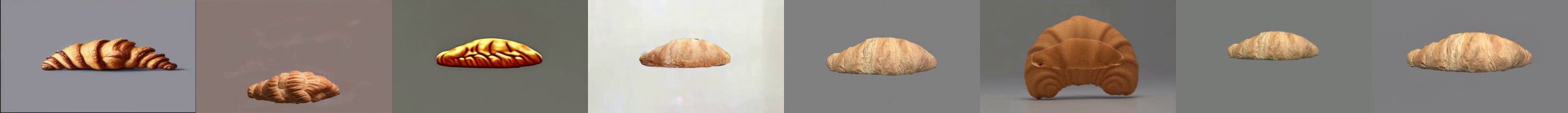}
  \end{subfigure}
  
    \vspace{3pt}
    {\captionsetup{font=footnotesize}
    \fbox{%
      \parbox{0.985\linewidth}{%
        \textbf{Prompt:} \textit{Rescale the \textbf{croissant} by $1.5000$ times in width, $1.5000$ times in height, and $1.5000$ times in depth. Keep background unchanged.}
      }%
    }}
    
  {%
    \fontsize{8pt}{9.6pt}\selectfont 
    \begin{tabular}{@{}*{9}{>{\centering\arraybackslash}m{0.125\linewidth}}@{}}
      \minibox{Source} & \minibox{Ground Truth} & \minibox{DiffHandles} & \minibox{Ditto} &
      \minibox{Flux-Fill} & \minibox{Flux-Kontext} & \minibox{FreeFine} & \minibox{GeoDiffuser}\\
    \end{tabular}%
  }

  \vspace{-1pt}
  \begin{subfigure}{\linewidth}
    \centering
        \includegraphics[width=\linewidth]{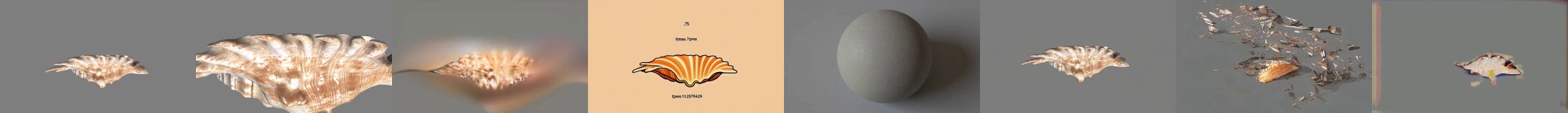}
  \end{subfigure}

  {%
    \fontsize{8pt}{9.6pt}\selectfont
    \vspace{-4pt}
    \begin{tabular}{@{}*{9}{>{\centering\arraybackslash}m{0.125\linewidth}}@{}}
      \minibox{HqEdit} & \minibox{InsV2V} & \minibox{InsVIE} & \minibox{Lucy-Edit} & \minibox{Qwen-Img-E}& \minibox{Señorita} &
      \minibox{Shape4Motion} & \minibox{ScaleVid (Ours)} \\
    \end{tabular}%
  }

  \vspace{-1pt}
  \begin{subfigure}{\linewidth}
    \centering
    \includegraphics[width=\linewidth]{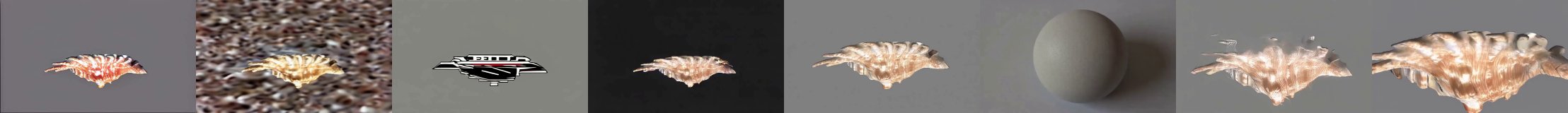}    
  \end{subfigure}
  
    \vspace{3pt}
    {\captionsetup{font=footnotesize}
    \fbox{%
      \parbox{0.985\linewidth}{%
        \textbf{Prompt:} \textit{Rescale the \textbf{shell} by $2.7500$ times in width, $2.6500$ times in height, and $0.8000$ times in depth. Keep background unchanged.}
      }%
    }}

  {%
    \fontsize{8pt}{9.6pt}\selectfont 
    \begin{tabular}{@{}*{9}{>{\centering\arraybackslash}m{0.125\linewidth}}@{}}
      \minibox{Source} & \minibox{Ground Truth} & \minibox{DiffHandles} & \minibox{Ditto} &
      \minibox{Flux-Fill} & \minibox{Flux-Kontext} & \minibox{FreeFine} & \minibox{GeoDiffuser}\\
    \end{tabular}%
  }

  \vspace{-1pt}
  \begin{subfigure}{\linewidth}
    \centering
        \includegraphics[width=\linewidth]{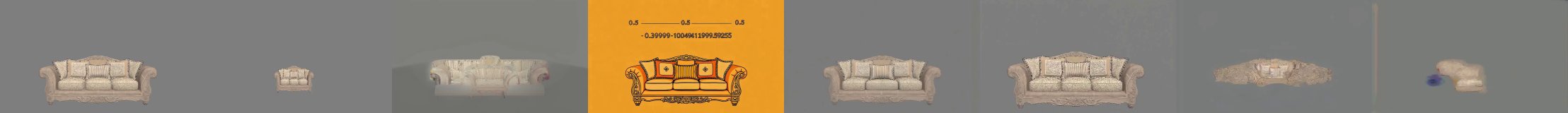}
  \end{subfigure}

  {%
    \fontsize{8pt}{9.6pt}\selectfont
    \vspace{-4pt}
    \begin{tabular}{@{}*{9}{>{\centering\arraybackslash}m{0.125\linewidth}}@{}}
      \minibox{HqEdit} & \minibox{InsV2V} & \minibox{InsVIE} & \minibox{Lucy-Edit} & \minibox{Qwen-Img-E}& \minibox{Señorita} &
      \minibox{Shape4Motion} & \minibox{ScaleVid (Ours)} \\
    \end{tabular}%
  }

  \vspace{-1pt}
  \begin{subfigure}{\linewidth}
    \centering
        \includegraphics[width=\linewidth]{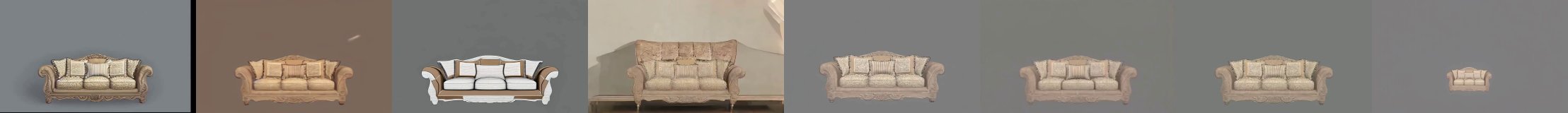}
  \end{subfigure}
  
    \vspace{3pt}
    {\captionsetup{font=footnotesize}
    \fbox{%
      \parbox{0.985\linewidth}{%
        \textbf{Prompt:} \textit{Rescale the \textbf{sofa} by $0.3500$ times in width, $0.5500$ times in height, and $0.5000$ times in depth. Keep background unchanged.}
      }%
    }}
    
  \caption{Comparison results of \abbr{} and baselines on the Geometry Benchmark.}
  \label{fig:bench_compare}
\end{figure*}

\begin{figure*}[!htp]
  \centering
  \renewcommand{\arraystretch}{1.0}
  \setlength{\tabcolsep}{0pt}

  \vspace{3pt}

  {%
    \fontsize{8pt}{9.6pt}\selectfont 
    \begin{tabular}{@{}*{8}{>{\centering\arraybackslash}m{0.125\linewidth}}@{}}
      \minibox{Source} & \minibox{Ground Truth} & \minibox{DiffHandles} & \minibox{Ditto} &
      \minibox{Flux-Fill} & \minibox{Flux-Kontext} & \minibox{FreeFine} & \minibox{GeoDiffuser}\\
    \end{tabular}%
  }

  \vspace{-1pt}
  \begin{subfigure}{\linewidth}
    \centering
    \animategraphics[width=\linewidth]{12}{images/davis/bear/1/}{1}{1}
  \end{subfigure}

  {%
    \fontsize{8pt}{9.6pt}\selectfont
    \vspace{-4pt}
    \begin{tabular}{@{}*{8}{>{\centering\arraybackslash}m{0.125\linewidth}}@{}}
      \minibox{HqEdit} & \minibox{InsV2V} & \minibox{InsVIE} & \minibox{Lucy-Edit} & \minibox{Qwen-Img-E}& \minibox{Señorita} &
      \minibox{Shape4Motion} & \minibox{ScaleVid (Ours)} \\
    \end{tabular}%
  }

  \vspace{-1pt}
  \begin{subfigure}{\linewidth}
    \centering
    \animategraphics[width=\linewidth]{12}{images/davis/bear/2/}{1}{1}
  \end{subfigure}

    \vspace{3pt}
    {\captionsetup{font=footnotesize}
    \fbox{%
      \parbox{0.985\linewidth}{%
        \textbf{Prompt:} \textit{Rescale the \textbf{bear} by $2.2391$ times in width, $1.7172$ times in height, and $1.1951$ times in depth. Keep background unchanged.}
      }%
    }}

  {%
    \fontsize{8pt}{9.6pt}\selectfont 
    \begin{tabular}{@{}*{9}{>{\centering\arraybackslash}m{0.125\linewidth}}@{}}
      \minibox{Source} & \minibox{Ground Truth} & \minibox{DiffHandles} & \minibox{Ditto} &
      \minibox{Flux-Fill} & \minibox{Flux-Kontext} & \minibox{FreeFine} & \minibox{GeoDiffuser}\\
    \end{tabular}%
  }

  \vspace{-1pt}
  \begin{subfigure}{\linewidth}
    \centering
    \animategraphics[width=\linewidth]{12}{images/davis/car-turn/1/}{1}{1}
  \end{subfigure}

  {%
    \fontsize{8pt}{9.6pt}\selectfont
    \vspace{-4pt}
    \begin{tabular}{@{}*{9}{>{\centering\arraybackslash}m{0.125\linewidth}}@{}}
      \minibox{HqEdit} & \minibox{InsV2V} & \minibox{InsVIE} & \minibox{Lucy-Edit} & \minibox{Qwen-Img-E}& \minibox{Señorita} &
      \minibox{Shape4Motion} & \minibox{ScaleVid (Ours)} \\
    \end{tabular}%
  }

  \vspace{-1pt}
  \begin{subfigure}{\linewidth}
    \centering
    \animategraphics[width=\linewidth]{12}{images/davis/car-turn/2/}{1}{1}
  \end{subfigure}

    \vspace{3pt}
    {\captionsetup{font=footnotesize}
    \fbox{%
      \parbox{0.985\linewidth}{%
        \textbf{Prompt:} \textit{Rescale the \textbf{car} by $0.8567$ times in width, $1.7076$ times in height, and $2.2359$ times in depth. Keep background unchanged.}
      }%
    }}

  {%
    \fontsize{8pt}{9.6pt}\selectfont 
    \begin{tabular}{@{}*{9}{>{\centering\arraybackslash}m{0.125\linewidth}}@{}}
      \minibox{Source} & \minibox{Ground Truth} & \minibox{DiffHandles} & \minibox{Ditto} &
      \minibox{Flux-Fill} & \minibox{Flux-Kontext} & \minibox{FreeFine} & \minibox{GeoDiffuser}\\
    \end{tabular}%
  }

  \vspace{-1pt}
  \begin{subfigure}{\linewidth}
    \centering
    \animategraphics[width=\linewidth]{12}{images/davis/gold-fish/1/}{1}{1}
  \end{subfigure}

  {%
    \fontsize{8pt}{9.6pt}\selectfont
    \vspace{-4pt}
    \begin{tabular}{@{}*{9}{>{\centering\arraybackslash}m{0.125\linewidth}}@{}}
      \minibox{HqEdit} & \minibox{InsV2V} & \minibox{InsVIE} & \minibox{Lucy-Edit} & \minibox{Qwen-Img-E}& \minibox{Señorita} &
      \minibox{Shape4Motion} & \minibox{ScaleVid (Ours)} \\
    \end{tabular}%
  }

  \vspace{-1pt}
  \begin{subfigure}{\linewidth}
    \centering
    \animategraphics[width=\linewidth]{12}{images/davis/gold-fish/2/}{1}{1}
  \end{subfigure}
  
    \vspace{3pt}
    {\captionsetup{font=footnotesize}
    \fbox{%
      \parbox{0.985\linewidth}{%
        \textbf{Prompt:} \textit{Rescale the \textbf{fishes} by $2.3211$ times in width, $1.4238$ times in height, and $0.8527$ times in depth. Keep background unchanged.}
      }%
    }}
    
  {%
    \fontsize{8pt}{9.6pt}\selectfont 
    \begin{tabular}{@{}*{9}{>{\centering\arraybackslash}m{0.125\linewidth}}@{}}
      \minibox{Source} & \minibox{Ground Truth} & \minibox{DiffHandles} & \minibox{Ditto} &
      \minibox{Flux-Fill} & \minibox{Flux-Kontext} & \minibox{FreeFine} & \minibox{GeoDiffuser}\\
    \end{tabular}%
  }

  \vspace{-1pt}
  \begin{subfigure}{\linewidth}
    \centering
    \animategraphics[width=\linewidth]{12}{images/davis/horse/1/}{1}{1}
  \end{subfigure}

  {%
    \fontsize{8pt}{9.6pt}\selectfont
    \vspace{-4pt}
    \begin{tabular}{@{}*{9}{>{\centering\arraybackslash}m{0.125\linewidth}}@{}}
      \minibox{HqEdit} & \minibox{InsV2V} & \minibox{InsVIE} & \minibox{Lucy-Edit} & \minibox{Qwen-Img-E}& \minibox{Señorita} &
      \minibox{Shape4Motion} & \minibox{ScaleVid (Ours)} \\
    \end{tabular}%
  }

  \vspace{-1pt}
  \begin{subfigure}{\linewidth}
    \centering
    \animategraphics[width=\linewidth]{12}{images/davis/horse/2/}{1}{1}
  \end{subfigure}
  
    \vspace{3pt}
    {\captionsetup{font=footnotesize}
    \fbox{%
      \parbox{0.985\linewidth}{%
        \textbf{Prompt:} \textit{Rescale the \textbf{horse} and \textbf{rider} by $1.1826$ times in width, $2.1752$ times in height, and $1.2873$ times in depth. Keep background unchanged.}
      }%
    }}

  {%
    \fontsize{8pt}{9.6pt}\selectfont 
    \begin{tabular}{@{}*{9}{>{\centering\arraybackslash}m{0.125\linewidth}}@{}}
      \minibox{Source} & \minibox{Ground Truth} & \minibox{DiffHandles} & \minibox{Ditto} &
      \minibox{Flux-Fill} & \minibox{Flux-Kontext} & \minibox{FreeFine} & \minibox{GeoDiffuser}\\
    \end{tabular}%
  }

  \vspace{-1pt}
  \begin{subfigure}{\linewidth}
    \centering
    \animategraphics[width=\linewidth]{12}{images/davis/kid/1/}{1}{1}
  \end{subfigure}

  {%
    \fontsize{8pt}{9.6pt}\selectfont
    \vspace{-4pt}
    \begin{tabular}{@{}*{9}{>{\centering\arraybackslash}m{0.125\linewidth}}@{}}
      \minibox{HqEdit} & \minibox{InsV2V} & \minibox{InsVIE} & \minibox{Lucy-Edit} & \minibox{Qwen-Img-E}& \minibox{Señorita} &
      \minibox{Shape4Motion} & \minibox{ScaleVid (Ours)} \\
    \end{tabular}%
  }

  \vspace{-1pt}
  \begin{subfigure}{\linewidth}
    \centering
    \animategraphics[width=\linewidth]{12}{images/davis/kid/2/}{1}{1}
  \end{subfigure}
  
    \vspace{3pt}
    {\captionsetup{font=footnotesize}
    \fbox{%
      \parbox{0.985\linewidth}{%
        \textbf{Prompt:} \textit{Rescale the \textbf{kid} and \textbf{football} by $0.5851$ times in width, $1.6757$ times in height, and $2.2253$ times in depth. Keep background unchanged.}
      }%
    }}
    
  \caption{Comparison results of \abbr{} and baselines on DAVIS.}
  \label{fig:davis_compare}
\end{figure*}

\begin{figure*}[!htp]
  \centering
  \renewcommand{\arraystretch}{1.0}
  \setlength{\tabcolsep}{0pt}

  \vspace{3pt}

  {%
    \fontsize{8pt}{9.6pt}\selectfont 
    \begin{tabular}{@{}*{9}{>{\centering\arraybackslash}m{0.125\linewidth}}@{}}
      \minibox{Source} & \minibox{Mask} & \minibox{DiffHandles} & \minibox{Ditto} &
      \minibox{Flux-Fill} & \minibox{Flux-Kontext} & \minibox{FreeFine} & \minibox{GeoDiffuser}\\
    \end{tabular}%
  }

  \vspace{-1pt}
  \begin{subfigure}{\linewidth}
    \centering
    \animategraphics[width=\linewidth]{12}{images/pexels/boombox/1/}{1}{1}
  \end{subfigure}

  {%
    \fontsize{8pt}{9.6pt}\selectfont
    \vspace{-4pt}
    \begin{tabular}{@{}*{9}{>{\centering\arraybackslash}m{0.125\linewidth}}@{}}
      \minibox{HqEdit} & \minibox{InsV2V} & \minibox{InsVIE} & \minibox{Lucy-Edit} & \minibox{Qwen-Img-E}& \minibox{Señorita} &
      \minibox{Shape4Motion} & \minibox{ScaleVid (Ours)} \\
    \end{tabular}%
  }

  \vspace{-1pt}
  \begin{subfigure}{\linewidth}
    \centering
    \animategraphics[width=\linewidth]{12}{images/pexels/boombox/2/}{1}{1}
  \end{subfigure}

    \vspace{3pt}
    {\captionsetup{font=footnotesize}
    \fbox{%
      \parbox{0.985\linewidth}{%
        \textbf{Prompt:} \textit{Rescale the \textbf{boombox} by $1.5776$ times in width, $2.2173$ times in height, and $1.3454$ times in depth. Keep background unchanged.}
      }%
    }}

  {%
    \fontsize{8pt}{9.6pt}\selectfont 
    \begin{tabular}{@{}*{9}{>{\centering\arraybackslash}m{0.125\linewidth}}@{}}
      \minibox{Source} & \minibox{Mask} & \minibox{DiffHandles} & \minibox{Ditto} &
      \minibox{Flux-Fill} & \minibox{Flux-Kontext} & \minibox{FreeFine} & \minibox{GeoDiffuser}\\
    \end{tabular}%
  }

  \vspace{-1pt}
  \begin{subfigure}{\linewidth}
    \centering
    \animategraphics[width=\linewidth]{12}{images/pexels/mountain/1/}{1}{1}
  \end{subfigure}

  {%
    \fontsize{8pt}{9.6pt}\selectfont
    \vspace{-4pt}
    \begin{tabular}{@{}*{9}{>{\centering\arraybackslash}m{0.125\linewidth}}@{}}
      \minibox{HqEdit} & \minibox{InsV2V} & \minibox{InsVIE} & \minibox{Lucy-Edit} & \minibox{Qwen-Img-E}& \minibox{Señorita} &
      \minibox{Shape4Motion} & \minibox{ScaleVid (Ours)} \\
    \end{tabular}%
  }

  \vspace{-1pt}
  \begin{subfigure}{\linewidth}
    \centering
    \animategraphics[width=\linewidth]{12}{images/pexels/mountain/2/}{1}{1}
  \end{subfigure}

    \vspace{3pt}
    {\captionsetup{font=footnotesize}
    \fbox{%
      \parbox{0.985\linewidth}{%
        \textbf{Prompt:} \textit{Rescale the \textbf{mountain} by $1.7379$ times in width, $1.0378$ times in height, and $1.5644$ times in depth. Keep background unchanged.}
      }%
    }}

  {%
    \fontsize{8pt}{9.6pt}\selectfont 
    \begin{tabular}{@{}*{9}{>{\centering\arraybackslash}m{0.125\linewidth}}@{}}
      \minibox{Source} & \minibox{Mask} & \minibox{DiffHandles} & \minibox{Ditto} &
      \minibox{Flux-Fill} & \minibox{Flux-Kontext} & \minibox{FreeFine} & \minibox{GeoDiffuser}\\
    \end{tabular}%
  }

  \vspace{-1pt}
  \begin{subfigure}{\linewidth}
    \centering
    \animategraphics[width=\linewidth]{12}{images/pexels/bottle/1/}{1}{1}
  \end{subfigure}

  {%
    \fontsize{8pt}{9.6pt}\selectfont
    \vspace{-4pt}
    \begin{tabular}{@{}*{9}{>{\centering\arraybackslash}m{0.125\linewidth}}@{}}
      \minibox{HqEdit} & \minibox{InsV2V} & \minibox{InsVIE} & \minibox{Lucy-Edit} & \minibox{Qwen-Img-E}& \minibox{Señorita} &
      \minibox{Shape4Motion} & \minibox{ScaleVid (Ours)} \\
    \end{tabular}%
  }

  \vspace{-1pt}
  \begin{subfigure}{\linewidth}
    \centering
    \animategraphics[width=\linewidth]{12}{images/pexels/bottle/2/}{1}{1}
  \end{subfigure}
  
    \vspace{3pt}
    {\captionsetup{font=footnotesize}
    \fbox{%
      \parbox{0.985\linewidth}{%
        \textbf{Prompt:} \textit{Rescale the \textbf{bottle} by $1.3458$ times in width, $0.6378$ times in height, and $0.4284$ times in depth. Keep background unchanged.}
      }%
    }}
    
  {%
    \fontsize{8pt}{9.6pt}\selectfont 
    \begin{tabular}{@{}*{9}{>{\centering\arraybackslash}m{0.125\linewidth}}@{}}
      \minibox{Source} & \minibox{Mask} & \minibox{DiffHandles} & \minibox{Ditto} &
      \minibox{Flux-Fill} & \minibox{Flux-Kontext} & \minibox{FreeFine} & \minibox{GeoDiffuser}\\
    \end{tabular}%
  }

  \vspace{-1pt}
  \begin{subfigure}{\linewidth}
    \centering
    \animategraphics[width=\linewidth]{12}{images/pexels/flower/1/}{1}{1}
  \end{subfigure}

  {%
    \fontsize{8pt}{9.6pt}\selectfont
    \vspace{-4pt}
    \begin{tabular}{@{}*{9}{>{\centering\arraybackslash}m{0.125\linewidth}}@{}}
      \minibox{HqEdit} & \minibox{InsV2V} & \minibox{InsVIE} & \minibox{Lucy-Edit} & \minibox{Qwen-Img-E}& \minibox{Señorita} &
      \minibox{Shape4Motion} & \minibox{ScaleVid (Ours)} \\
    \end{tabular}%
  }

  \vspace{-1pt}
  \begin{subfigure}{\linewidth}
    \centering
    \animategraphics[width=\linewidth]{12}{images/pexels/flower/2/}{1}{1}
  \end{subfigure}
  
    \vspace{3pt}
    {\captionsetup{font=footnotesize}
    \fbox{%
      \parbox{0.985\linewidth}{%
        \textbf{Prompt:} \textit{Rescale the \textbf{leaves} and \textbf{flowers} by $2.2579$ times in width, $1.3811$ times in height, and $1.0421$ times in depth. Keep background unchanged.}
      }%
    }}

  {%
    \fontsize{8pt}{9.6pt}\selectfont 
    \begin{tabular}{@{}*{9}{>{\centering\arraybackslash}m{0.125\linewidth}}@{}}
      \minibox{Source} & \minibox{Mask} & \minibox{DiffHandles} & \minibox{Ditto} &
      \minibox{Flux-Fill} & \minibox{Flux-Kontext} & \minibox{FreeFine} & \minibox{GeoDiffuser}\\
    \end{tabular}%
  }

  \vspace{-1pt}
  \begin{subfigure}{\linewidth}
    \centering
    \animategraphics[width=\linewidth]{12}{images/pexels/man1/1/}{1}{1}
  \end{subfigure}

  {%
    \fontsize{8pt}{9.6pt}\selectfont
    \vspace{-4pt}
    \begin{tabular}{@{}*{9}{>{\centering\arraybackslash}m{0.125\linewidth}}@{}}
      \minibox{HqEdit} & \minibox{InsV2V} & \minibox{InsVIE} & \minibox{Lucy-Edit} & \minibox{Qwen-Img-E}& \minibox{Señorita} &
      \minibox{Shape4Motion} & \minibox{ScaleVid (Ours)} \\
    \end{tabular}%
  }

  \vspace{-1pt}
  \begin{subfigure}{\linewidth}
    \centering
    \animategraphics[width=\linewidth]{12}{images/pexels/man1/2/}{1}{1}
  \end{subfigure}
  
    \vspace{3pt}
    {\captionsetup{font=footnotesize}
    \fbox{%
      \parbox{0.985\linewidth}{%
        \textbf{Prompt:} \textit{Rescale the \textbf{man on the left} by $1.4131$ times in width, $0.7471$ times in height, and $2.3173$ times in depth. Keep background unchanged.}
      }%
    }}
    
  \caption{Comparison results of \abbr{} and baselines on Pexels.}
  \label{fig:pexels_compare}
\end{figure*}

\clearpage
\begin{figure*}[!htp]
  \centering
  \renewcommand{\arraystretch}{1.0}
  \setlength{\tabcolsep}{0pt}
      \includegraphics[width=\linewidth]{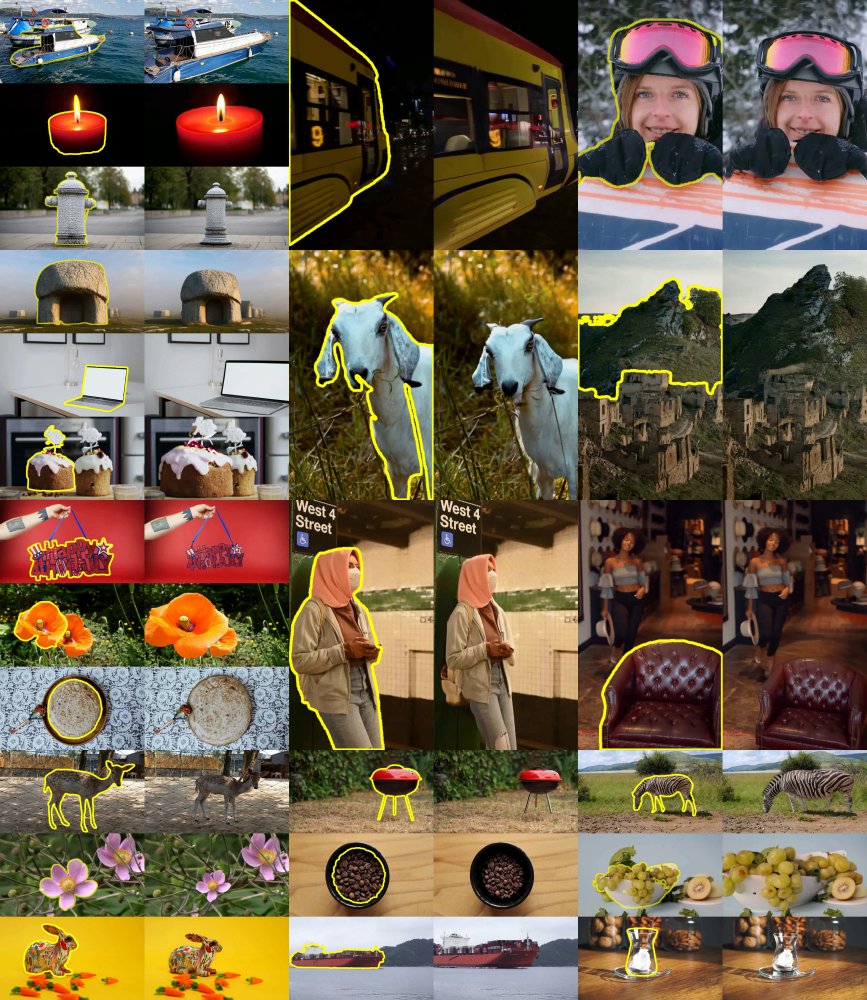}
  \caption{Visual results of \abbr{}. In each pair, the left frame shows the source with the target object highlighted in yellow, and the right frame shows the ScaleVid output.}
  \label{fig:eof}
\end{figure*}


\end{document}